\documentclass{elsarticle}
\usepackage[T1]{fontenc}
\usepackage[english]{babel}
\usepackage{lineno,hyperref}
\modulolinenumbers[5]
\usepackage{mathrsfs}
\usepackage{amsmath,amsfonts,amssymb,amsthm}
\usepackage{newpxtext}
\usepackage{mathabx}
\usepackage{multicol}
\usepackage{xcolor}
\usepackage{tikz}
\usepackage{bm}
\usepackage{pgfplots}
\usepackage{xargs}                      % Use more than one optional parameter in a new commands
\usepackage[colorinlistoftodos,prependcaption,textsize=tiny,textwidth=4.5cm]{todonotes}
\newcommandx{\SB}[2][1=]{\todo[linecolor=red,backgroundcolor=red!25,bordercolor=red,#1]{SB:#2}}
\newcommandx{\AG}[2][1=]{\todo[linecolor=blue,backgroundcolor=blue!25,bordercolor=blue,#1]{AG:#2}}
\newcommandx{\MS}[2][1=]{\todo[linecolor=OliveGreen,backgroundcolor=OliveGreen!25,bordercolor=OliveGreen,#1]{MS: #2}}
\newcommandx{\LS}[2][1=]{\todo[linecolor=yellow,backgroundcolor=yellow!25,bordercolor=yellow,#1]{LS #2}}
\usepackage{subcaption}
\usepackage{geometry}
\usepackage[shortlabels]{enumitem}
\usepackage{makecell}
\usepackage{array}
\usepackage{pifont}
\usepackage{float}
\usepackage{natbib}
\usepackage{threeparttable,booktabs}
\newdefinition{definition}{Definition}
\newdefinition{example}{Example}

\def\C{\mathcal{C}}

\def\G{\mathcal{G}}

\def\K{\mathcal{K}}
\def\Lg{\mathcal{L}}

\def\Pr{\mathcal{P}}

\def\R{\mathbb{R}}

\def\T{\mathcal{T}}

\def\bt{\mathbf{t}}
\def\bu{\mathbf{u}}
\def\bv{\mathbf{v}}
\def\bv{\mathbf{v}}
\def\bw{\mathbf{w}}

\def\imp{\rightarrow}

\def\bt{\mathbf{t}}
\def\F{\mathcal{F}}

\DeclareMathOperator*{\argmax}{argmax}
\DeclareMathOperator*{\argmin}{argmin}
\DeclareMathOperator*{\Agg}{Agg}
\DeclareMathOperator*{\AggQ}{Agg(Q)}
\DeclareMathOperator*{\All}{Agg(\forall)}
\DeclareMathOperator*{\Exi}{Agg(\exists)}
\DeclareMathOperator*{\Sat}{SatAgg}
\newcommand{\Tnorm}{T}
\newcommand{\Snorm}{S}
\newcommand{\ImpOp}{I}
\newcommand{\Neg}{N}

\newcommand{\Tmin}{T_{M} }
\newcommand{\Tprod}{T_{P}}
\newcommand{\Tluk}{T_{L}}
\newcommand{\Smin}{S_{M} }
\newcommand{\Sprod}{S_{P}}
\newcommand{\Sluk}{S_{L}}
\newcommand{\Ns}{N_{S}}
\newcommand{\Ikd}{I_{KD}}
\newcommand{\Igodel}{I_{G}}
\newcommand{\Ir}{I_{R}}
\newcommand{\Iprod}{I_{P}}
\newcommand{\Iluk}{I_{Luk}}
\newcommand{\Amean}{A_M}
\newcommand{\Apmean}{A_{pM}}
\newcommand{\ApmeanError}{A_{pME}}

\newcommand{\cmark}{\ding{51}}
\newcommand{\xmark}{\ding{55}}

\newlength\myheight
\newlength\mydepth
\settototalheight\myheight{Xygp}
\newcommand*\inlinegraphics[1]{%
  \settototalheight\myheight{Xygp}%
  \settodepth\mydepth{Xygp}%
  \raisebox{-\mydepth}{\includegraphics[height=\myheight]{#1}}%
}

\begin{document}

\title{Logic Tensor Networks}

\author[1,2]{Samy Badreddine\corref{cor1}}
\ead{badreddine.samy@gmail.com}

\author[3]{Artur d'Avila Garcez}
\ead{a.garcez@city.ac.uk}

\author[4]{Luciano Serafini}
\ead{serafini@fbk.eu}

\author[1,2]{Michael Spranger}
\ead{michael.spranger@sony.com}

\cortext[cor1]{Corresponding author}
\affiliation[1]{organization={Sony Computer Science Laboratories Inc},
    addressline={3-14-13 Higashigotanda},
    postcode={141-0022},
    city={Tokyo},
    country={Japan}}

\affiliation[2]{organization={Sony AI Inc},
    addressline={1-7-1 Konan},
    postcode={108-0075},
    city={Tokyo},
    country={Japan}}

\affiliation[3]{
    organization={City, University of London},
    addressline={Northampton Square},
    postcode={EC1V 0HB},
    city={London},
    country={United Kingdom}}

\affiliation[4]{organization={Fondazione Bruno Kessler},
    addressline={Via Sommarive 18},
    postcode={38123},
    city={Trento},
    country={Italy}}

\begin{abstract}
Attempts at combining logic and neural networks into neurosymbolic approaches have been on the increase in recent years. 
In a neurosymbolic system, symbolic knowledge assists deep learning, which typically uses a sub-symbolic distributed representation, to learn and reason at a higher level of abstraction.
We present Logic Tensor Networks (LTN), a neurosymbolic framework that supports querying, learning and reasoning with both rich data and abstract knowledge about the world. 
LTN introduces a fully differentiable logical language, called Real Logic, whereby the elements of a first-order logic signature are grounded onto data using neural computational graphs and first-order fuzzy logic semantics.
We show that LTN provides a uniform language to represent and compute efficiently many of the most important AI tasks
such as multi-label classification, relational learning, data clustering, semi-supervised learning, regression, embedding learning and query answering.
We implement and illustrate each of the above tasks with several simple explanatory examples using TensorFlow 2. The results indicate that LTN can be a general and powerful framework for neurosymbolic AI.
\end{abstract}


\maketitle

\section{Introduction}

Artificial Intelligence (AI) agents are required to learn from their surroundings and reason about what has been learned to make decisions, act in the world, or react to various stimuli.
The latest Machine Learning (ML) has adopted mostly a pure sub-symbolic learning approach.
Using distributed representations of entities, the latest ML performs quick decision-making without building a comprehensible model of the world.
While achieving impressive results in computer vision, natural language, game playing, and multimodal learning, such approaches are known to be data inefficient and to struggle at out-of-distribution generalization. Although the use of appropriate inductive biases can alleviate such shortcomings, in general, sub-symbolic models lack comprehensibility. By contrast, symbolic AI is based on rich, high-level representations of the world that use human-readable symbols.
By \emph{rich knowledge}, we refer to logical representations which are more expressive than propositional logic or propositional probabilistic approaches, and which can express knowledge using full first-order logic, including universal and existential quantification ($\forall x$ and $\exists y$), arbitrary $n$-ary relations over variables, e.g. $R(x,y,z,\dots)$, and function symbols, 
e.g. $\mathrm{fatherOf}(x)$, $x+y$, etc.
Symbolic AI has achieved success at theorem proving, logical inference, and verification.
However, it also has shortcomings when dealing with incomplete knowledge. It can be inefficient with large amounts of inaccurate data and lack robustness to outliers. Purely symbolic decision algorithms usually have high computational complexity making them impractical for the real world. 
It is now clear that the predominant approach to ML, where learning is based on recognizing the latent structures hidden in the data, is insufficient and may benefit from symbolic AI \cite{3rdWave}. In this context, neurosymbolic AI, which stems from \emph{neural} networks and \emph{symbolic} AI, attempts to combine the strength of both paradigms (see \cite{DBLP:journals/flap/GarcezGLSST19,DBLP:conf/ijcai/LambGGPAV20,raedt2020statistical} for recent surveys).
That is to say, combine reasoning with complex representations of knowledge (knowledge-bases, semantic networks, ontologies, trees, and graphs) with learning from complex data (images, time series, sensorimotor data, natural language).
Consequently, a main challenge for neurosymbolic AI is the grounding of symbols, including constants, functional and relational symbols, into real data, which is akin to the longstanding \emph{symbol grounding} problem \cite{harnad1990symbol}.

Logic Tensor Networks (LTN) are a neurosymbolic framework and computational model that supports learning and reasoning about data with rich knowledge. 
In LTN, one can represent and effectively compute the most important tasks of deep learning with a fully differentiable first-order logic language, called Real Logic, which adopts infinitely many truth-values in the interval [0,1] \cite{fagin2020,manyvalue}. 
In particular, LTN supports the specification and computation of the following AI tasks uniformly using the same language: 
  data clustering, 
  classification, 
  relational learning, 
  query answering, 
  semi-supervised learning, 
  regression, and 
  embedding learning.

LTN and Real Logic were first introduced in \cite{serafini2016learning}.
Since then, LTN has been applied to different AI tasks involving perception, learning, and reasoning about relational knowledge. 
In \cite{donadello2019compensating,LTNIJCAI}, LTN was applied to semantic image interpretation whereby relational knowledge about objects was injected into deep networks for object relationship detection. 
In \cite{ReasoningLTN}, LTN was evaluated on its capacity to perform reasoning about ontological knowledge. 
Furthermore, \cite{bianchi2019complementing} shows how LTN can be used to learn an embedding of concepts into a latent real space by taking into consideration ontological knowledge about such concepts. 
In \cite{badreddine_injecting_2019}, LTN is used to annotate a reinforcement learning environment with prior knowledge and incorporate latent information into an agent.
In \cite{manigrasso_faster-ltn_2021}, authors embed LTN in a state-of-the-art convolutional object detector.
Extensions and generalizations of LTN have also been proposed in the past years, such as LYRICS \cite{marra2019lyrics} and \emph{Differentiable Fuzzy Logic} (DFL) \cite{vanKrieken2020Analyzing,van_krieken_analyzing_2020}. 
LYRICS provides an input language allowing one to define background knowledge using a first-order logic where predicate and function symbols are grounded onto any computational graph.
DFL analyzes how a large collection of fuzzy logic operators behave in a differentiable learning setting. 
DFL also introduces new semantics for fuzzy logic implications called sigmoidal implications, and it shows that such semantics outperform other semantics in several semi-supervised machine learning tasks.

This paper provides a thorough description of the full formalism and several extensions of LTN. We
show using an extensive set of explanatory examples, how LTN can be applied to solve many ML tasks with the help of logical knowledge.
In particular, the earlier versions of LTN have been extended with:
(1) \emph{Explicit domain declaration:} constants, variables, functions and predicates are now domain typed (e.g. the constants \emph{John} and \emph{Paris} can be from the domain of \emph{person} and \emph{city}, respectively). The definition of structured domains is also possible (e.g. the domain \emph{couple} can be defined as the Cartesian product of two domains of persons); (2) \emph{Guarded quantifiers:} guarded universal and existential quantifiers now allow the user to limit the quantification to the elements that satisfy some Boolean condition, e.g. $\forall x : \mathrm{age}(x) < 10 \: (\mathrm{playsPiano}(x) \rightarrow \mathrm{enfantProdige}(x))$ restricts the quantification to the cases where $age$ is lower than 10; 
(3) \emph{Diagonal quantification:} Diagonal quantification allows the user to write statements about specific tuples extracted in order from $n$ variables. For example, if the variables $\mathrm{capital}$ and $\mathrm{country}$ both have $k$ instances such that the $i$-th instance of $\mathrm{capital}$ corresponds to the $i$-th instance of $\mathrm{country}$, one can write $\forall \mathrm{Diag}(\mathrm{capital},\mathrm{country}) \: \mathrm{capitalOf}(\mathrm{capital},\mathrm{country})$. 

Inspired by the work of \cite{van_krieken_analyzing_2020}, this paper also extends the product t-norm configuration of LTN with the generalized mean aggregator, and it introduces solutions to the vanishing or exploding gradient problems. 
Finally, the paper formally defines a semantic approach to \emph{refutation-based reasoning} in Real Logic to verify if a statement is a logical consequence of a knowledge base.
Example \ref{s:ex_reasoning} proves that this new approach can better capture logical consequences compared to simply querying unknown formulas after learning (as done in \cite{ReasoningLTN}).

The new version of LTN has been implemented in TensorFlow 2 \cite{tensorflow2015-whitepaper}. Both the LTN library and the code for the examples used in this paper are available at \url{https://github.com/logictensornetworks/logictensornetworks}.

The remainder of the paper is organized as follows: 
In Section \ref{s:reallogic}, we define and illustrate Real Logic as a fully-differentiable first-order logic. In Section \ref{s:ltn}, we specify learning and reasoning in Real Logic and its modeling into deep networks with Logic Tensor Networks (LTN).
In Section \ref{s:examples}, we illustrate the reach of LTN by investigating a range of learning problems from clustering to embedding learning.
In Section \ref{s:relwork}, we place LTN in the context of the latest related work in neurosymbolic AI. In Section 
\ref{s:concl} we conclude and discuss directions for future work. The Appendix contains information about the implementation of LTN in TensorFlow 2, experimental set-ups, the different options for the differentiable logic operators, and a study of their relationship with gradient computations.

\section{Real Logic}

\def\D{\mathcal{D}}
\def\CD{\mathcal{D^*}}
\def\dom{D}
\def\N{\mathbb{N}}
\def\dim{\mathrm{dim}}
\def\X{\mathcal{X}}
\def\Ran{\mathcal{R}}
\def\card#1{|#1|}
\def\tor{^{\R}}
\def\bi{\mathbf{i}}
\def\bj{\mathbf{j}}
\def\by{\mathbf{y}}
\def\domof{\mathbf{D}}
\def\domofout{\mathbf{D_{out}}}
\def\domofin{\mathbf{D_{in}}}
\def\x{\mathbf{x}}
\def\Gp{\hat{\G}}
\def\phiAgg{\mathbf{A}}
\def\diag{\mathrm{Diag}}
\def\alice{\mathrm{Alice}}
\def\bob{\mathrm{Bob}}
\def\charlie{\mathrm{Charlie}}
\def\people{\mathrm{People}}
\def\town{\mathrm{Town}}
\def\rome{\mathrm{Rome}}
\def\seoul{\mathrm{Seoul}}
\def\LivesIn{\mathrm{lives\_in}}
\def\IsFriend{\mathrm{is\_friend}}
\def\Italian{\mathrm{Italian}}
\def\digitsImage{\mathrm{digit\_images}}

\label{s:reallogic}

\subsection{Syntax}
Real Logic forms the basis of Logic Tensor Networks. Real Logic is defined on a first-order language $\Lg$ with a signature that contains a
set $\C$ of constant symbols (objects), a set $\F$ of functional
symbols, a set $\Pr$ of relational symbols (predicates), and a set $\X$ of variable
symbols. $\Lg$-formulas allow us to specify relational knowledge with variables, 
e.g. the atomic formula $\IsFriend(v_1,v_2)$ may state that the person $v_1$ is a friend of the person $v_2$, 
the formula $\forall x \forall y(\IsFriend(x,y)\imp \IsFriend(y,x))$
states that the relation $\IsFriend$ is symmetric, 
and the formula $\forall x (\exists y (\Italian(x)\land\IsFriend(x,y)))$
states that every person has a friend that is Italian. 
Since we are interested in learning and reasoning in real-world scenarios where degrees of truth are often fuzzy and exceptions are present, 
formulas can be partially true, and therefore we adopt fuzzy semantics.

Objects can be of different types. Similarly, functions and
predicates are typed. Therefore, we assume there exists a non-empty set of symbols $\D$ called \emph{domain symbols}.
To assign types to the elements of $\Lg$ we introduce
the functions $\domof$, $\domofin$ and $\domofout$ such that:
\begin{itemize}
\item $\domof: \X\cup\C\rightarrow\D$. Intuitively, $\domof(x)$ and $\domof(c)$ returns the domain of a variable $x$ or a constant $c$.
\item $\domofin: \F\cup\Pr\rightarrow\D^*$, where $\D^*$ is the Kleene star of $\D$, that is the set of all finite sequences of symbols in $\D$.
Intuitively, $\domofin(f)$ and $\domofin(p)$ returns the domains of the arguments of a function $f$ or a predicate $p$.
If $f$ takes two arguments (for example, $f(x,y)$), $\domofin(f)$ returns two domains, one per argument.
\item $\domofout:\F\rightarrow\D$. Intuitively, $\domofout(f)$ returns the range of a function symbol.
\end{itemize}

Real Logic may also contain propositional
variables, as follows: if $P$ is a 0-ary predicate with
  $\domofin(P)=\left<\:\right>$ (the empty sequence of domains) then $P$ is
a propositional variable (an atom with truth-value in the interval [0,1]). 

A term is constructed recursively in the usual way from constant symbols, variables, and function symbols. 
An expression formed by applying a predicate symbol to an appropriate number of terms with appropriate domains is called an atomic formula, which evaluates to $\mathit{true}$ or $\mathit{false}$ in classical logic and a number in $[0,1]$ in the case of Real Logic.
We define the set of terms 
of the language as follows:

\begin{itemize}
\item each element $t$ of $\X\cup\C$ is a term of the domain $\domof(t)$;
\item if $t_i$ is a term of domain $\domof(t_i)$ for $1\leq i\leq n$
  then $t_1t_2\dots t_n$ (the sequence composed of $t_1$ followed by $t_2$ and so on, up to $t_n$) is a term of the domain
  $\domof(t_1)\domof(t_2)\dots\domof(t_n)$;
\item if $t$ is a term of the domain $\domofin(f)$ then $f(t)$
  is a term of the domain $\domofout(f)$.
\end{itemize}

We allow the following set of formula in $\Lg$:
\begin{itemize}
\item $t_1=t_2$ is an atomic formula for any terms $t_1$ and $t_2$
  with $\domof(t_1) = \domof(t_2)$;
\item $p(t)$ is an atomic formula if $\domof(t)=\domofin(p)$;
\item If $\phi$ and $\psi$ are formula and $x_1,\dots,x_n$ are
  $n$ distinct variable symbols then $\diamond\phi$, 
  $\phi\circ\psi$ and $Q x_1\dots x_n\phi$ are formula, where 
$\diamond$ is a unary connective, $\circ$ is a binary connective and 
$Q$ is a quantifier. 
\end{itemize}
We use $\diamond\in\{\lnot\}$ (negation),
$\circ\in\{\land,\lor,\imp,\leftrightarrow\}$ (conjunction, disjunction, implication and bi-conditional, respectively) and
$Q\in\{\forall,\exists\}$ (universal and existential, respectively).

\begin{example}
  Let $\town$ denote the domain of towns in the world and $\people$ denote the
  domain of living people. Suppose that $\Lg$ contains the
  constant symbols $\alice$, $\bob$ and $\charlie$ of domain $\people$, and
  $\rome$ and $\seoul$ of domain $\town$. 
  Let $x$ be a variable of domain $\people$ and $u$ be a variable
  of domain $\town$. The term  $x,u$ (i.e. the sequence $x$ followed by $u$) has domain $\people,\town$ which denotes the Cartesian product between $\people$ and $\town$ ($\people$ $\bigtimes$ $\town$). 
  $\alice,\rome$ is interpreted as an element of the domain $\people,\town$.
  Let $\LivesIn$ be a predicate with input domain $\domofin(\LivesIn) = \people,\town$.
  $\LivesIn(\alice,\rome)$ is a well-formed expression, whereas $\LivesIn(\bob,\charlie)$ is not.

\end{example}

\subsection{Semantics of Real Logic}
The semantics of Real Logic departs from the standard
abstract semantics of First-order Logic (FOL). In Real Logic, domains are interpreted concretely by tensors in the real field.\footnote{In the rest of the paper, we commonly use "tensor" to designate "tensor in the real field".} Every object denoted by constants, variables, and terms, is interpreted as a tensor of real values.
Functions are interpreted as real functions or tensor operations.  
Predicates are interpreted as functions or tensor operations
projecting onto a value in the interval $[0,1]$.

To emphasize the fact that in Real Logic symbols are grounded onto real-valued features, we use the term \emph{grounding}, denoted by $\G$, in place of
\emph{interpretation}\footnote{An interpretation is an assignment of
  truth-values $\mathit{true}$ or $\mathit{false}$, or in the case of
  Real Logic a value in [0,1], to a formula. A model is an
  interpretation that maps a formula to $\mathit{true}$}.
Notice that this is different from the common use of the term \emph{grounding} in logic, which indicates the operation of replacing the variables of a term or formula with constants or terms containing no variables. To avoid
  confusion, we use the synonym \emph{instantiation} for this purpose. 
    $\G$ associates a tensor of real numbers to any term of $\Lg$, and a real number in the interval $[0,1]$ to any formula $\phi$ of $\Lg$.
Intuitively, $\G(t)$ are the numeric features of the objects denoted by $t$, and $\G(\phi)$ represents the system's degree of confidence in the truth of $\phi$; the higher the value, the higher the confidence.

\subsubsection{Grounding domains and the signature}
A grounding for a logical language $\Lg$ on the set of domains $\D$
provides the interpretation of both the domain symbols in $\D$ and
the non-logical symbols in $\Lg$. 

\begin{definition}
  A grounding $\G$ associates to each domain $\dom \in \D$ a  
  set  $\G(\dom)\subseteq\bigcup\limits_{n_1 \dots n_d \in \N^*} \R^{n_1 \times \dots \times n_d}$. 
  For every $\dom_1\dots \dom_n\in\D^*$, $\G(\dom_1\dots \dom_n)=\bigtimes_{i=1}^n\G(\dom_i)$, that is $\G(\dom_1)\bigtimes\G(\dom_2)\bigtimes...\bigtimes\G(\dom_n)$.
\end{definition}
Notice that the elements in $\G(\dom)$ may be tensors
of any rank $d$ and any dimensions $n_1 \times \dots \times n_d$, as $\N^*$ denotes the Kleene star of $\N$.\footnote{ A tensor of rank $0$ corresponds to a scalar, a
  tensor of rank $1$ to a vector, a tensor of rank $2$ to a matrix and
  so forth, in the usual way.}
\def\IsDigit{\mathrm{is\_digit}}
\def\digits{\mathrm{digits}}
\begin{example}
Let $\digitsImage$ denote a domain of images of handwritten digits.
If we use images of $256 \times 256$ RGB pixels, then $\G(\digitsImage) \subseteq \R^{256\times 256 \times 3}$. 
Let us consider the predicate $\IsDigit(\includegraphics[width=1em]{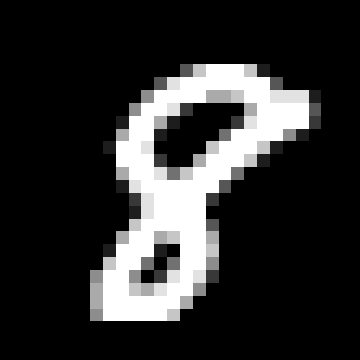},8)$.
The terms $\includegraphics[width=1em]{figures/mnist_eight.png},8$ have domains $\digitsImage,\digits$.
Any input to the predicate is a tuple in $\G(\digitsImage,\digits) = \G(\digitsImage) \times \G(\digits)$.
\end{example}

A grounding assigns to each constant symbol $c$, a tensor $\G(c)$ in
the domain $\G(\domof(c))$; 
It assigns to a variable $x$ a finite sequence of tensors $d_1 \dots d_k$, each in $\G(\domof(x))$.
These tensors represent the instances of $x$.
Differently from in FOL where a variable is assigned to a single value of the domain of interpretations at a time, in Real Logic a variable is assigned to a sequence of values in its domain, the $k$ examples of $x$.
A grounding assigns to a function symbol $f$ a function taking tensors from $\G(\domofin(f))$ as input, and producing a tensor in $\G(\domofout(f))$ as output.
Finally, a grounding assigns to a predicate symbol $p$ a function taking tensors from $\G(\domofin(p))$ as input, and producing a truth-value in the interval $[0,1]$ as output.

\begin{definition}\label{def:grounding}
    A \emph{grounding $\G$} of $\Lg$ is a function defined on the signature 
    of $\Lg$ that satisfies the following conditions:
    \begin{enumerate}
        \item $\G(x) = \left<d_1 \dots d_k\right>
        \in \bigtimes_{i=1}^{k}\G(\domof(x))$ for
          every variable symbol $x\in\X$, with $k \in \N_0^+$. Notice that $\G(x)$ is a sequence and not a set, meaning that the same value of $\G(\domof(x))$ can occur multiple times in $\G(x)$, as is usual in a Machine Learning data set with ``attributes'' and ``values'';
        \item $\G(f)\in \G(\domofin(f)) \rightarrow \G(\domofout(f))$ for every function symbol $f\in\F$;
        \item $\G(p)\in \G(\domofin(p)) \rightarrow [0,1]$\/ for every predicate symbol $p\in\Pr$.
\end{enumerate}
\end{definition}

If a grounding depends on a set of parameters $\theta$, we denote it as $\G_\theta(\cdot)$ or $\G(\cdot \mid \theta)$ interchangeably.
Section \ref{s:examples} describes how such parameters can be learned using the concept of satisfiability.

\subsubsection{Grounding terms and atomic formulas}\label{GroundingTerms}
We now extend the definition of grounding to all first-order terms and
atomic formulas.
Before formally defining these groundings, we describe on a high level what happens when grounding terms that contain free variables.
\footnote{
  We assume the usual syntactic definition of free and bound variables in FOL. A variable is free if it is not bound by a quantifier ($\forall,\exists$).
} 
\def\height{\mathrm{height}}

Let $x$ be a variable that denotes people.
As explained in Definition \ref{def:grounding}, $x$ is grounded as an explicit sequence of $k$ instances ($k = |\G(x)|$).
Consequently, a term $\height(x)$ is also grounded in $k$ height values, each corresponding to one instance.
We can generalize to expressions with multiple free variables, as shown in Example \ref{example:formulas_tensors}.

In the formal definition below, instead of considering a single term at a time,
it is convenient to consider sequences of terms
$\bt = t_1t_2\dots t_k$ and define the grounding on $\bt$ (with the
definition of the grounding of a single term being derived as a
special case).
The fact that the sequence of terms $\bt$ contains $n$ distinct
variables $x_1,\dots,x_n$ is denoted by $\bt(x_1,\dots,x_n)$. The grounding of $\bt(x_1,\dots,x_n)$, denoted by
$\G(\bt(x_1,\dots,x_n))$, is a tensor with $n$ corresponding axes, one
for each free variable, defined as follows:

\begin{definition}
  Let $\bt(x_1,\dots,x_n)$ be a sequence $t_1\dots t_m$ of $m$ terms
  containing $n$ distinct variables
  $x_1,\dots,x_n$. Let each term $t_i$ in $\bt$ contain $n_i$ variables
  $x_{j_{i1}},\dots,x_{j_{in_i}}$. 
  \begin{itemize}
\item $\G(\bt)$ is a tensor with dimensions
  $(|\G(x_1)|,\dots,|\G(x_n)|)$ such that the element of this tensor indexed by
  $k_1,\dots,k_n$, written as $\G(\bt)_{k_1 \dots k_n}$, is equal to
  the concatenation of $\G(t_i)_{k_{j_{i1}}\dots k_{j_{in_i}}}$ for
  $1\leq i\leq m$;

\item $\G(f(\bt))_{i_1 \dots i_n} = \G(f)(\G(\bt)_{i_1 \dots i_n})$,
  i.e. the element-wise
  application of $\G(f)$ to $\G(\bt)$;
\item $\G(p(\bt))_{i_1 \dots i_n} = \G(p)(\G(\bt)_{i_1 \dots i_n})$,
  i.e. the element-wise
  application of $\G(p)$ to $\G(\bt)$.
\end{itemize}
\end{definition}    
If term $t_i$ contains $n_i$ variables $x_{j_1},\dots,x_{j_{n_i}}$
selected from $x_1, \dots, x_n$ then
$\G(t_i)_{k_{j_1} \dots k_{j_{n_i}}}$ can be obtained from
$\G(\bt)_{i_1 \dots i_n}$ with an appropriate mapping of indices $i$
to $k$.

\newpage
\begin{example}
  \label{example:formulas_tensors}
  Suppose that $\Lg$ contains the variables $x$ and $y$, the function $f$, the predicate $p$ and the set of domains $\D=\{V,W\}$. 
  Let $\domof(x)=V$, $\domof(y)=W$,  $\domofin(f)=V W$, $\domofout(f)=W$ and $\domof(p)=VW$. 
  In what follows, an example of the grounding of $\Lg$ and $\D$ is shown on the left, and the  grounding of some examples of possible terms and atomic formulas is shown on the right.

  \begin{center}
    \begin{minipage}[t]{.40\textwidth}
    \begin{align*}
        \G(V) & = \R^+ \\
        \G(W) & = \R^- \\
        \G(x) & = \left< v_1, v_2, v_3 \right> \\
        \G(y) &  = \left< w_1, w_2 \right> \\
        \G(p) & : x,y\mapsto \sigma(x+y)   \\
        \G(f) & : x,y\mapsto x\cdot y  \\
      \end{align*}
    \end{minipage}
    \quad
    \begin{minipage}[t]{.50\textwidth}
      \begin{align*}  
        \G(f(x,y)) & = \begin{pmatrix}
          v_1\cdot w_1 & v_1\cdot w_2 & \\
          v_2\cdot w_1 & v_2\cdot w_2 & \\
          v_3\cdot w_1 & v_3\cdot w_2 & 
        \end{pmatrix} 
        \\
        \\ 
        \G(p(x,f(x,y))) & = \begin{pmatrix}
          \scriptstyle{\sigma(v_1+ v_1\cdot w_1)} &  \scriptstyle{\sigma(v_1+v_1\cdot w_2)}  \\
          \scriptstyle{\sigma(v_2+ v_2\cdot w_1)} &  \scriptstyle{\sigma(v_2+v_2\cdot w_2)}  \\
          \scriptstyle{\sigma(v_3+ v_3\cdot w_1)} &  \scriptstyle{\sigma(v_3+v_3\cdot w_2)} 
        \end{pmatrix} \\
      \end{align*}
      \end{minipage}
  \end{center}

\begin{figure}
    \centering
    \includegraphics[width=\textwidth]{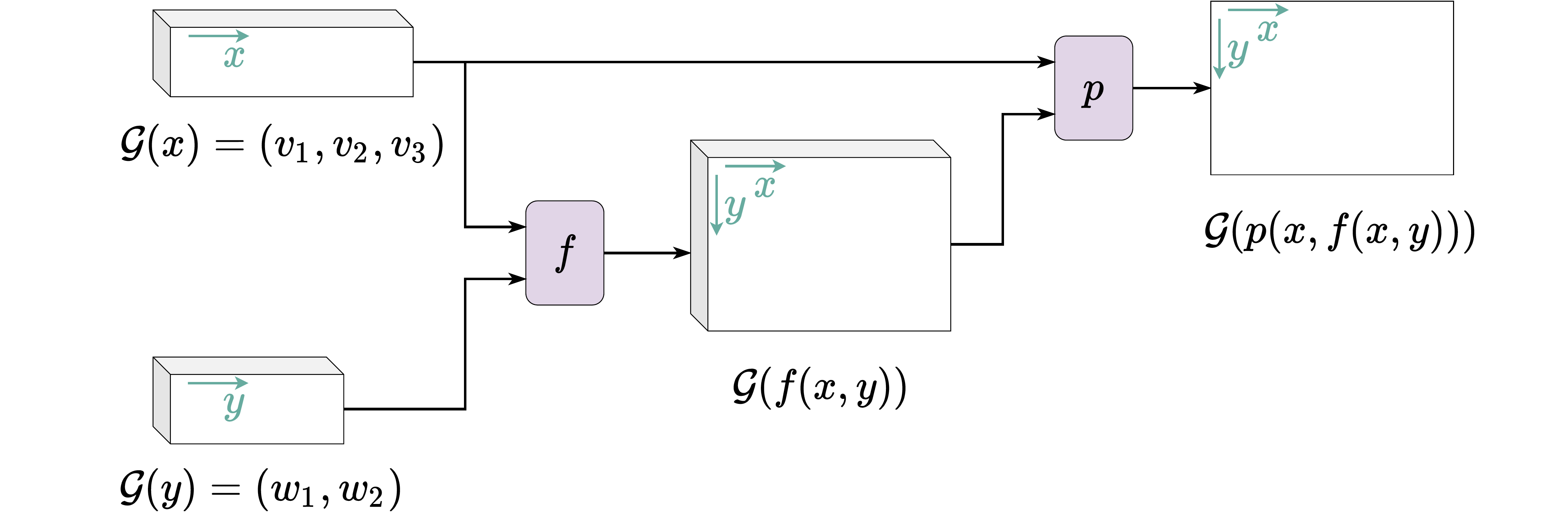}
    \caption{
      Illustration of Example \ref{example:formulas_tensors}:
      \protect\inlinegraphics{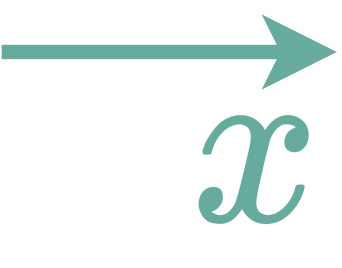} and \protect\inlinegraphics{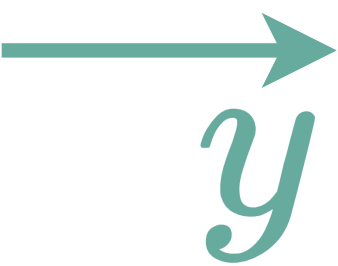} indicate dimensions associated with the free variables $x$ and $y$. 
      A tensor representing a term that includes a free variable $x$ will have an axis \protect\inlinegraphics{figures/illustration_xaxis.PNG}. 
      One can index \protect\inlinegraphics{figures/illustration_xaxis.PNG} to obtain results calculated using each of the $v_1$, $v_2$ or $v_3$ values of $x$. 
      In our graphical convention, the depth of the boxes indicates that the tensor can have \emph{feature dimensions} (refer to the end of Example \ref{example:formulas_tensors}).
    }
    \label{f:illustration_ltn_basic}
\end{figure}

Notice the dimensions of the results.
$\G(f(x,y))$ and $\G(p(x,f(x,y)))$ return $\lvert \G(x) \rvert \times \lvert \G(y) \rvert = 3 \times 2$ values, one for each combination of individuals that occur in the variables.
For functions, we can have additional dimensions associated to the output domain.
Let us suppose a different grounding such that $\G(\domofout(f)) = \R^m$. 
Then the dimensions of $\G(f(x,y))$ would have been $\lvert \G(x) \rvert \times \lvert \G(y) \rvert \times m$, 
where $\lvert \G(x) \rvert \times \lvert \G(y) \rvert $ are the dimensions for indexing the free variables and $m$ are dimensions associated to the output domain of $f$. 
Let us call the latter \emph{feature dimensions}, as captioned in Figure \ref{f:illustration_ltn_basic}.
Notice that $\G(p(x,f(x,y)))$  will always return a tensor with the exact dimensions $\lvert \G(x) \rvert \times \lvert \G(y) \rvert \times 1$ because, under any grounding, a predicate always returns a value in $[0,1]$.
Therefore, as the "feature dimensions" of predicates is always $1$, we choose to "squeeze it" and not to represent it in our graphical convention 
(see Figure \ref{f:illustration_ltn_basic}, the box output by the predicate has no depth).
\end{example}

\subsubsection{Connectives and Quantifiers}

\def\fuzzyop{\mathrm{FuzzyOp}}

\begin{figure}
  \centering
  \includegraphics[width=0.8\textwidth]{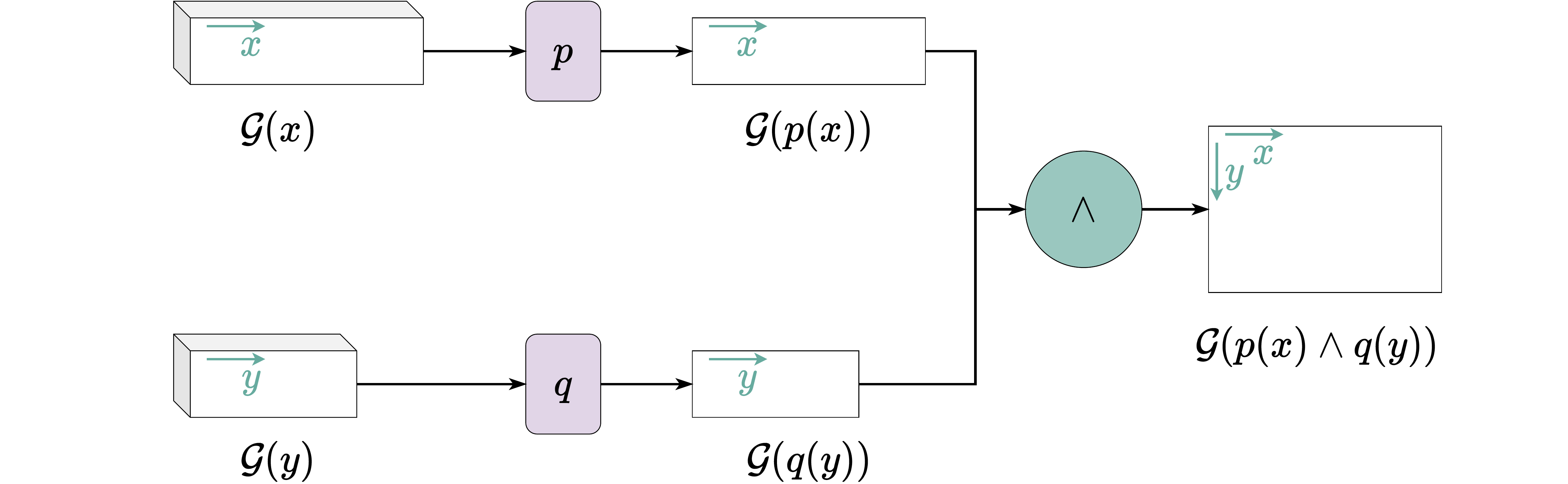}
  \caption{Illustration of an element-wise operator implementing conjunction ($p(x) \land q(y)$). 
  We assume that $x$ and $y$ are two different variables.
  The result has one number in the interval $[0,1]$ to every combination of individuals from $\G(x)$ and $\G(y)$.}
  \label{f:illustration_ltn_connect}
\end{figure}

The semantics of the connectives is defined according to the semantics
of first-order fuzzy logic \cite{hajek_metamathematics_1998}.
Conjunction ($\land$), disjunction ($\lor$), implication ($\imp$) and
negation ($\lnot$) are associated, respectively, with a t-norm
($\Tnorm$), a t-conorm ($\Snorm$), a fuzzy implication ($\ImpOp$) and
a fuzzy negation ($\Neg$) operation
$\fuzzyop \in \{\Tnorm,\Snorm,\ImpOp,\Neg\}$. Definitions of some common fuzzy operators
are presented in \ref{a:operators}.  Let $\phi$ and $\psi$ be two
formulas with free variables $x_1,\dots,x_m$ and $y_1,\dots,y_n$,
respectively. Let us assume that the first $k$ variables are common to
$\phi$ and $\psi$. Recall that $\diamond$ and $\circ$ denote the set
of unary and binary connectives, respectively. 
Formally:

\begin{align}
  \label{eq:connective_semantics}
  \G(\diamond\phi)_{i_1,\dots,i_{m}} & =
   \fuzzyop(\diamond)(\G(\phi)_{i_1,\dots,i_m}) \\
   \label{eq:connective_semantics2}
  \G(\phi\circ\psi)_{i_1,\dots,i_{m+n-k}} & =
   \fuzzyop(\circ)(\G(\phi)_{i_1,\dots,i_k,i_{k+1},\dots,i_m}\G(\psi)_{i_1,\dots,i_k,i_{m+1},\dots,i_{m+n-k}})
\end{align}
In \eqref{eq:connective_semantics2}, $(i_1,\dots,i_k)$ denote the indices of the $k$ common variables, 
$(i_{k+1},\dots,i_{m})$ denote the indices of the $m-k$ variables appearing only in $\phi$,
and $(i_{m+1},\dots,i_{m+n-k})$ denote the indices of the $n-k$ variables appearing only in $\psi$.
Intuitively, $\G(\phi \circ \psi)$ is a tensor whose elements are obtained by applying 
$\fuzzyop(\circ)$ element-wise to every combination of individuals from $x_1,\dots,x_m$ and $y_1,\dots,y_n$ (see Figure \ref{f:illustration_ltn_connect}).

\begin{figure}
    \centering
    \includegraphics[width=\textwidth]{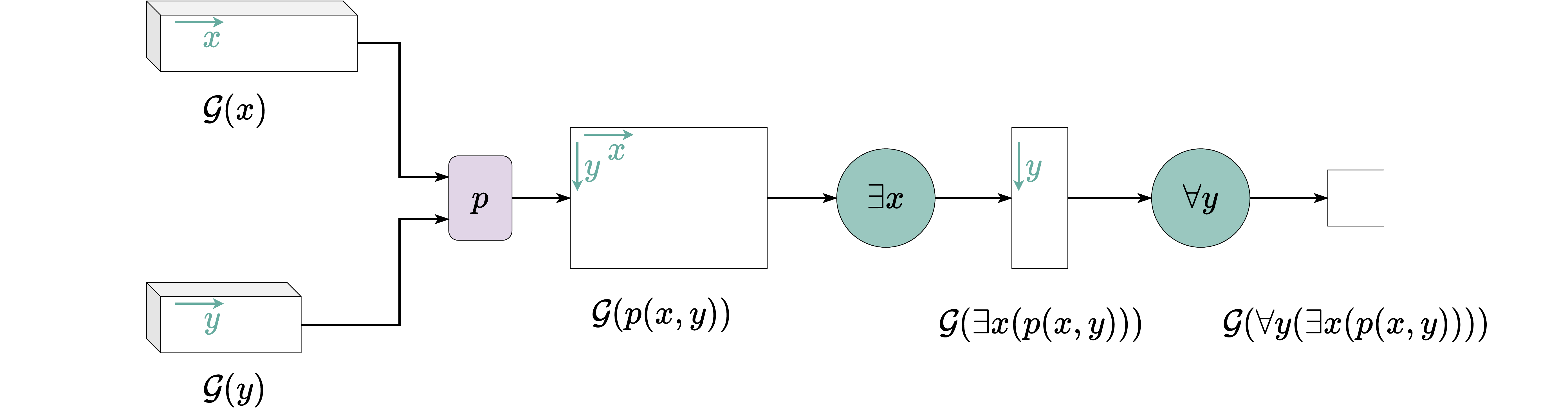}
    \caption{Illustration of an aggregation operation implementing quantification ($\forall y \exists x$) over variables $x$ and $y$.
    We assume that $x$ and $y$ have different domains.
    The result is a single number in the interval [0,1].}
    \label{f:illustration_ltn_aggreg}
\end{figure}

The semantics of the quantifiers ($\{\forall,\exists\}$) is defined with the use of aggregation. Let $\Agg$ be a symmetric and continuous aggregation operator, 
$\Agg:\bigcup\limits_{n \in \mathbb{N}} [0,1]^n \rightarrow [0,1]$. An analysis of suitable aggregation operators is presented in Appendix \ref{a:operators}.
For every formula $\phi$ containing $x_1,\dots,x_n$ free variables, suppose, without loss of generality, that quantification applies to the first $h$ variables.
We shall therefore apply $\Agg$ to the first $h$ axes of $\G(\phi)$, as follows: 
\begin{equation}
    \label{eq:aggreg_semantics}
    \G(Qx_1,\dots,x_h (\phi))_{i_{h+1},\ldots,i_{n}} =
\AggQ_{\substack{i_1=1,\dots,|\G(x_1)|\\\vdots\\ i_h=1,\dots,|\G(x_h)|}}\G(\phi)_{i_1,\dots,i_h,i_{h+1},\dots,i_n}
\end{equation}
where $\AggQ$ is the aggregation operator associated with the
quantifier $Q$. 
Intuitively, we obtain $\G(Qx_1,\dots,x_h (\phi))$ by reducing the dimensions associated with $x_1,\dots,x_h$ using the operator $\AggQ$ (see Figure \ref{f:illustration_ltn_aggreg}).

Notice that the above \emph{grounded} semantics can assign different meanings to the three formulas:
$$
\forall xy \big(\phi(x,y)\big) \ \ \ \ \
\forall x \big( \forall y \big(\phi(x,y)\big) \big) \ \ \ \ \
\forall y \big( \forall x \big(\phi(x,y)\big) \big)
$$
The semantics of the three formulas will coincide if the aggregation operator is bi-symmetric.

LTN also allows the following form of quantification, here called diagonal quantification ($\diag$): 
\begin{equation}
    \G(Q\ \diag(x_1,\dots,x_h) (\phi))_{i_{h+1},\ldots,i_{n}} =
\AggQ_{i=1,\dots,\min_{1 \leq j \leq h}|\G(x_j)|}
\G(\phi)_{i,\dots,i,i_{h+1},\dots,i_n}
\end{equation}

\begin{figure}
    \centering
    \includegraphics[width=\textwidth]{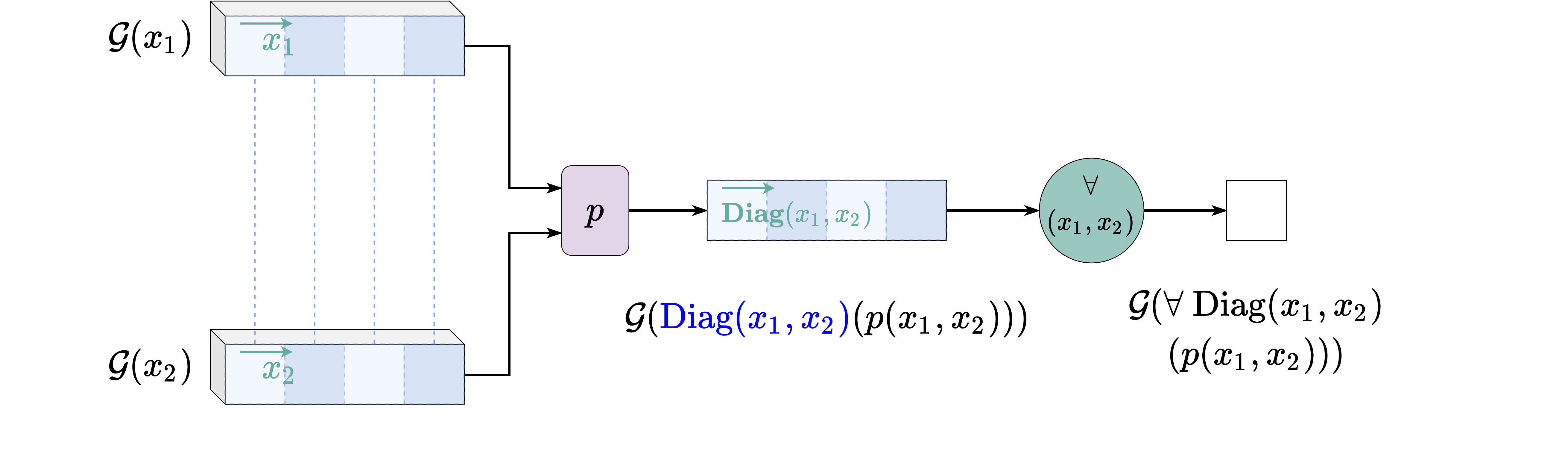}
    \caption{Diagonal Quantification: $\mathrm{Diag}(x_1,x_2)$ quantifies over specific tuples only, such that the $i$-th tuple contains the $i$-th instances of the variables $x_1$ and $x_2$ in the groundings $\G(x_1)$ and $\G(x_2)$, respectively.
    $\mathrm{Diag}(x_1,x_2)$ assumes, therefore, that $x_1$ and $x_2$ have the same number of instances as in the case of samples $x_1$ and their labels $x_2$ in a typical supervised learning tasks. 
    \label{f:illustration_ltn_diag}
    }
\end{figure}

$\diag (x_1,\dots,x_h)$ quantifies over specific tuples such that the $i$-th tuple contains the $i$-th instance of each of the variables in the argument of $\diag$, under the assumption that all variables in the argument are grounded onto sequences with the same number of instances. $\diag (x_1,\dots,x_h)$ is called diagonal quantification because it quantifies over the diagonal of $\G(\phi)$ along the axes associated with $x_1$...$x_h$, although in practice only the diagonal is built and not the entire $\G(\phi)$, as shown in Figure \ref{f:illustration_ltn_diag}. 
For example, given a data set with samples $x$ and target labels $y$, if looking to write a statement $p(x,y)$ that holds true for each pair of sample and label, one can write $\forall \mathrm{Diag}(x,y) \ p(x,y)$ given that $|\G(x)|=|\G(y)|$. 
As another example, given two variables $x$ and $y$ whose groundings contain 10 instances of $x$ and $y$ each, the expression $\forall \diag(x,y) \ p(x,y)$ produces 10 results such that the $i$-th result corresponds to the $i$-th instances of each grounding. 
Without $\diag$, the expression would be evaluated for all $10 \times 10$ combinations of the elements in $\G(x)$ and $\G(y)$.\footnote{Notice how $\diag$ is not simply "syntactic sugar" for creating a new variable $\mathrm{pairs\_xy}$ by stacking pairs of examples from $\G(x)$ and $\G(y)$. If the groundings of $x$ and $y$ have incompatible ranks (for instance, if $x$ denotes images and $y$ denotes their labels), stacking them in a tensor $\G(\mathrm{pairs\_xy})$ is non-trivial, requiring several reshaping operations.
} $\mathrm{Diag}$ will find much application in the examples and experiments to follow. 

\subsection{Guarded Quantifiers}
In many situations, one may wish to quantify over a set of elements of
a domain whose grounding satisfy some condition. In particular, one may wish to express such condition using formulas of the language of the form: 

\begin{align}
\label{eq:ex-masked-variable}
\forall y \ (\exists x : \mathrm{age}(x) > \mathrm{age}(y) \ ( \mathrm{parent}(x,y))) 
\end{align}

The grounding of such a formula is obtained by aggregating the
values of $\mathrm{parent}(x,y)$ only for the instances of $x$ that
satisfy the condition $\mathrm{age}(x)>\mathrm{age}(y)$, that is: 

$$
\All_{j=1,\dots,|\G(y)|}
\Exi_{\substack{i=1,\dots,|\G(x)| \ s.t. \\ \G(\mathrm{age}(x))_i > \G(\mathrm{age}(y))_j 
  }}\G(\mathrm{parent}(x,y))_{i,j}
$$

The evaluation of which tuple is safe is purely symbolic and non-differentiable.
Guarded quantifiers operate over only a subset of the variables, when this symbolic knowledge is crisp and available.
More generally, in what follows, $m$ is a symbol representing the condition, which we shall call a \emph{mask}, and $\G(m)$ associates a function
    \footnote{In some edge cases, a masking may produce an empty sequence, e.g. if for some value of $\G(y)$, there is no value in $\G(x)$ that satisfies $\mathrm{age}(x) > \mathrm{age}(y)$, we resort to the concept of an \emph{empty semantics}: $\forall$ returns $1$ and $\exists$ returns $0$.}
returning a Boolean to $m$.

\begin{equation}
  \G(Q\ x_1,\dots,x_h: m(x_1,\dots,x_n) (\phi))_{i_{h+1},\ldots,i_{n}} \stackrel{\text{def}}{=}
  \AggQ_{
    \substack{i_1=1,\dots,|\G(x_1)|\\
      \vdots\\
      i_h=1,\dots,|\G(x_h)| \ s.t.\\
      \G(m)(\G(x_1)_{i_1},\dots,\G(x_n)_{i_n} )
  }}\G(\phi)_{i_1,\dots,i_h,i_{h+1},\dots,i_n}
\end{equation}

Notice that the semantics of a guarded sentence $\forall x : m(x) (\phi(x))$ is different than 
the semantics of $\forall x (m(x) \imp \phi(x))$.
In crisp and traditional FOL, the two statements would be equivalent. 
In Real Logic, they can give different results.
Let $\G(x)$ be a sequence of 3 values, $\G(m(x)) = (0,1,1)$ and $\G(\phi(x)) = (0.2,0.7,0.8)$.
Only the second and third instances of $x$ are safe, that is, are in the masked subset.
Let $\imp$ be defined using the Reichenbach operator $\Ir(a,b) = 1-a+ab$ and $\forall$ be defined using the mean operator. 
We have $\G(\forall x (m(x) \imp \phi(x))) = \frac{1+0.7+0.8}{3} = 0.833\ldots$ 
whereas $\G(\forall x : m(x) (\phi(x))) = \frac{0.7+0.8}{2} = 0.75$.
Also, in the computational graph of the guarded sentence, there are no gradients attached to the instances that do not verify the mask.
Similarly, the semantics of $\exists x : m(x) (\phi(x))$ is not equivalent to that of $\exists x (m(x)\land \phi(x))$.

\begin{figure}
    \centering
    \includegraphics[width=\textwidth]{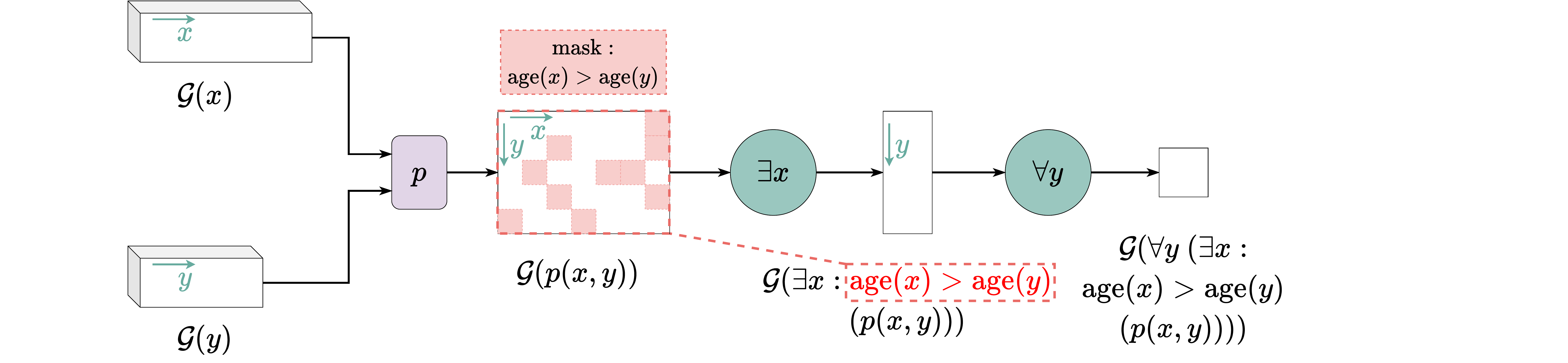}
    \caption{Example of Guarded Quantification: One can filter out elements of the various domains that do not satisfy some condition before the aggregation operators for $\forall$ and $\exists$ are applied.}
    \label{f:illustration_ltn_masking}
\end{figure}

\subsection{Stable Product Real Logic}
\label{s:stableproduct}
It has been shown in \cite{van_krieken_analyzing_2020} that not all first-order fuzzy logic semantics are equally suited for gradient-descent optimization. 
Many fuzzy logic operators can lead to 
vanishing or exploding gradients.
Some operators are also \emph{single-passing}, in that they propagate
gradients to only one input at a time.

In general, the best performing symmetric configuration\footnote{
        We define a symmetric configuration as a set of fuzzy operators such that conjunction and disjunction are defined by a t-norm and its dual t-conorm, respectively, and the implication operator is derived from such conjunction or disjunction operators and standard negation (c.f. \ref{a:operators} for details).
        In \cite{van_krieken_analyzing_2020}, van Krieken et al. also analyze non-symmetric configurations 
        and even operators that do not strictly verify fuzzy logic semantics.}
for the connectives uses the product t-norm $\Tprod$ for conjunction, its dual t-conorm $\Sprod$ for disjunction, standard negation $\Ns$, and the Reichenbach implication 
$\Ir$ (the corresponding S-Implication to the above operators). This subset 
of Real Logic where the grounding of the connectives is restricted to the product configuration is called \emph{Product Real Logic} in \cite{van_krieken_analyzing_2020}. 
Given $a$ and $b$ two truth-values in $[0,1]$:
\begin{align}
    \lnot : \Ns(a) &= 1-a \\
    \land : \Tprod(a,b) &= ab \\
    \lor : \Sprod(a,b) &= a+b-ab \\
    \imp : \Ir(a,b) &= 1 -a +ab 
\end{align}

Appropriate aggregators for $\exists$ and $\forall$ are the generalized mean $\Apmean$ with $p \geq 1$ to approximate the existential quantification, and the generalized mean w.r.t. the error $\ApmeanError$ with $p \geq 1$ to approximate the universal quantification.
They can be understood as a smooth maximum and a smooth minimum, respectively.
Given $n$ truth-values $a_1$, \dots, $a_n$ all in $[0,1]$:
\begin{align}
    \exists : \Apmean(a_1,\dots,a_n) 
        &= \biggl( \frac{1}{n} \sum\limits_{i=1}^n a_i^p \biggr)^{\frac{1}{p}}
        \qquad p \geq 1 \\
    \forall : \ApmeanError(a_1,\dots,a_n) 
        &= 1 - \biggl( \frac{1}{n} \sum\limits_{i=1}^n (1-a_i)^p \biggr)^{\frac{1}{p}}
        \qquad p \geq 1
\end{align}

$\ApmeanError$ measures the power of the deviation of each value from the ground truth $1$. 
With $p=2$, it is equivalent to $1-\mathsf{RMSE}(\bm{a},\bm{1})$, where $\mathsf{RMSE}$ is the root-mean-square error, $\bm{a}$ is the vector of truth-values and $\bm{1}$ is a vector of $1$'s.

The intuition behind the choice of $p$ is that the higher that $p$ is, the more weight that $\Apmean$ (resp. $\ApmeanError$) will give to $\mathit{true}$ (resp. $\mathit{false}$) truth-values, converging to the $\max$ (resp. $\min$) operator. 
Therefore, the value of $p$ can be seen as a hyper-parameter as it offers flexibility to account for outliers in the data depending on the application.

Nevertheless, \emph{Product Real Logic} still has the following gradient problems:
$\Tprod(a,b)$ has vanishing gradients on the edge case $a=b=0$;
$\Sprod(a,b)$ has vanishing gradients on the edge case $a=b=1$;
$\Ir(a,b)$ has vanishing gradients on the edge case $a=0$,$b=1$;
$\Apmean(a_1,\dots,a_n)$ has exploding gradients when $\sum_i (a_i)^p$ tends to $0$;
$\ApmeanError(a_1,\dots,a_n)$ has exploding gradients when $\sum_i (1-a_i)^p$ tends to $0$ (see \ref{a:gradients} for details).

\def\notzero{\pi_{0}}
\def\notone{\pi_{1}}

To address these problems, we define the projections $\notzero$ and 
$\notone$ below with $\epsilon$ an arbitrarily small positive real number:
\begin{align}
    \notzero:[0,1]\rightarrow]0,1]:&\ a\rightarrow (1-\epsilon)a + \epsilon \\
    \notone:[0,1]\rightarrow[0,1[:&\ a\rightarrow (1-\epsilon)a
\end{align}

We then derive the following stable operators to produce what we call the 
\emph{Stable Product Real Logic} configuration:
\begin{align}
  \Ns'(a) & = \Ns(a) \\
  \Tprod'(a,b) & = \Tprod(\notzero(a),\notzero(b)) \\
  \Sprod'(a,b) & = \Sprod(\notone(a),\notone(b)) \\
  \Ir'(a,b) & = \Ir(\notzero(a),\notone(b)) \\
  \Apmean'(a_1,\dots,a_n) & = \Apmean(\notzero(a_1),\dots,\notzero(a_n)) \qquad p \geq 1\\
  \ApmeanError'(a_1,\dots,a_n) & = \ApmeanError(\notone(a_1),\dots,\notone(a_n)) \qquad p \geq 1
\end{align}

It is important noting that the conjunction operator in stable product semantics is not a T-norm \footnote{Recall
  that a T-norm is a function $T:[0,1]\times[0,1]\rightarrow[0,1]$
  satisfying commutativity, monotonicity, associativity and identity,
  that is, $T(a,1)=a$.}.  $\Tprod'(a,b)$ does not satisfy identity in
$[0,1[$ since for any $0\leq a < 1$,
$\Tprod'(a,1)=(1-\epsilon)a+\epsilon\neq a$, although $\epsilon$ can be chosen arbitrarily small. 
In the experimental evaluations reported in Section~\ref{s:examples}, we find that the adoption of the stable product semantics is an important practical step to improve the numerical stability of the learning system.

\section{Learning, Reasoning, and Querying in Real Logic}

\label{s:ltn}

In Real Logic, one can define the tasks of \emph{learning}, \emph{reasoning} and
\emph{query-answering}. Given a Real Logic theory that represents the knowledge of an
agent at a given time, \emph{learning} is the task of making generalizations from specific observations obtained from data.
This is often called inductive inference. \emph{Reasoning} 
is the task of deriving what knowledge follows from the facts which are currently known. \emph{Query answering} is the task of evaluating the truth
value of a certain logical expression (called a query), or finding the set of objects in the data that evaluate a certain expression to $\mathit{true}$. In what follows, we define and exemplify each of these tasks. To do so, we first need to specify which types of knowledge can be
represented in Real Logic. 

\subsection{Representing Knowledge with Real Logic}
In logic-based knowledge representation
systems, knowledge is represented by logical formulas whose intended
meanings are propositions about a domain of interest. 
The connection between the symbols occurring in the formulas 
and what holds in the domain
is not represented in the knowledge base and is left
implicit since it does not have any effect on the logic computations. In
Real Logic, by contrast, the connection between the symbols and the
domain is represented explicitly in the language by the grounding $\G$, which plays an important role in both learning and reasoning.
$\G$ is an integral part of the knowledge represented by Real Logic. 
A Real Logic knowledge base is therefore defined by the formulas of
the logical language and knowledge about
the domain in the form of groundings obtained from data. The following types of knowledge can be represented in Real Logic.

\subsubsection{Knowledge through symbol groundings}

\begin{description}
  \item[\emph{Boundaries for domain grounding.}] 
  These are constraints specifying that 
  the value of a certain logical expression must be within a certain
  range. For instance, one may specify that the domain $\dom$ must be
  interpreted in the $[0,1]$ hyper-cube or in the standard $n$-simplex,
  i.e. the set ${d_1,\dots,d_n}\in(\R^+)^n$ such that
  $\sum_id_i=1$. Other intuitive examples of range constraints include the
  elements of the domain ``colour'' grounded onto points in
  $[0,1]^3$ such that every element is associated with the triplet of values
  $(R,G,B)$ with $R,G,B\in[0,1]$, or the range of a function $age(x)$ as
  an integer between $0$ and $100$.

  \item[\emph{Explicit definition of grounding for symbols.}] 
  Knowledge can be more strictly incorporated by fixing the grounding of some symbols.
    If a constant $c$ denotes an object with known features $\bv_c\in\R^n$, we can fix its grounding $\G(c)=\bv_c$.
    Training data that consists in a set of $n$ data items such as $n$ images (or tuples known as training examples) can be specified
    in Real Logic by $n$ constants, e.g. $\mathit{img}_1,\mathit{img}_2,\dots,\mathit{img}_n$, and
    by their groundings, e.g. 
    $\G(\mathit{img}_1)=\includegraphics[width=1em]{figures/mnist_eight.png},\
    \G(\mathit{img}_2)=\includegraphics[width=1em]{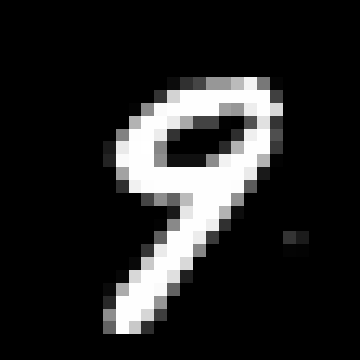},\
    \dots,\
    \G(\mathit{img}_n)=\includegraphics[width=1em]{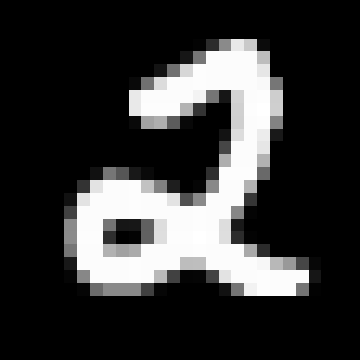}$.
    These can be gathered in a variable $\mathit{imgs}$.
    A binary predicate $\mathrm{sim}$ that measures the similarity of two objects
    can be grounded as, e.g., a cosine similarity function of two vectors $\bv$ and $\bw$,
      $(\bv,\bw)\mapsto \frac{\bv \cdot \bw}{||\bv|| \
        ||\bw||}$.
    The output layer of the neural network associated with a multi-class single-label predicate $P(x,\mathit{class})$
    can be a $\mathtt{softmax}$ function normalizing the output such that it guarantees exclusive classification, i.e. $\sum_{\mathit{i}}P(x,i)=1$.\footnote{Notice that $\mathtt{softmax}$ is often used as the last layer in neural networks to turn logits into a probability distribution. However, we do not use the $\mathtt{softmax}$ function as such here. Instead, we use it here to enforce an exclusivity constraint on satisfiability scores.}
  Grounding of constants and functions allows the computation of the grounding of their results. If, for example, $\G(\mathit{transp})$ is the function
  that transposes a matrix then
  $\G(\mathit{transp}(\mathit{img}_1)) =
  \includegraphics[width=1em]{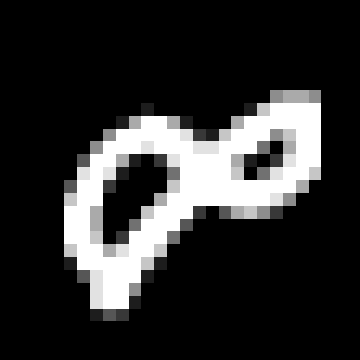}$.

  \item[\emph{Parametric definition of grounding for symbols.}]
  Here, the
  exact grounding of a symbol $\sigma$ is not known, but it is known
  that it can be obtained by finding a set of real-valued
  parameters, that is, via learning. 
  To emphasize this fact, we adopt the notation
  $\G(\sigma)=\G(\sigma\mid\bm\theta_\sigma)$ where
  $\bm\theta_\sigma$ is the set of parameter values that
  determines the value of $\G(\sigma)$. The typical example of 
  parametric grounding for constants is the learning of an embedding.  
  Let $\mathit{emb}(\mathit{word}\mid\bm\theta_{\mathit{emb}})$ be a word embedding with
  parameters $\bm\theta_{\mathit{emb}}$ which takes as input a word and
  returns its embedding in $\R^n$. If the words of a vocabulary
  $W=\{w_1,\dots,w_{|W|}\}$ are constant symbols, their groundings
  $\G(w_i\mid\bm\theta_{\mathit{emb}})$ are defined parametrically w.r.t.
  $\bm\theta_{\mathit{emb}}$ as $\mathit{emb}(w_i\mid\bm\theta_{\mathit{emb}})$. 
  An example of parametric grounding for a function symbol $f$
  is to assume that $\G(f)$ is a linear function such that $\G(f):\R^m\rightarrow\R^n$ 
  maps each $\bv\in\R^m$ into $\bm A_f\bv+\bm b_f$, with
  $\bm A_f$ a matrix of real numbers and $\bm b$ a vector of real numbers. In this case,
  $\G(f)=\G(f\mid\bm\theta_f)$, where $\bm\theta_f=\{\bm A_f,
  \bm b_f\}$. Finally, the grounding of a predicate symbol can be
  given, for example, by a neural network $N$ with parameters
  $\bm\theta_N$. As an example, consider a neural network
  $N$ trained for image classification into $n$ classes: $\mathit{cat},\mathit{dog},\mathit{horse},$ etc. 
  $N$ takes as input a vector $\bv$ of pixel values and produces as
  output a vector $\by=(y_{\mathit{cat}},y_{\mathit{dog}},y_{\mathit{horse}},\dots)$ in $[0,1]^n$ such that $\by=N(\bv\mid\bm\theta_N)$, where $y_c$ is the probability that input image $\bv$ is of
  class $c$. In case classes are, alternatively, chosen to be represented by 
  unary predicate symbols such as $\mathrm{cat(\bv)},\mathrm{dog(\bv)},\mathrm{horse(\bv)}$,\dots then
  $\G(\mathrm{cat(\bv)}) = N(\bv\mid\theta_N)_\mathit{cat}$,
  $\G(\mathrm{dog(\bv)}) = N(\bv\mid\theta_N)_\mathit{dog}$,
  $\G(\mathrm{horse(\bv)}) = N(\bv\mid\theta_N)_\mathit{horse}$, etc.

\end{description}

\subsubsection{Knowledge through formulas}

\begin{description}
  \item[\emph{Factual propositions.}] 
  Knowledge about the properties of specific objects in the domain is represented, as usual, by logical propositions, as exemplified below:
   Suppose that it is known that $\mathit{img}_1$ is a number eight, $\mathit{img}_2$ is a
  number nine, and $\mathit{img}_n$ is a number two. This can be represented by adding the following facts to the knowledge-base:
  $\mathrm{nine}(\mathit{img}_1),$ $\mathrm{eight}(\mathit{img}_2),\dots,$ $\mathrm{two}(\mathit{img}_n)$. Supervised learning, that is, learning with the use of training examples which include target values (labelled data), is specified in Real Logic by combining grounding
  definitions and factual propositions. For example, the fact that an image
  $\includegraphics[width=1em]{figures/mnist_nine.png}$ is a positive
  example for the class $\mathrm{nine}$ and a negative example for the class
  $\mathrm{eight}$ is specified by defining
  $\G(\mathit{img}_1)=\includegraphics[width=1em]{figures/mnist_nine.png}$
  alongside the propositions $\mathrm{nine}(\mathit{img}_1)$ and $\neg
  \mathrm{eight}(\mathit{img}_1)$. Notice how semi-supervision can be specified naturally in
  Real Logic by adding propositions containing disjunctions,
  e.g. $\mathrm{eight}(\mathit{img}_1)\lor \mathrm{nine}(\mathit{img}_1)$, which state that $\mathit{img}_1$ is either
  an eight or a nine (or both). Finally, relational learning can be achieved by relating logically multiple objects (defined as constants or variables or even as more complex sequences of terms) such as e.g.: $\mathrm{nine}(\mathit{img}_1)\rightarrow \lnot \mathrm{nine}(\mathit{img}_2)$ (if $\mathit{img}_1$ is a nine then $\mathit{img}_2$ is not a nine) or $\mathrm{nine}(\mathit{img})\rightarrow \lnot \mathrm{eight}(\mathit{img})$ (if an image is a nine then it is not an eight). The use of more complex knowledge including the use of variables such as $\mathit{img}$ above is the topic of \emph{generalized propositions}, discussed next. 
  
  \item[\emph{Generalized propositions.}]
  General knowledge about all or some of the objects of some domains can be specified in Real Logic by using first-order logic formulas with quantified variables. This general type of knowledge allows one to specify arbitrary constraints on the groundings independently from the specific data available. 
  It allows one to specify, in a concise way, knowledge that holds true for all the objects of a domain. 
  This is especially useful in Machine Learning in the semi-supervised and unsupervised settings, where
  there is no specific knowledge about a single individual. For
  example, as part of a task of multi-label classification with constraints on the labels \cite{chen2013improving}, a positive label constraint may express
   that if an example is labelled with $l_1,\dots,l_k$ then it should also be labelled with $l_{k+1}$. 
   This can be specified in Real Logic with a universally quantified formula:
  $\forall x\ (l_1(x)\wedge\dots\wedge l_k(x) \rightarrow l_{k+1}(x))$.\footnote{This can also be specified using a guarded quantifier
  $\forall x:( (l_1(x)\wedge\dots\wedge l_k(x))>\mathit{th} ) \ \  l_{k+1}(x)$ where $\mathit{th}$ is a threshold value in $[0,1]$.} 
  Another example of soft constraints used in Statistical Relational
  Learning associates the labels of related examples. For instance, in Markov Logic Networks \cite{MLN}, as part of the well-known \emph{Smokers and Friends} example, people who are smokers are associated by the friendship relation. In Real Logic, the formula
  $\forall x y \ ((\mathrm{smokes}(x) \land \mathrm{friend}(x,y)) \rightarrow \mathrm{smokes}(y))$
  would be used to encode the soft constraint that friends of smokers are normally smokers. 
\end{description}

\subsubsection{Knowledge through fuzzy semantics}
\label{s:knowledge_fuzzy_ops}
\begin{description}

  \item[\emph{Definition for operators.}]
  The grounding of a formula $\phi$ depends on the operators approximating the connectives and quantifiers that appear in $\phi$.
  Different operators give different interpretations of the satisfaction associated with the formula.
  For instance, the operator $\ApmeanError(a_1,\dots,a_n)$ that approximates universal quantification can be understood as a smooth minimum.
  It depends on a hyper-parameter $p$ (the exponent used in the generalized mean).
  If $p=1$ then $\ApmeanError(a_1,\dots,a_n)$ corresponds to the arithmetic mean. 
  As $p$ increases, given the same input, the value of the universally quantified formula will decrease as $\ApmeanError$ converges to the $\min$ operator. 
  To define how strictly the universal quantification should be interpreted in each proposition, one can use different values of $p$ for different propositions of the knowledge base.
  For instance, a formula $\forall x\ P(x)$ where $\ApmeanError$ is used with a low value for $p$ will in fact denote that $P$ holds for \emph{some} $x$, 
  whereas a formula $\forall x\ Q(x)$ with a higher $p$ may denote that $Q$ holds for \emph{most} $x$. 
 \end{description}

\subsubsection{Satisfiability}

 In summary, a Real Logic knowledge-base has three components:
 the first describes knowledge about the grounding of symbols (domains, constants, variables, functions, and predicate symbols); 
 the second is a set of closed logical formulas describing factual propositions and general knowledge;
 the third lies in the operators and the hyperparameters used to evaluate each formula. 
 The definition that follows formalizes this notion.

 \begin{definition}[Theory/Knowledge-base]
  A \emph{theory} of Real Logic is a triple
  $\T=\left<\K ,\G(\ \cdot\mid\bm\theta), \bm\Theta\right>$, where $\K$
  is a set of closed first-order logic formulas defined on the set of
  symbols $S=D\cup X \cup C\cup F\cup P$ denoting, respectively, domains,
  variables, constants, function and predicate symbols; 
  $\G(\ \cdot\mid\bm\theta)$
  is a parametric grounding for all the symbols $s\in S$ and all the logical operators; and 
  $\bm\Theta=\{\bm\Theta_s\}_{s\in S}$ is the hypothesis space for
  each set of parameters $\bm\theta_s$ associated with symbol $s$. 
\end{definition}

Learning and reasoning in a Real Logic theory are both associated with searching and applying the set of values of parameters $\bm\theta$ from the
hypothesis space $\bm\Theta$ that maximize the satisfaction of the
formulas in $\K$. We use the term \emph{grounded theory}, denoted by $\left<\K,\G_{\bm\theta}\right>$, to refer to a Real Logic theory
with a specific set of learned parameter values. This idea
shares some similarity with the weighted MAX-SAT problem
\cite{manquinho2009algorithms}, where the weights for formulas in $\K$ are given by their fuzzy truth-values obtained by choosing the parameter values of the grounding. To define this optimization problem, we aggregate the truth-values 
of all the formulas in $\K$ by selecting a \emph{formula
  aggregating} operator $\Sat:[0,1]^*\rightarrow [0,1]$.
\begin{definition}
  The \emph{satisfiability} of a theory 
  $\T=\left<\K,\G_{\bm\theta}\right>$ with respect to the \emph{aggregating operator}
  $\Sat$ is defined as $\Sat_{\phi\in\K}\G_{\bm\theta}(\phi)$. 
\end{definition}

\subsection{Learning} 
\label{s:learning}
Given a Real Logic theory $\T=(\K,\G(\ \cdot\mid\bm\theta),\bm\Theta)$,
\emph{learning} is the process of searching for the set of
parameter values $\bm\theta^\ast$ that maximize the satisfiability of $\T$ w.r.t. a given
aggregator: 
$$
    \bm\theta^\ast = \argmax_{\bm\theta\in\bm\Theta}\ \Sat_{\phi\in\K}\G_{\bm\theta}(\phi)
$$

Notice that with this general formulation, one can learn the grounding
of constants, functions, and predicates. The learning of the grounding
of constants corresponds to the learning of \emph{embeddings}. The learning of
the grounding of functions corresponds to the learning of \emph{generative} models or a \emph{regression} task. 
Finally, the learning of the grounding of
predicates corresponds to a \emph{classification} task in Machine Learning.

In some cases, it is useful to impose some regularization (as done customarily in ML) on the set of parameters $\bm\theta$, thus encoding a preference on the hypothesis space $\bm\Theta$, such as a preference for smaller parameter values. In this case, learning
is defined as follows: 
$$
    \bm\theta^\ast = \argmax_{\bm\theta\in\bm\Theta}\left(\Sat_{\phi\in\K}\G_{\bm\theta}(\phi) - \lambda \mathrm{R}(\bm\theta)\right)
$$
where $\lambda\in\R^+$ is the regularization parameter and $\mathrm{R}$ is a
regularization function, e.g. $L_1$ or $L_2$ regularization,
that is, $L_1(\bm\theta)=\sum_{\theta\in\bm\theta}|\theta|$ and 
$L_2(\bm\theta)=\sum_{\theta\in\bm\theta}\theta^2$.

LTN can generalize and extrapolate when querying formulas grounded with unseen data (for example, new individuals from a domain), using knowledge learned with previous groundings (for example, re-using a trained predicate). 
This is explained in Section \ref{s:querying}.

\subsection{Querying}
\label{s:querying}
Given a grounded theory $\T=(\K,\G_{\bm\theta})$, \emph{query answering} allows one to check if a certain fact is true (or, more precisely, by how much it is
true since in Real Logic truth-values are real numbers in the interval [0,1]). There are various types of queries that can be asked of a grounded theory.

A first type of query is called \emph{truth queries}. Any formula in the language of $\T$ can be a truth query. The answer to a truth query $\phi_q$ is the truth
value of $\phi_q$ obtained by computing its grounding, i.e.
$\G_{\bm\theta}(\phi_q)$. Notice that, if $\phi_q$ is a closed
formula, the answer is a scalar in $[0,1]$ denoting the truth-value of $\phi_q$ according to $\G_{\bm\theta}$. if $\phi_q$ contains $n$ free variables $x_1,\dots,x_n$, 
the answer to the query is a tensor of order $n$ such that the component indexed by $i_1\dots i_n$ is the truth-value of $\phi_q$
evaluated in $\G_{\bm\theta}(x_1)_{i_1},\dots,\G_{\bm\theta}(x_n)_{i_n}$.

The second type of query is called \emph{value queries}. Any term in the language of $\T$ can be a value query. The answer to a value query $t_q$ is a tensor of real numbers obtained by computing the grounding of the term, i.e.
$\G_{\bm\theta}(t_q)$. 
Analogously to truth queries, the answer to a
value query is a ``tensor of tensors'' if $t_q$ contains variables. 
Using value queries,
one can inspect how a constant or a term, more generally, is embedded in the manifold.

The third type of query is called \emph{generalization truth queries}. With generalization truth queries, we are interested in knowing the truth-values of formulas when
these are applied to a new (unseen) set of objects of a domain, such as a validation or a test set of examples typically used in the evaluation of machine learning systems. A generalization truth query is a pair $(\phi_q(x),\bm U)$, where $\phi_q$ is a
formula with a free variable $x$ and $\bm U=(\bu^{(1)},\dots,\bu^{(k)})$ is a set of unseen examples whose dimensions are compatible with those of the domain of $x$. 
The answer to the query $(\phi_q(x),\bm U)$ is $\G_{\theta}(\phi_q(x))$ for $x$ taking each value $\bu^{(i)}$, $1 \leq i \leq k$, in $\bm U$.
The result
of this query is therefore a vector of $|\bm U|$ truth-values corresponding to the evaluation of $\phi_q$ on  new data $\bu^{(1)},\dots,\bu^{(k)}$.

The fourth and final type of query is \emph{generalization value
  queries}. These are analogous to generalization truth queries with the difference that they evaluate a term $t_q(x)$, and not a formula, on new data $\bm U$. 
  The result, therefore, is a vector of $|\bm U|$ values corresponding to the evaluation of the trained model on a regression task using test data $\bm U$. 

\subsection{Reasoning} 
\label{s:def_reasoning}

\subsubsection{Logical consequence in Real Logic}
From a pure logic perspective, reasoning is the task of verifying if a
formula is a logical consequence of a set of formulas. This can be achieved semantically using model theory ($\models$) or syntactically via a proof theory ($\vdash$). 
To characterize reasoning in Real Logic, we adapt the notion of logical
consequence for fuzzy logic provided in
\cite{bvehounek2011introduction}: 
A formula $\phi$ is a fuzzy logical consequence of a finite set of formulas
$\Gamma$, in symbols 
$
\Gamma\models\phi
$
if for every fuzzy interpretation $f$, if all the formulas in $\Gamma$
are true (i.e. evaluate to 1) in $f$ then $\phi$ is true in $f$. In other words, every model of $\Gamma$ is a model of $\phi$.  
A direct application of this definition to Real Logic is not practical
since in most practical cases the level of satisfiability of a
grounded theory $(\K,\G_{\bm\theta})$ will not be equal to 1. We 
therefore define an interval $[q,1]$ with $\frac12 < q < 1$ and assume that a formula is true if its truth-value is in the interval $[q,1]$.
This leads to the following definition: 
\begin{definition}
  \label{def:rl-logical-consequence}
  A closed formula $\phi$ is a logical consequence of a knowledge-base
  $(\K,\G(\ \cdot\mid\bm\theta),\bm\Theta)$, in symbols 
  $(\K,\G(\ \cdot\mid\bm\theta),\bm\Theta)\models_q\phi$, 
  if, for every grounded theory $\left<\K,\G_{\bm\theta}\right>$, if
  $\Sat(\K,\G_{\bm\theta})\geq q$ then $\G_{\bm\theta}(\phi)\geq q$. 
\end{definition}

\subsubsection{Reasoning by optimization}
Logical consequence by direct application of
Definition~\ref{def:rl-logical-consequence} requires querying the truth
value of $\phi$ for a potentially infinite set of groundings. 
Therefore, we consider in practice the following directions:

\paragraph{Reasoning Option 1 (Querying after learning)}
This is approximate logical inference by considering
only the grounded theories that maximally satisfy
$(\K,\G(\ \cdot\mid\bm\theta),\bm\Theta)$. 
We therefore define that $\phi$ is a \emph{brave logical
  consequence} of a Real Logic knowledge-base
$(\K,\G(\ \cdot\mid\bm\theta),\bm\Theta)$ if 
$\G_{\bm\theta^*}(\phi)\geq q$ for all the $\bm\theta^*$ such that: 
$$
\bm\theta^*=\argmax_{\bm\theta}\Sat(\K,\G_{\bm\theta})
\ \ \ \ \ \ \ \ \mbox{and}\ \ \ \ \ \ \ \  \Sat(\K,\G_{\bm\theta^*})\geq q
$$
The objective is to find all $\bm \theta^*$ that optimally satisfy the knowledge base and to measure if they also satisfy $\phi$.
One can search for such $\bm \theta^*$ by running multiple optimizations with the objective function of Section \ref{s:learning}. 

This approach is somewhat naive. 
Even if we run the optimization multiple times with multiple parameter initializations (to, hopefully, reach different optima in the search space),
the obtained groundings may not be representative of other optimal or close-to-optimal groundings.
In Section \ref{s:ex_reasoning}, we give an example that shows the limitations of this approach and motivates the next one.

\paragraph{Reasoning Option 2 (Proof by Refutation)} 
Here, we reason by refutation and search for a counter-example to the logical consequence by introducing an alternative search objective.
Normally, according to Definition \ref{def:rl-logical-consequence}, one tries to verify that:\footnote{
  For simplicity, we temporarily define the notation $\G(\K) := \Sat_{\phi \in \K}(\K,\G)$.
}
\begin{equation}
  \label{def:objective_reasoning_trad}
  \text{for all}\ \theta \in \Theta ,\ \text{if}\ \G_\theta(\K) \geq q \ \text{then}\ \G_\theta(\phi) \geq q.
\end{equation}

Instead, we solve the dual problem:
\begin{equation}
  \label{def:objective_reasoning_refut}
  \text{there exists}\ \theta \in \Theta\ \text{such that} \ \G_\theta(\K) \geq q \ \text{and} \ \G_\theta(\phi) < q.
\end{equation}
If Eq.\eqref{def:objective_reasoning_refut} is true then a counterexample to Eq.\eqref{def:objective_reasoning_trad} has been found and the logical consequence does not hold.
If Eq.\eqref{def:objective_reasoning_refut} is false then no counterexample to Eq.\eqref{def:objective_reasoning_trad} has been found and the logical consequence is assumed to hold true.
A search for such parameters $\theta$ (the counterexample) can be performed by minimizing $\G_\theta(\phi)$ while imposing a constraint that seeks to invalidate results where $\G_\theta(\K)<q$. We therefore define:
\begin{center}
$\mathrm{penalty}(\G_\theta,q)=\begin{cases}
  c \ \text{if}\ \G_\theta(\K) < q,\\
  0 \ \text{otherwise},
\end{cases}$ where $c>1$.\footnote{
  In the objective function, $\G^\ast$ should satisfy $\G^*(\K) \geq q$ before reducing $\G^*(\phi)$ because the penalty $c$ which is greater than 1 is higher than any potential reduction in $\G(\phi)$ which is smaller or equal to 1.
} 
\end{center}

\begin{figure}
\centering  
  \begin{tikzpicture}
  \begin{axis}
    [
      domain=-3:3,
      legend pos = north west,
      style={font=\scriptsize},
    ]
    \addplot[color=blue,mark=None,samples=100,domain=-3:3]
    {
      (x>=0) * x +
      (x<0) * 1 * (exp(x)-1)
    };
    \addlegendentry{$\alpha=1,\beta=1$}
    \addplot[color=red,mark=None,samples=100,domain=-3:3]
    {
      (x>=0) * 2 * x +
      (x<0) * 0.5 * (exp(x)-1)
    };
    \addlegendentry{$\alpha=0.5,\beta=2$}
  \end{axis}
  \end{tikzpicture}
\caption{$\mathtt{elu}(\alpha,\beta x)$ where $\alpha \geq 0$ and $\beta \geq 0$ are hyper-parameters. 
The function $\mathtt{elu}(\alpha,\beta (q-\G_\theta(\K)))$ with $\alpha$ low and $\beta$ high is a soft constraint for $\mathrm{penalty}(\G_\theta,q)$ suitable for learning.}
\label{f:elu}
\end{figure}
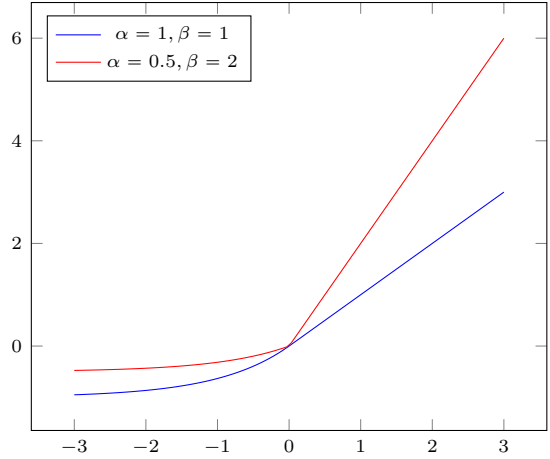

Given $\G^\ast$ such that:
\begin{equation}
  \label{eq:reasoning_refut_obj}  
  \G^\ast = \argmin_{\G_\theta} (\G_\theta(\phi) + \mathrm{penalty}(\G_\theta,q))
\end{equation}
\begin{itemize}
  \item If $\G^\ast(\K) < q$ : \ \  Then for all $\G_\theta$, $\G_\theta(\K) < q$ and therefore $(\K,\G(\ \cdot\mid\bm\theta),\bm\Theta)\models_q\phi$.
  \item If $\G^\ast(\K) \geq q \ \text{and}\ \G^\ast(\phi) \geq q$ : \ \ Then for all $\G_\theta$ with $\G_\theta(\K) \geq q$, we have that $\G_\theta(\phi) \geq \G^\ast(\phi) \geq q$ and therefore $(\K,\G(\ \cdot\mid\bm\theta),\bm\Theta)\models_q\phi$.  
  \item If $\G^\ast(\K) \geq q \ \text{and}\ \G^\ast(\phi) < q$ : \ \ Then $(\K,\G(\ \cdot\mid\bm\theta),\bm\Theta) \nvDash_q\phi$.
\end{itemize}
Clearly, Equation \eqref{eq:reasoning_refut_obj} cannot be used as an objective function for gradient-descent due to null derivatives.
Therefore, we propose to approximate the penalty function with the soft constraint:

\begin{center}
 $\mathtt{elu}(\alpha,\beta (q-\G_\theta(\K)))=\begin{cases}
  \beta  (q-\G_\theta(\K))\ &\text{if}\ \G_\theta(\K) \leq q,\\
  \alpha (e^{q-\G_\theta(\K)}-1) \ &\text{otherwise},
\end{cases}$ 
\end{center}
where $\alpha \geq 0$ and $\beta \geq 0$ are hyper-parameters (see Figure \ref{f:elu}).
When $\G_\theta(\K) < q$, the penalty is linear in $q-\G_\theta(\K)$ with a slope of $\beta$. 
Setting $\beta$ high, the gradients for $\G_\theta(\K)$ will be high in absolute value if the knowledge-base is not satisfied.
When $\G_\theta(\K) > q$, the penalty is a negative exponential that converges to $-\alpha$.
Setting $\alpha$ low but non-zero seeks to ensure that the gradients do not vanish when the penalty should not apply (when the knowledge-base is satisfied). We obtain the following approximate objective function:
\begin{equation}
  \G^\ast = \argmin_{\G_\theta} (\G_\theta(\phi) + \mathtt{elu}(\alpha,\beta (q-\G_\theta(\K)))
\end{equation}

Section \ref{s:ex_reasoning} will illustrate the use of reasoning by refutation with an example in comparison with reasoning as \emph{querying after learning}. Of course, other forms of reasoning are possible, not least that adopted in \cite{ReasoningLTN}, but a direct comparison is outside the scope of this paper and left as future work.

\section{The Reach of Logic Tensor Networks}
\label{s:examples}

The objective of this section is to show how the language of Real Logic can be used to specify a number of tasks that involve learning from data and reasoning. Examples of such tasks are classification, regression, clustering, and link prediction. 
The solution of a problem specified in Real Logic is obtained by
interpreting such a specification in \emph{Logic Tensor Networks}. The LTN library implements Real Logic in Tensorflow 2~\cite{tensorflow2015-whitepaper} and is available from GitHub\footnote{https://github.com/logictensornetworks/logictensornetworks}. Every logical operator is grounded using Tensorflow primitives such that LTN implements directly a Tensorflow graph. Due to Tensorflow built-in optimization, LTN is relatively efficient while providing the expressive power of first-order logic. Details on the implementation of the examples
  described in this section are reported in \ref{a:impl}. The implementation of the examples presented here is also available from the LTN repository on GitHub.
Except when stated otherwise, the results reported are the average result over 10 runs using a 95\% confidence interval.
Every example uses a stable real product configuration to approximate the Real Logic operators and the Adam optimizer~\cite{kingma_adam_2017} with a learning rate of $0.001$. Table \ref{tab:network_architectures} in the Appendix gives an overview of the network architectures used to obtain the results reported in this section.

\subsection{Binary Classification}
\label{s:ex_binary}
\begin{figure}
    \centering
    \includegraphics[width=1.\textwidth]{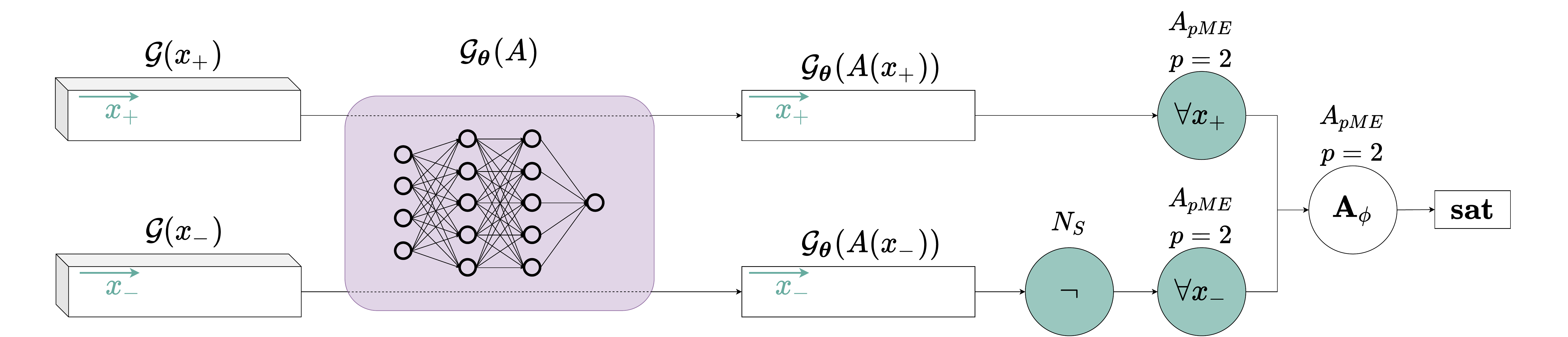}
    \caption{Symbolic Tensor Computational Graph for the Binary Classification Example. In the figure, $\G{x_+}$ and $\G{x_-}$ are inputs to the network $\G_\theta(A)$ and the dotted lines indicate the propagation of activation from each input through the network, which produces two outputs.}
    \label{f:graph_binary}
\end{figure}

\begin{figure}
\centering
\includegraphics[width=0.8\textwidth]{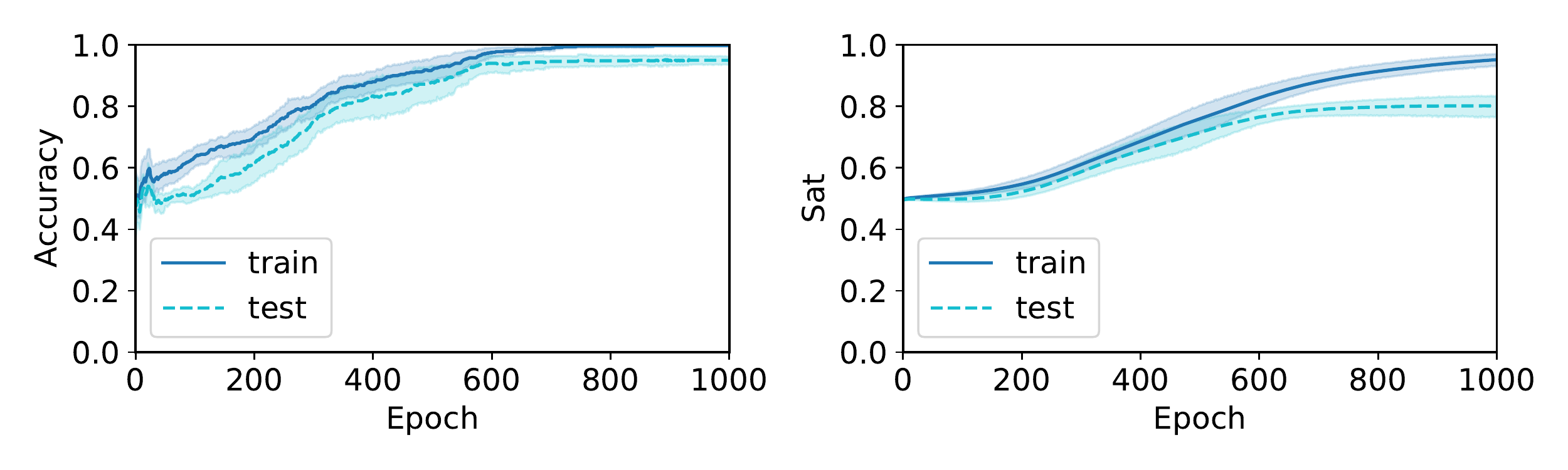}
\caption{Binary Classification task (training and test set performance): Average accuracy (left) and satisfiability (right). Due to the random initializations, accuracy and satisfiability start on average at $0.5$ with performance increasing rapidly after a few epochs.
}
\label{f:crv_binary_classification}
\end{figure}

The simplest machine learning task is binary classification.
Suppose that one wants to learn a binary classifier $A$ for a set of points in $[0,1]^2$.
Suppose that a set of positive and negative training examples is given.
LTN uses the following language and grounding:
\begin{description}
\item[Domains:] \ \\
    $\mathrm{points}$ (denoting the examples).
\item[Variables:] \ \\
    $x_+$ for the positive examples. \\
    $x_-$ for the negative examples. \\
    $x$ for all examples. \\
    $\domof(x) = \domof(x_+) = \domof(x_-) = \mathrm{points}$.
\item[Predicates:] \ \\
    $A(x)$ for the trainable classifier.\\
    $\domofin(A) = \mathrm{points}$.
\item[Axioms:]
\begin{align}
    \forall x_+ & \ A(x_+) \\
    \forall x_- & \ \neg A(x_-)
\end{align}

\item[Grounding:] \ \\
$\G(\mathrm{points}) = [0,1]^2$.\\
$\G(x) \in [0,1]^{m\times 2}$ ($\G(x)$ is a sequence of $m$ points, that is, $m$ examples).\\
$\G(x_+) = \left< d \in \G(x) \mid \lVert d - (0.5,0.5) \rVert < 0.09 \right>$.\footnote{$\G(x_+)$ are, by definition in this example, the training examples with Euclidean distance to the center $(0.5,0.5)$ smaller than the threshold of $0.09$.}\\
$\G(x_-) = \left< d \in \G(x) \mid \lVert d - (0.5,0.5) \rVert \geq 0.09 \right>$.\footnote{
 $\G(x_-)$ are, by definition, the training examples with Euclidean distance to the centre $(0.5,0.5)$ larger or equal to the threshold of $0.09$.}\\
$\G(A \mid \theta) : x \mapsto \mathtt{sigmoid}(\mathtt{MLP}_\theta(x))$, where $\mathtt{MLP}$ is a Multilayer Perceptron with a single output neuron, whose parameters $\theta$ are to be learned\footnote{
    $\mathtt{sigmoid}(x)=\frac{1}{1+e^-x}$
}.

\item[Learning:] \ \\
Let us define $\bm D$ the data set of all examples.
The objective function with $\K=\{\forall x_+ \ A(x_+), \forall x_- \ \neg A(x_-)\}$ is given by $\argmax_{\bm\theta\in\bm\Theta}\ \Sat_{\phi\in\K}\G_{\bm\theta, x \leftarrow \bm D}(\phi)$.
    \footnote{
        The notation $\G_{x \leftarrow \bm D}(\phi(x))$ means that the variable $x$ is grounded with the data $\bm D$ (that is, $\G(x) := \bm D$) when grounding $\phi(x)$. 
    }
In practice, the optimizer uses the following loss function:
$$
    \bm L = (1 - \Sat_{\phi\in\K}\G_{\bm\theta,x \leftarrow \bm B}(\phi))
$$
where $\bm B$ is a mini-batch sampled from $\bm D$.\footnote{
    As usual in ML, while it is possible to compute the loss function and gradients over the entire data set, it is preferred to use mini-batches of the examples.
}
The objective and loss functions depend on the following hyper-parameters:
\begin{itemize}
    \item the choice of fuzzy logic operator semantics used to approximate each connective and quantifier,
    \item the choice of hyper-parameters underlying the operators, such as the value of the exponent $p$ in any generalized mean,
    \item the choice of formula aggregator function.
\end{itemize}
Using the stable product configuration to approximate connectives and quantifiers, and $p=2$ for every occurrence of $\ApmeanError$, and using for the formula aggregator also $\ApmeanError$ with $p=2$, yields the following satisfaction equation:
\begin{align}
\Sat_{\phi \in \K}\G_\theta(\phi) 
    = & 1 - \frac{1}{2} \Big( 1 - \Big( 1 - \Big( \frac{1}{\lvert \G(x_+)\rvert} \sum\limits_{v \in \G(x_+)} \big(1-\mathtt{sigmoid}(\mathtt{MLP_\theta}(v))\big)^2 \Big)^{\frac{1}{2}\cdot 2}\Big) \notag \\
    & \qquad + 1 - \Big( 1 - \Big( \frac{1}{\lvert \G(x_-)\rvert} \sum\limits_{v \in \G(x_-)} \big(\mathtt{sigmoid}(\mathtt{MLP_\theta}(v))\big)^2 \Big)^{\frac{1}{2}\cdot 2}\Big)
    \Big)^{\frac{1}{2}} \notag 
\end{align}

The computational graph of Figure \ref{f:graph_binary} shows $\Sat_{\phi\in\K}\G_{\bm\theta}(\phi))$ as used with the above loss function.

We are therefore interested in learning the parameters $\theta$ of the $\mathtt{MLP}$ used to model the binary classifier.
We sample 100 data points uniformly from $[0,1]^2$ to populate the data set of positive and negative examples. 
The data set was split into 50 data points for training and 50 points for testing. 
The training was carried out for a fixed number of 1000 epochs using backpropagation with the Adam optimizer \cite{kingma_adam_2017} with a batch size of $64$ examples.
Figure \ref{f:crv_binary_classification} shows the classification accuracy and satisfaction level of the LTN on both training and test sets averaged over 10 runs using a 95\% confidence interval.
The accuracy shown is the ratio of examples correctly classified, with an example deemed as being positive if the classifier outputs a value higher than $0.5$.

Notice that a model can reach an accuracy of $100\%$ while satisfaction of the knowledge base is yet not maximized. 
For example, if the threshold for an example to be deemed as positive is $0.7$, all examples may be classified correctly with a confidence score of $0.7$. In that case, while the accuracy is already maximized, the satisfaction of $\forall x_+ A(x_+)$ would still be $0.7$, and can still improve until the confidence for every sample reaches $1.0$. 
\end{description}

\begin{figure}
\centering
\includegraphics[width=.8\textwidth]{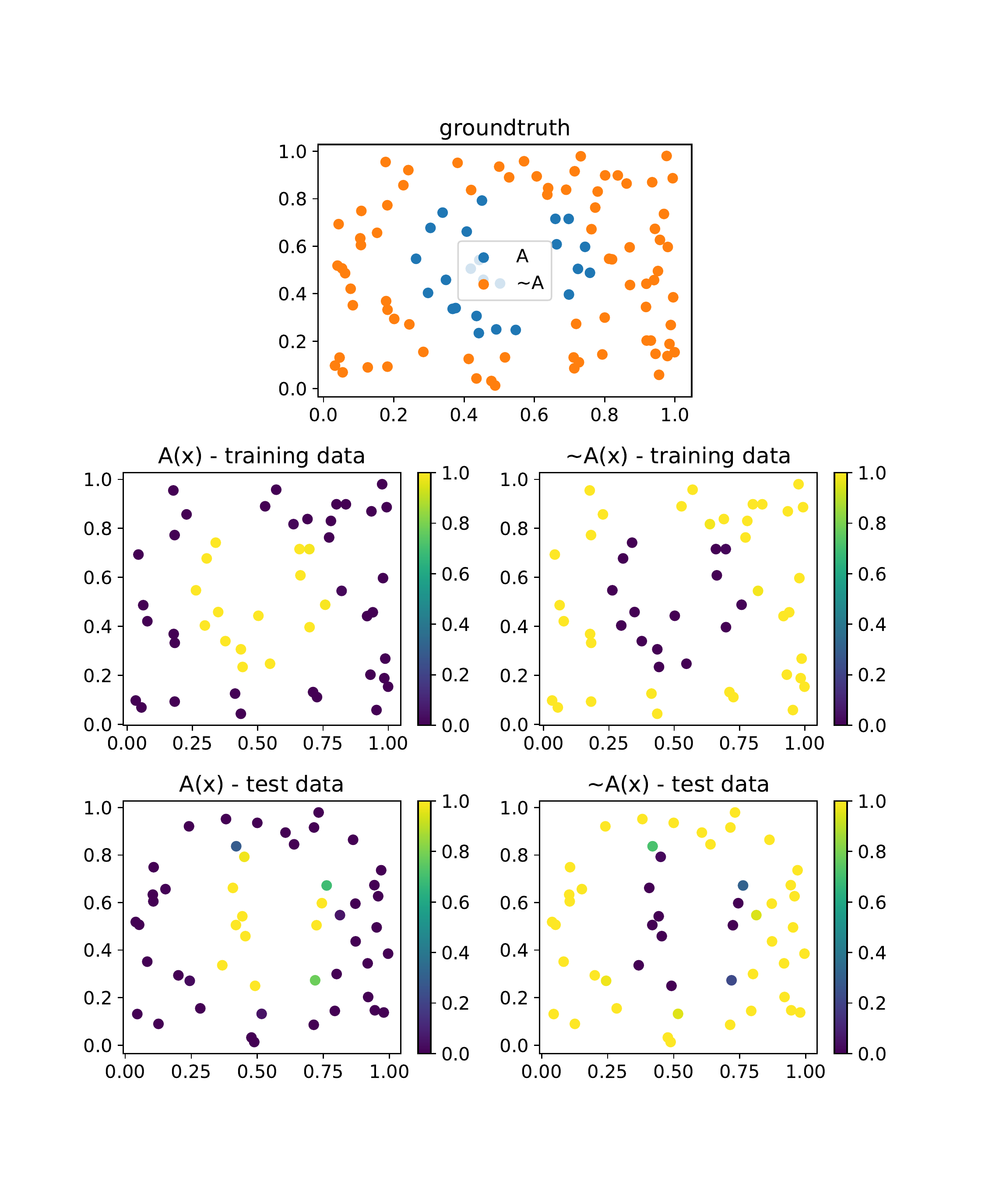}
\caption{Binary Classification task (querying the trained predicate $A(x)$): It is interesting to see how $A(x)$ could be appropriately named as denoting the inside of the central region shown in the figure, and therefore $\neg A(x)$ represents the outside of the region.}
\label{f:example_binary}
\end{figure}

This first example, although straightforward, illustrates step-by-step the process of using LTN in a simple setting. 
Notice that, according to the nomenclature of Section \ref{s:querying}, measuring accuracy amounts to querying the \emph{truth query} (respectively, the \emph{generalization truth query}) $A(x)$ for all the examples of the training set (respectively, test set) and comparing the results with the classification threshold. 
In Figure \ref{f:example_binary}, we show the results of such queries $A(x)$ after optimization.
Next, we show how the LTN language can be used to solve progressively more complex problems by combining learning and reasoning.

\subsection{Multi-Class Single-Label Classification} 
\label{example:multiclasssinglelabel}
The natural extension of binary classification is a multi-class classification task.
We first approach multi-class single-label classification, which assumes
that each example is assigned to one and only one label.

\begin{figure}
    \centering
    \includegraphics[width=1.\textwidth]{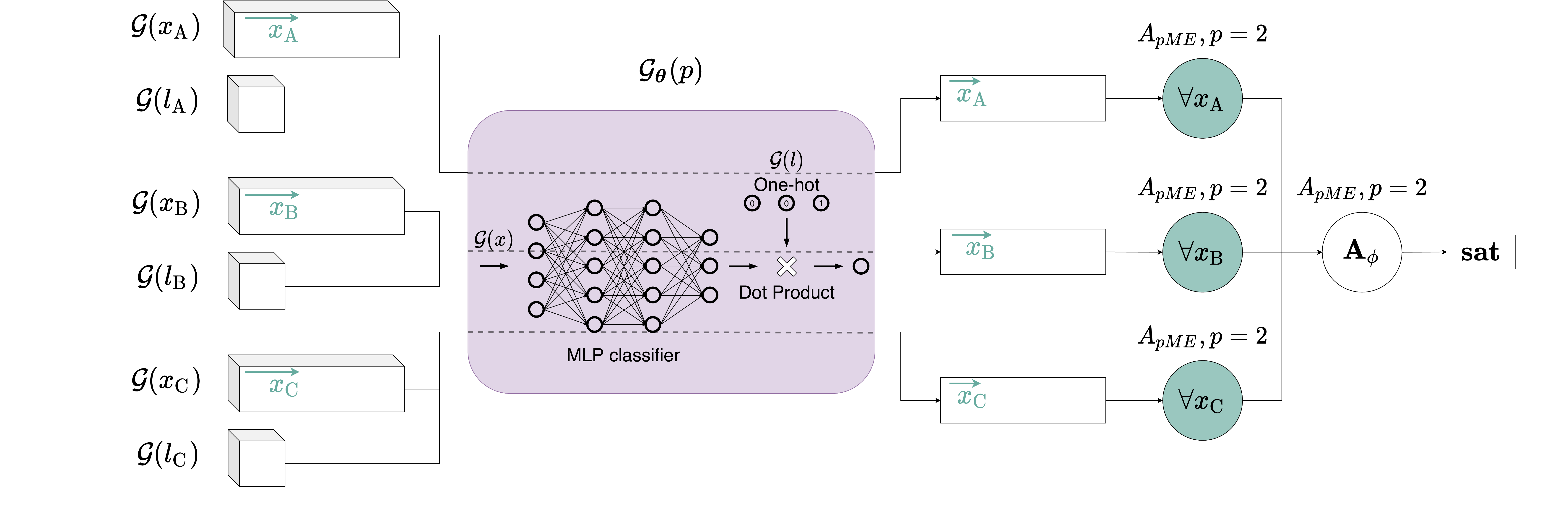}
    \caption{Symbolic Tensor Computational Graph for the Multi-Class Single-Label Problem. As before, the dotted lines in the figure indicate the propagation of activation from each input through the network, in this case producing three outputs.}
    \label{f:graph_multiclass}
\end{figure}

\begin{figure}
    \centering
    \includegraphics[width=0.8\textwidth]{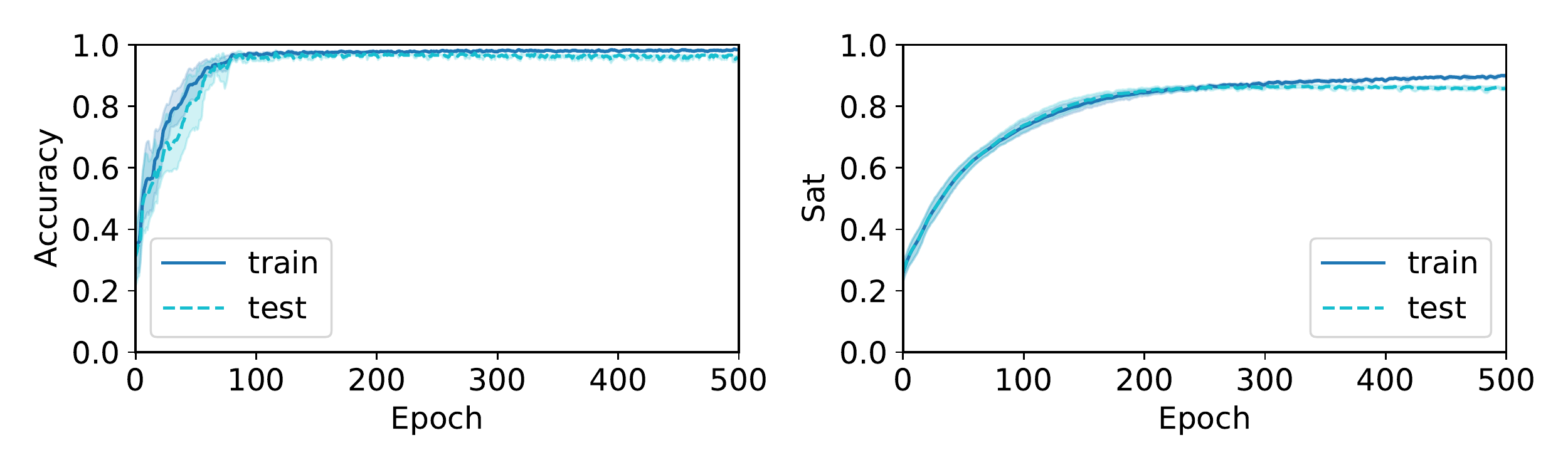}
    \caption{Multi-Class Single-Label Classification: Classification accuracy (left) and satisfaction level (right).}
    \label{f:crv_multiclass_singlelabel}
\end{figure}

For illustration purposes, we use the \emph{Iris flower} data set \cite{Dua:2019}, which consists of classification into three mutually exclusive classes; call these $A$, $B$, and $C$.
While one could train three unary predicates $A(x)$, $B(x)$ and $C(x)$, it turns out to be more effective if this problem is modeled by a single binary predicate $P(x,l)$, where $l$ is a variable denoting a multi-class label, in this case, classes $A$, $B$ or $C$.
This syntax allows one to write statements quantifying over the classes, e.g. $\forall x ( \exists l ( P(x,l)))$.
Since the classes are mutually exclusive, the output layer of the $\mathtt{MLP}$ representing $P(x,l)$ will be a $\mathtt{softmax}$ layer, instead of a $\mathtt{sigmoid}$ function, to ensure the exclusivity constraint on satisfiability scores.\footnote{$\texttt{softmax}(x)=e^{x_i}/\sum_j e^{x_j}$}. 
The problem can be specified as follows:

\begin{description}
    \item[Domains:] \ \\
        $\mathrm{items}$, denoting the examples from the Iris flower data set.\\
        $\mathrm{labels}$, denoting the class labels.
    \item[Variables:] \ \\
        $x_A$, $x_B$, $x_C$ for the positive examples of classes $A$, $B$, $C$. \\
        $x$ for all examples.\\
        $\domof(x_A) = \domof(x_B) = \domof(x_C) = \domof(x) = \mathrm{items}$.
    \item[Constants:] \ \\
        $l_A$, $l_B$, $l_C$, the labels of classes $A$ (Iris setosa), $B$ (Iris virginica), $C$ (Iris versicolor), 
        respectively.\\
        $\domof(l_A) = \domof(l_B) = \domof(l_C) = \mathrm{labels}$.
    \item[Predicates:] \ \\
        $P(x,l)$ denoting the fact that item $x$ is classified as $l$.\\
        $\domofin(P) = \mathrm{items},\mathrm{labels}$.
    \item[Axioms:] 
    \begin{align}
    \forall x_A &\ P(x_A,l_A)\\
    \forall x_B &\ P(x_B,l_B)\\
    \forall x_C &\ P(x_C,l_C)
    \end{align}
    Notice that rules about exclusiveness such as $\forall x (P(x,l_A) \rightarrow (\lnot P(x,l_B) \wedge \lnot P(x,l_C)))$ are not included since such constraints are already imposed by the grounding of $P$ below, more specifically the $\mathtt{softmax}$ function.
    \item[Grounding:] \ \\
    $\G(\mathrm{items}) = \R^4$, items are described by $4$ features: the length and the width of the sepals and petals, in centimeters.\\
    $\G(\mathrm{labels}) = \mathbb{N}^3$, we use a one-hot encoding to represent classes.\\
    $\G(x_A) \in \R^{m_1 \times 4}$, that is, $\G(x_A)$ is a sequence of $m_1$ examples of class $A$.\\
    $\G(x_B) \in \R^{m_2 \times 4}$, $\G(x_B)$ is a sequence of $m_2$ examples of class $B$.\\
    $\G(x_C) \in \R^{m_3 \times 4}$, $\G(x_C)$ is a sequence of $m_3$ examples of class $C$.\\
    $\G(x) \in \R^{(m_1+m_2+m_3)\times 4}$, $\G(x)$ is a sequence of all the examples.\\ 
    $\G(l_A) = [1,0,0]$, $\G(l_B) = [0,1,0]$, $\G(l_C) = [0,0,1]$.\\
    $\G(P \mid \theta) : x,l \mapsto l^\top \cdot \mathtt{softmax}(\mathtt{MLP}_\theta(x))$, where the $\mathtt{MLP}$ has three output neurons corresponding to as many classes, and 
    $\cdot$ denotes the dot product as a way of selecting an output for $\G(P \mid \theta)$; multiplying the $\mathtt{MLP}$'s output by the one-hot vector $l^\top$ gives the truth degree corresponding to the class denoted by $l$.
    \item[Learning:] \ \\
    The logical operators and connectives are approximated using the stable product configuration with $p=2$ for $\ApmeanError$.
    For the formula aggregator,  $\ApmeanError$ is used also with $p=2$.\\
    The computational graph of Figure \ref{f:graph_multiclass} illustrates how $\Sat_{\phi \in \K} \G_\theta(\phi)$ is obtained.
    If $\bm U$ denotes batches sampled from the data set of all examples, the loss function (to minimize) is:
    $$\bm L = 1-\Sat_{\phi \in \K }\G_{\theta,x\leftarrow \bm B}(\phi).$$
    Figure \ref{f:crv_multiclass_singlelabel} shows the result of training with the Adam optimizer with batches of 64 examples. 
    Accuracy measures the ratio of examples correctly classified, with example $x$ labeled as $\argmax_{l}(P(x,l))$.\footnote{This is also known as \emph{top-1} accuracy, as proposed in \cite{10.5555/2999134.2999257}. Cross-entropy results $\sum(t \ log(y))$ could have been reported here as is common with the use of softmax, although it is worth noting that, of course, the loss function used by LTN is different.}
    Classification accuracy reaches an average value near $1.0$ for both the training and test data after some 100 epochs. Satisfaction levels of the Iris flower predictions continue to increase for the rest of the training (500 epochs) to more than $0.8$.
    
    It is worth contrasting the choice of using a binary predicate $(P(x,l))$ in this example with the option of using multiple unary predicates $(l_A(x),l_B(x),l_C(x))$, one for each class. Notice how each predicate is normally associated with an output neuron. In the case of the unary predicates, the networks would be disjoint (or modular), whereas weight-sharing takes place with the use of the binary predicate. Since $l$ is instantiated into $l_A,l_B,l_C$, in practice $P(x,l)$ becomes $P(x,l_A)$, $P(x,l_B)$, $P(x,l_C)$, which is implemented via three output neurons to which a softmax function applies. 
    
\end{description}

\subsection{Multi-Class Multi-Label Classification}
\label{s:ex_multilabel}
We now turn to multi-label classification, whereby multiple labels can be assigned to each example. As a first example of the reach of LTNs, we shall see how the previous example can be extended naturally using LTN to account for multiple labels, not always a trivial extension for most ML algorithms.    
The standard approach to the multi-label problem is to provide explicit negative examples for each class. By contrast, LTN can use background knowledge to relate classes directly to each other, thus becoming a powerful tool in the case of the multi-label problem when typically the labeled data is scarce. 
We explore the \emph{Leptograpsus crabs} data set~\cite{campbell_multivariate_1974} consisting of 200 examples of 5 morphological measurements of 50 crabs. The task is to classify the crabs according to their color and sex. There are four labels: blue, orange, male, and female.
The color labels are mutually exclusive, and so are the labels for sex. LTN will be used to specify such information logically.

\begin{figure}
    \centering
    \includegraphics[width=1.\textwidth]{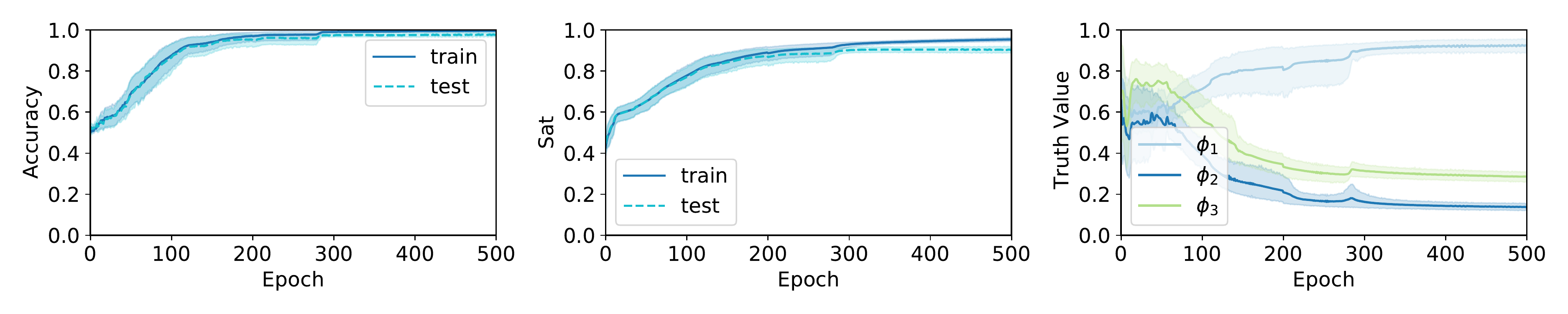}
    \caption{Multi-Class Multi-Label Classification: Classification Accuracy (left), Satisfiability level (middle), and Querying of Constraints (right).}
    \label{f:crv_multiclass_multilabel}
\end{figure}

\begin{description}
    \item[Domains:] \ \\
        $\mathrm{items}$ denoting the examples from the crabs dataset.\\
        $\mathrm{labels}$ denoting the class labels.
    \item[Variables:] \ \\
        $x_{\mathrm{blue}}$, $x_{\mathrm{orange}}$,  $x_{\mathrm{male}}$, $x_{\mathrm{female}}$ for the positive examples of each class.\\
        $x$, used to denote all the examples. \\
        $\domof(x_{\mathrm{blue}}) = \domof(x_{\mathrm{orange}}) = \domof(x_{\mathrm{male}}) = \domof(x_{\mathrm{female}}) = \domof(x) = \mathrm{items}$.
    \item[Constants:] \ \\
        $l_{\mathrm{blue}}$,  $l_{\mathrm{orange}}$, $l_{\mathrm{male}}$, $l_{\mathrm{female}}$ (the labels for each class).\\
        $\domof(l_{\mathrm{blue}})=\domof(l_{\mathrm{orange}})=\domof(l_{\mathrm{male}})=\domof(l_{\mathrm{female}})=\mathrm{labels}$.
    \item[Predicates:] \ \\
        $P(x,l)$, denotes the fact that item $x$ is labelled as $l$.\\
        $\domofin(P) = \mathrm{items},\mathrm{labels}$.
    \item[Axioms:] \ \\
        \begin{align}
        &\forall x_{\mathrm{blue}} \ P(x_{\mathrm{blue}},l_{\mathrm{blue}})\\
        &\forall x_{\mathrm{orange}} \ P(x_{\mathrm{orange}},l_{\mathrm{orange}})\\
        &\forall x_{\mathrm{male}} \ P(x_{\mathrm{male}},l_{\mathrm{male}})\\
        &\forall x_{\mathrm{female}} \ P(x_{\mathrm{female}},l_{\mathrm{female}})\\
        &\forall x \ \lnot(P(x,l_{\mathrm{blue}}) \land P(x,l_{\mathrm{orange}})) \label{ex:axiom:exclusivecolors}\\
        &\forall x \ \lnot(P(x,l_{\mathrm{male}}) \land P(x,l_{\mathrm{female}})) \label{ex:axiom:exclusivegenders}
        \end{align}
        Notice how logical rules \ref{ex:axiom:exclusivecolors} and \ref{ex:axiom:exclusivegenders} above represent the mutual exclusion of the labels on colour and sex, respectively. As a result, negative examples are not used explicitly in this specification.
    \item[Grounding:] \ \\
        $\G(\mathrm{items})=\R^5$; the examples from the data set are described using 5 features.\\
        $\G(\mathrm{labels})=\mathbb{N}^4$; one-hot vectors are used to represent class labels.\footnote{There are two possible approaches here: either each item is labeled with one multi-hot encoding or each item is labeled with several one-hot encodings. The latter approach was used in this example.}\\
        
        $\G(x_\mathrm{blue}) \in \R^{m_1 \times 5}$, $\G(x_\mathrm{orange}) \in \R^{m_2 \times 5}$, $\G(x_\mathrm{male}) \in \R^{m_3 \times 5}$, $\G(x_\mathrm{female}) \in \R^{m_4 \times 5}$.
        These sequences are not mutually-exclusive, one example can for instance be in both $x_\mathrm{blue}$ and $x_\mathrm{male}$.\\
        $\G(l_\mathrm{blue}) = [1,0,0,0]$, $\G(l_\mathrm{orange}) = [0,1,0,0]$,$\G(l_\mathrm{male}) = [0,0,1,0]$,$\G(l_\mathrm{female}) = [0,0,0,1]$.\\
        $\G(P \mid \theta) : x,l \mapsto l^\top \cdot \mathtt{sigmoid}(\mathtt{MLP}_\theta(x))$, with the $\mathtt{MLP}$ having four output neurons corresponding to as many classes. As before, $\cdot$ denotes the dot product which selects a single output.
        By contrast with the previous example, notice the use of a $\mathtt{sigmoid}$ function instead of a $\mathtt{softmax}$ function.
    \item[Learning:] \ \\
        As before, the fuzzy logic operators and connectives are approximated using the stable product configuration with $p=2$ for $\ApmeanError$, and for the formula aggregator, $\ApmeanError$ is also used with $p=2$.\\
        Figure \ref{f:crv_multiclass_multilabel} shows the result of the Adam optimizer using backpropagation trained with batches of 64 examples.
        This time, the accuracy is defined as $1-\mathrm{HL}$, where $\mathrm{HL}$ is the average Hamming loss, i.e. the fraction of labels predicted incorrectly, with a classification threshold of $0.5$ (given an example $u$, if the model outputs a value greater than $0.5$ for class $C$ then $u$ is deemed as belonging to class $C$). The rightmost graph in Figure \ref{f:crv_multiclass_multilabel} illustrates how LTN learns the constraint that a crab cannot have both blue and orange color, which is discussed in more detail in what follows.
    \item[Querying:] \ \\
        To illustrate the learning of constraints by LTN, we have queried three formulas that were not explicitly part of the knowledge-base, over time during learning:
        \begin{align}
            \phi_1 &:& \forall x \ &( P(x, l_{\mathrm{blue}}) \imp \lnot P(x, l_{\mathrm{orange}})) \\
            \phi_2 &:& \forall x \ &( P(x, l_{\mathrm{blue}}) \imp P(x, l_{\mathrm{orange}})) \\
            \phi_3 &:& \forall x \ &( P(x, l_{\mathrm{blue}}) \imp P(x, l_{\mathrm{male}})) 
        \end{align}
        For querying, we use $p=5$ when approximating the universal quantifiers with $\ApmeanError$.
        A higher $p$ denotes a stricter universal quantification with a stronger focus on outliers (see Section \ref{s:stableproduct}).
        \footnote{
            Training should usually not focus on outliers, as optimizers would struggle to generalize and tend to get stuck in local minima.
            However, when querying $\phi_1$,$\phi_2$,$\phi_3$, we wish to be more careful about the interpretation of our statement.
            See also \ref{s:knowledge_fuzzy_ops}. 
        }
        We should expect $\phi_1$ to hold true (every blue crab cannot be orange and vice-versa\footnote{Notice how, strictly speaking, other colours remain possible since the prior knowledge did not specify the bi-conditional: 
        $\forall x \ (P(x,l_{\mathrm{blue}}) \leftrightarrow \lnot P(x,l_{\mathrm{orange}}))$}, and we should expect $\phi_2$ (every blue crab is also orange) and $\phi_3$ (every blue crab is male) to be false. 
        The results are reported in the rightmost plot of Figure \ref{f:crv_multiclass_multilabel}. 
        Prior to training, the truth-values of $\phi_1$ to $\phi_3$ are non-informative. During training one can see, with the maximization of the satisfaction of the knowledge-base, a trend towards the satisfaction of $\phi_1$, and an opposite trend of $\phi_2$ and $\phi_3$ towards \emph{false}.
        
\end{description}

\subsection{Semi-Supervised Pattern recognition}
\label{s:ex_mnist}

Let us now explore two, more elaborate, classification tasks, which showcase the benefit of using logical reasoning alongside machine learning. With these two examples, we also aim to provide a more direct comparison with a related neurosymbolic system DeepProbLog \cite{DeepProbLog}. The benchmark examples below were introduced in the DeepProbLog paper \cite{DeepProbLog}.

\def\mnistthree{\inlinegraphics{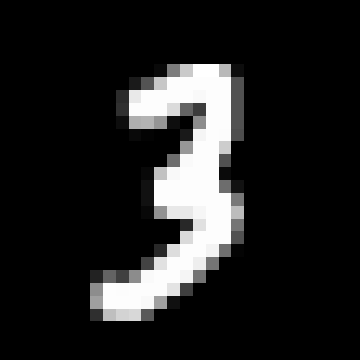}}
\def\mnisteight{\inlinegraphics{figures/mnist_eight.png}}
\def\mnistnine{\inlinegraphics{figures/mnist_nine.png}}
\def\mnisttwo{\inlinegraphics{figures/mnist_two.png}}

\begin{description}
    \item[Single Digits Addition:] 
    Consider the predicate $\mathtt{addition(X,Y,N)}$, 
    where $\mathtt{X}$ and $\mathtt{Y}$ are images of digits (the MNIST data set will be used), and 
    $\mathtt{N}$ is a natural number corresponding to the sum of these digits.
    This predicate should return an estimate of the validity of the addition.
    For instance, $\mathtt{addition(\mnistthree,\mnisteight,11)}$ is a valid addition;
    $\mathtt{addition(\mnistthree,\mnisteight,5)}$ is not.
    \item[Multi Digits Addition:]
    The experiment is extended to numbers with more than one digit. 
    Consider the predicate $\mathtt{addition([X_1,X_2],[Y_1,Y_2],N)}$.
    $\mathtt{[X_1,X_2]}$ and $\mathtt{[Y_1,Y_2]}$ are lists of images of digits, representing two multi-digit numbers; $\mathtt{N}$ is a natural number corresponding to the sum of the two multi-digit numbers.
    For instance, $\mathtt{addition([\mnistthree,\mnisteight],[\mnistnine,\mnisttwo],130)}$ is a valid addition; 
    $\mathtt{addition([\mnistthree,\mnisteight],[\mnistnine,\mnisttwo],26)}$ is not.    
\end{description}

A natural neurosymbolic approach is to seek to learn a single-digit classifier and benefit from knowledge readily available about the properties of addition in this case.
For instance, suppose that a predicate $\mathrm{digit}(x,d)$ gives the likelihood of an image $x$ being of digit $d$. 
A definition for $\mathtt{addition(\mnistthree,\mnisteight,11)}$ in LTN is:    
$$
\exists d_1,d_2 : d_1+d_2= 11 \ (\mathrm{digit}(\mnistthree,d_1)\land \mathrm{digit}(\mnisteight,d_2))
$$

In \cite{DeepProbLog}, the above task is made more complicated by not providing labels for the single-digit images during training. Instead, training takes place on pairs of images with labels made available for the result only, that is, the sum of the individual labels.
The single-digit classifier is not explicitly trained by itself; its output is a piece of latent information that is used by the logic.  
However, this does not pose a problem for end-to-end neurosymbolic systems such as LTN or DeepProbLog for which the gradients can propagate through the logical structures.

\begin{figure}
    \centering
    \includegraphics[width=1.\textwidth]{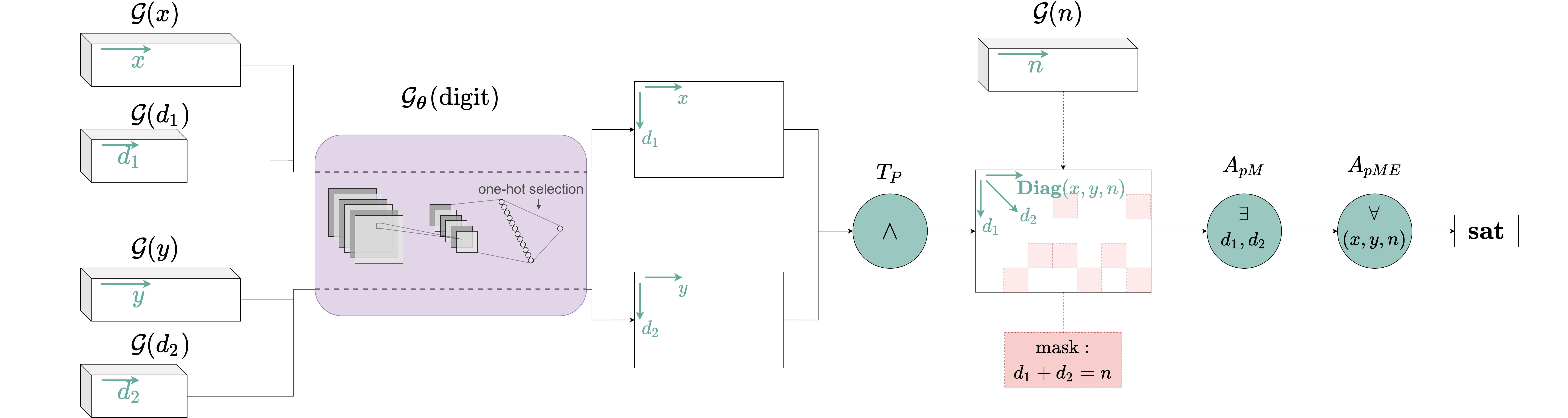}
    \caption{Symbolic Tensor Computational Graph for the Single Digit Addition task. Notice that the figure does not depict accurate dimensions for the tensors; $\G(x)$ and $\G(y)$ are in fact 4D tensors of dimensions $m \times 28 \times 28 \times 1$. Computing results with the variables $d_1$ or $d_2$ corresponds to the addition of a further axes of dimension $10$.}
    \label{f:graph_mnist}
\end{figure}

\begin{figure}
    \centering
    \includegraphics[width=1.\textwidth]{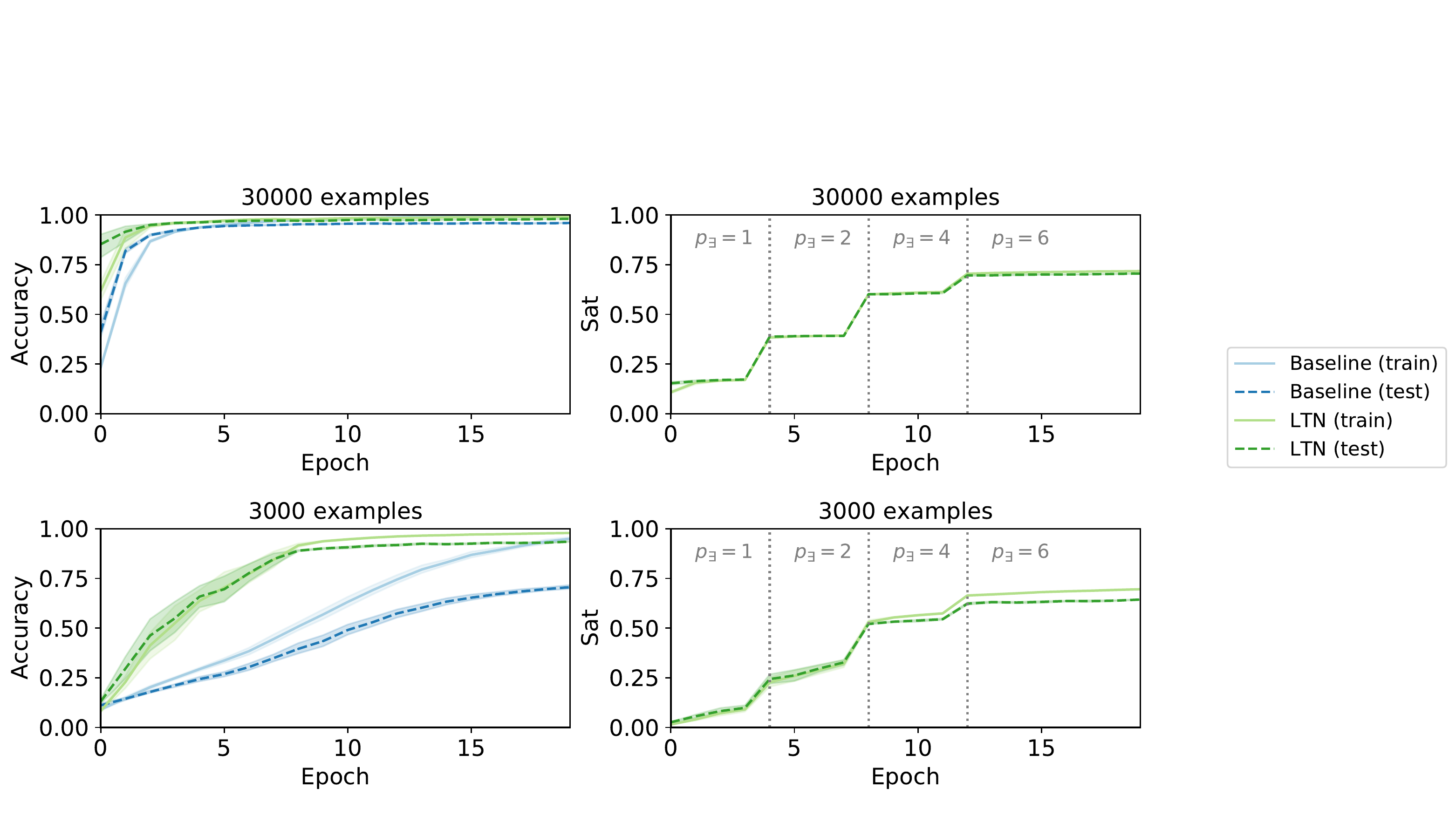}
    \caption{Single Digit Addition Task: Accuracy and satisfiability results (top) and results in the presence of fewer examples (bottom) in comparison with standard Deep Learning using a CNN (blue lines).}
    \label{f:crv_singledigits}
\end{figure}

\begin{figure}
    \centering
    \includegraphics[width=1.\textwidth]{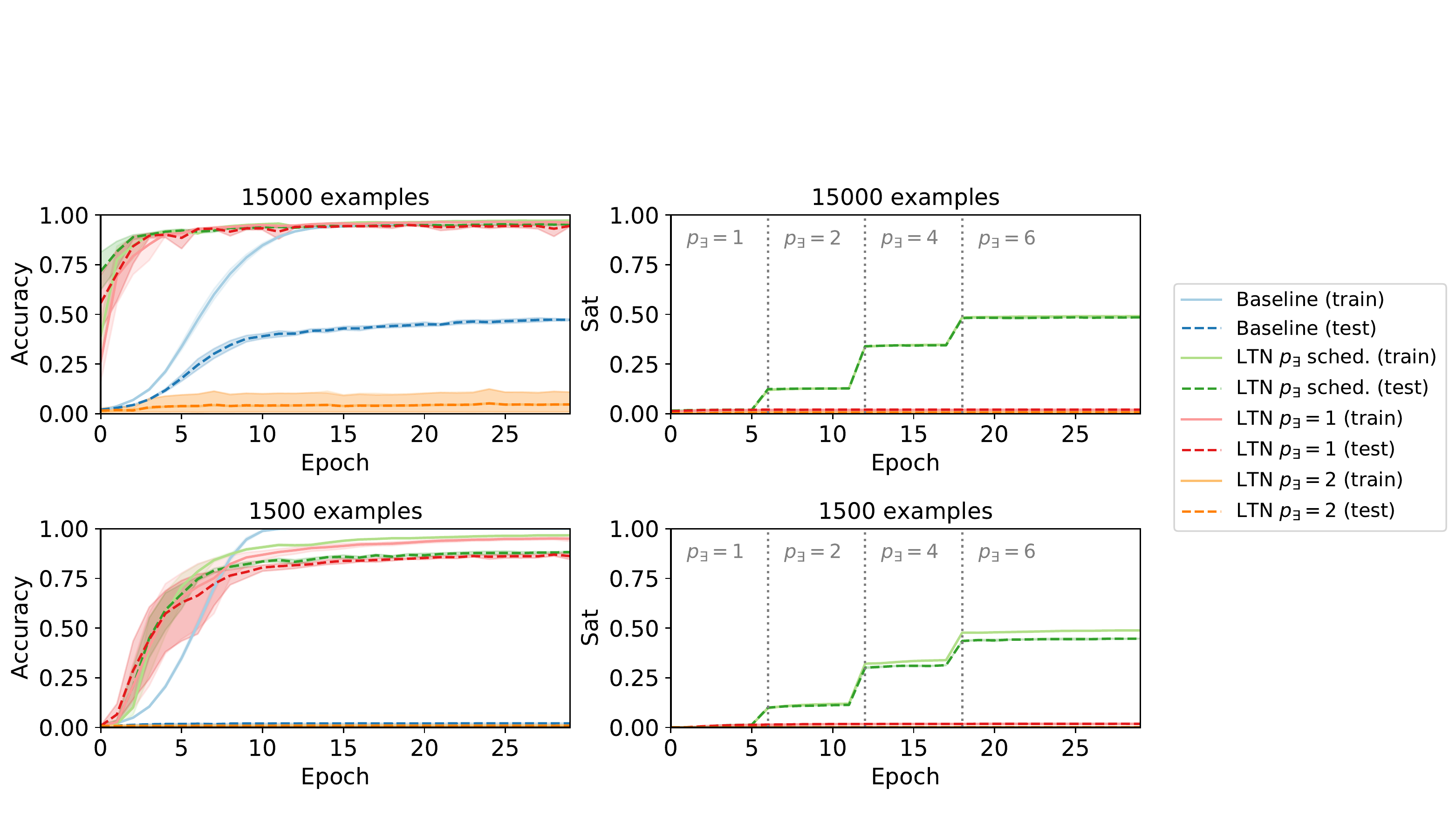}
    \caption{Multiple Digit Addition Task: Accuracy and satisfiability results (top) and results in the presence of fewer examples (bottom) in comparison with standard Deep Learning using a CNN (blue lines).}
    \label{f:crv_multidigits}
\end{figure}

We start by illustrating a LTN theory that can be used to learn the predicate $\mathrm{digit}$.
The specification of the theory below is for the single digit addition example, although it can be extended easily to the multiple digits case.
\begin{description}
    \item[Domains:] \ \\
    $\mathrm{images}$, denoting the MNIST digit images, \\
    $\mathrm{results}$, denoting the integers that label the results of the additions, \\
    $\mathrm{digits}$, denoting the digits from 0 to 9.
    \item[Variables:] \ \\
    $x$, $y$, ranging over the MNIST images in the data,\\
    $n$ for the labels, i.e. the result of each addition,\\
    $d_1$, $d_2$ ranging over digits.\\
    $\domof(x) = \domof(y) = \mathrm{images}$,\\
    $\domof(n) = \mathrm{results}$,\\
    $\domof(d_1) = \domof(d_2) = \mathrm{digits}$. 
    
    \item[Predicates:] \ \\
    $\mathrm{digit}(x,d)$ for the single digit classifier, where $d$ is a term denoting a digit constant or a digit variable. 
    The classifier should return the probability of an image $x$ being of digit $d$.\\
    $\domofin(\mathrm{digit}) = \mathrm{images},\mathrm{digits}$. 
    
    \item[Axioms:] \ \\
    Single Digit Addition:
        \begin{align}
            & \forall \mathrm{Diag}(x,y,n) \notag \\
            \label{eq:axiom_SD_addition}
            & \quad (\exists d_1,d_2 : d_1+d_2=n \\
            & \qquad (\mathrm{digit}(x,d_1) \land \mathrm{digit}(y,d_2))) \notag
        \end{align}
    Multiple Digit Addition:
        \begin{align}
            & \forall \mathrm{Diag}(x_1,x_2,y_1,y_2,n) \notag \\
            & \quad (\exists d_1,d_2,d_3,d_4 : 10d_1+d_2+10d_3+d_4=n \\
            & \qquad (\mathrm{digit}(x_1,d_1) \land \mathrm{digit}(x_2,d_2) \land \mathrm{digit}(y_1,d_3) \land \mathrm{digit}(y_2,d_4))) \notag
        \end{align}
    Notice the use of $\mathrm{Diag}$: 
    when grounding $x$,$y$,$n$ with three sequences of values, the $i$-th examples of each variable are matching.
    That is, $(\G(x)_i,\G(y)_i,\G(n)_i)$ is a tuple from our dataset of valid additions.
    Using the diagonal quantification, LTN aggregates pairs of images and their corresponding result, rather than any combination of images and results. 
    
    Notice also the guarded quantification: by quantifying only on the latent "digit labels" (i.e.\ $d_1$,$d_2$, \dots) that can add up to the result label ($n$, given in the dataset), we incorporate symbolic information into the system.
    For example, in \eqref{eq:axiom_SD_addition}, if $n=3$, the only valid tuples $(d_1,d_2)$ are $(0,3),(3,0),(1,2),(2,1)$. 
    Gradients will only backpropagate to these values.
    \item[Grounding:] \ \\
    $\G(\mathrm{images}) = [0,1]^{28\times 28\times 1}$.
    The MNIST data set has images of $28$ by $28$ pixels. 
    The images are grayscale and have just one channel.
    The RGB pixel values from $0$ to $255$ of the MNIST data set are converted to the range $[0,1]$.\\
    $\G(\mathrm{results}) = \mathbb{N}$.\\
    $\G(\mathrm{digits}) = \{0,1,\dots,9\}$.\\
    $\G(x) \in [0,1]^{m \times 28 \times 28 \times 1} $, $\G(y) \in [0,1]^{m \times 28 \times 28 \times 1} $, $\G(n) \in \mathbb{N}^m  $.\footnote{Notice the use of the same number $m$ of examples for each of these variables as they are supposed to match one-to-one due to the use of $\mathrm{Diag}$.}\\
    $\G(d_1) = \G(d_2) = \left<0,1,\dots,9\right>$.\\
    $\G(\mathrm{digit} \mid \theta) : x,d \mapsto \mathtt{onehot}(d)^\top \cdot \mathtt{softmax}(\mathtt{CNN}_\theta(x))$, 
    where $\mathtt{CNN}$ is a Convolutional Neural Network with $10$ output neurons for each class.
    Notice that, in contrast with the previous examples, $d$ is an integer label; 
    $\mathtt{onehot}(d)$ converts it into a one-hot label. 
    \item[Learning:] \ \\
    The computational graph of Figure \ref{f:graph_mnist} shows the objective function for the satisfiability of the knowledge base.
    A stable product configuration is used with hyper-parameter $p=2$ of the operator $\ApmeanError$ for universal quantification ($\forall$). 
    Let $p_\exists$ denote the exponent hyper-parameter used in the generalized mean $\Apmean$ for existential quantification ($\exists$). 
    Three scenarios are investigated and compared in the Multiple Digit experiment (Figure \ref{f:crv_multidigits}):
    \begin{enumerate}
        \item $p_\exists=1$ throughout the entire experiment,
        \item $p_\exists=2$ throughout the entire experiment, or
        \item $p_\exists$ follows a schedule, changing from $p=1$ to $p=6$ gradually with the number of training epochs.
    \end{enumerate}
    In the Single Digit experiment, only the last scenario above (schedule) is investigated (Figure \ref{f:crv_singledigits}).

    We train to maximize satisfiability by using batches of 32 examples of image pairs, labeled by the result of their addition.
    As done in \cite{DeepProbLog}, the experimental results vary the number of examples in the training set to emphasize the generalization abilities of a neurosymbolic approach. 
    Accuracy is measured by predicting the digit values using the predicate $\mathrm{digit}$ and reporting the ratio of examples for which the addition is correct.
    A comparison is made with the same baseline method used in \cite{DeepProbLog}:
    given a pair of MNIST images, a non-pre-trained CNN outputs embeddings for each image (Siamese neural network). 
    The embeddings are provided as input to dense layers that classify the addition into one of the 19 (respectively, 199) possible results of the Single Digit Addition (respectively, Multiple Digit Addition) experiments. 
    The baseline is trained using a cross-entropy loss between the labels and the predictions.
    As expected, such a standard deep learning approach struggles with the task without the provision of symbolic meaning about intermediate parts of the problem.
    
    Experimentally, we find that the optimizer for the neurosymbolic system gets stuck in a local optimum at the initialization in about 1 out of 5 runs.
    We, therefore, present the results on an average of the 10 best outcomes out of 15 runs of each algorithm (that is, for the baseline as well).
    The examples of digit pairs selected from the full MNIST data set are randomized at each run.

    Figure \ref{f:crv_multidigits} shows that the use of $p_\exists=2$ from the start produces poor results. 
    A higher value for $p_\exists$ in $\Apmean$ weighs up the instances with a higher truth-value (see also \ref{a:gradients} for a discussion).
    Starting already with a high value for $p_\exists$, the classes with a higher initial truth-value for a given example will have higher gradients and be prioritized for training, which does not make practical sense when randomly initializing the predicates.
    Increasing $p_\exists$ by following a schedule is the most promising approach.
    In this particular example, $p_\exists=1$ is also shown to be adequate purely from a learning perspective.
    However, $p_\exists=1$ implements a simple average which does not account for the meaning of $\exists$ well;
    the resulting satisfaction value is not meaningful within a reasoning perspective.
    
    Table \ref{table:cnn} shows that the training and test times of LTN are of the same order of magnitude as those of the CNN baselines. 
    Table \ref{table:acc_ltn_deepprob} shows that LTN reaches similar accuracy as that reported by DeepProbLog.

\end{description}

\begin{table}[h]
    \centering
    \begin{tabular}{
        p{2cm}
        >{\centering}m{2cm}
        >{\centering}m{2cm}
        >{\centering}m{2cm}
        >{\centering\arraybackslash}m{2cm} }
    \toprule
        & \multicolumn{2}{c}{(Single Digits)} & \multicolumn{2}{c}{(Multi Digits)} \\
        Model & Train & Test & Train & Test \\
    \midrule

        $\mathrm{baseline}$& $2.72 \pm 0.23 \mathrm{ms}$ &$1.45 \pm 0.21 \mathrm{ms}$ & $3.87 \pm 0.24 \mathrm{ms}$ &$2.10 \pm 0.30 \mathrm{ms}$ \\ 
        $\mathrm{LTN}$ & $5.36 \pm 0.25 \mathrm{ms}$& $3.44 \pm 0.39 \mathrm{ms}$ & $8.51 \pm 0.72 \mathrm{ms}$& $5.72 \pm 0.57 \mathrm{ms}$ \\
    \bottomrule
    \end{tabular}
    \caption{
        The computation time of training and test steps on the single and multiple digit addition tasks, 
        measured on a computer with a single Nvidia Tesla V100 GPU and averaged over 1000 steps.
        Each step operates on a batch of 32 examples.
        The computational efficiency of the LTN and the CNN baseline systems are of the same order of magnitude.
    }
    \label{table:cnn}
\end{table}

\begin{table}[h]
    \centering
    \begin{tabular}{
        p{2.5cm}
        >{\centering}m{1.75cm}
        >{\centering}m{1.75cm}
        >{\centering}m{1.75cm}
        >{\centering\arraybackslash}m{1.75cm} }
    \toprule
        & \multicolumn{4}{c}{Number of training examples} \\\cline{2-5}
        & \multicolumn{2}{c}{(Single Digits)} & \multicolumn{2}{c}{(Multi Digits)} \\
        Model & 30 000 & 3 000 & 15 000 & 1 500 \\
    \midrule
        $\mathrm{baseline}$ & $ 95.95 \pm 0.27 $ & $ 70.59 \pm 1.45 $ & $ 47.19 \pm 0.69 $ & $ 2.07 \pm 0.12 $ \\
        $\mathrm{LTN}$ & $ 98.04 \pm 0.13 $ & $ 93.49 \pm 0.28 $ & $ 95.37 \pm 0.29 $ & $ 88.21 \pm 0.63 $ \\
        $\mathrm{DeepProbLog}$ & $ 97.20 \pm 0.45 $ & $ 92.18 \pm 1.57 $ & $ 95.16 \pm 1.70 $ & $ 87.21 \pm 1.92 $ \\
    \bottomrule
    \end{tabular}
    \caption{Accuracy (in \%) on the test set: comparison of the final results obtained with LTN and those reported with DeepProbLog\cite{DeepProbLog}.
    Although it is difficult to compare directly the results over time (the frameworks are implemented in different libraries), while achieving similar computational efficiency as the CNN baseline, LTN also reaches similar accuracy as that reported by DeepProbLog.}
    \label{table:acc_ltn_deepprob}
\end{table}

\subsection{Regression}
\label{s:ex_regression}
Another important problem in Machine Learning is regression where a relationship is estimated between one independent variable $X$ and a continuous dependent variable $Y$. 
The essence of regression is, therefore, to approximate a function $f(x)=y$ by a function $f^*$, given examples $(x_i, y_i)$ such that $f(x_i) = y_i$. In LTN one can model a regression task by defining $f^*$ as a learnable function whose parameter values are constrained by data.
Additionally, a regression task requires a notion of equality. We, therefore, define the predicate $\mathrm{eq}$ as a smooth version of the symbol $=$ to turn the constraint $f(x_i)=y_i$ into a smooth optimization problem.

\begin{figure}
\centering
\includegraphics[width=0.7\textwidth]{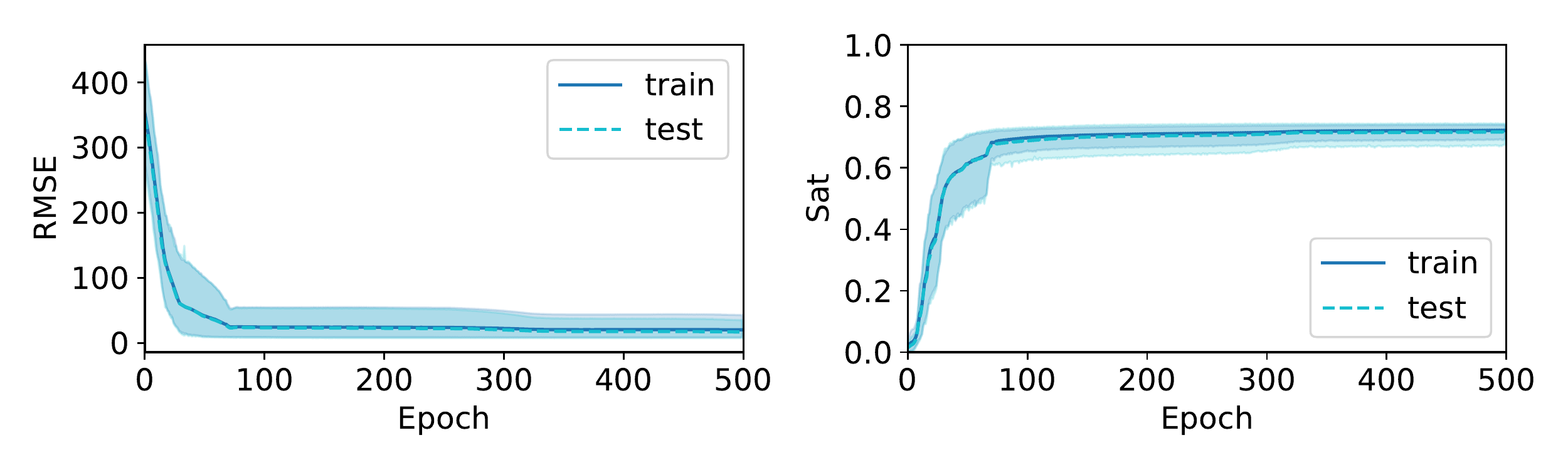}
\caption{Regression task: RMSE and satisfaction level over time.}
\label{f:crv_regression}
\end{figure}

\begin{figure}
\centering
\includegraphics[width=0.7\textwidth]{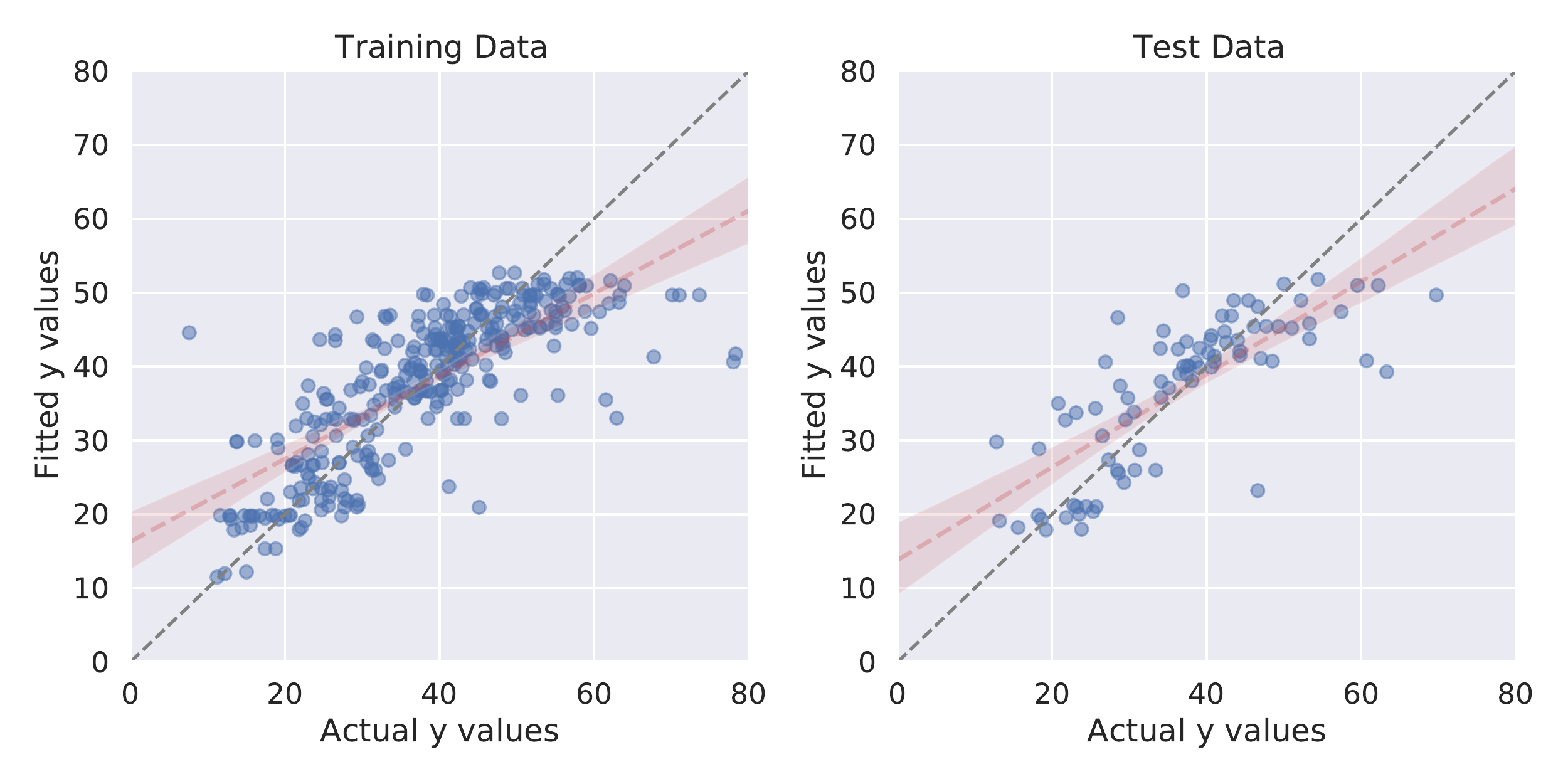}
\caption{Visualization of LTN solving a regression problem.}
\label{f:example_regression}
\end{figure}

In this example, we explore regression using a problem from a real estate data set\footnote{https://www.kaggle.com/quantbruce/real-estate-price-prediction} with
414 examples, each described in terms of 6 real-numbered features: 
the transaction date (converted to a float),
the age of the house, the distance to the nearest station, the number of convenience stores in the vicinity, and the latitude and longitude coordinates. The model has to predict the house price per unit area.
\begin{description}
\item[Domains:] \ \\
    $\mathrm{samples}$, denoting the houses and their features.\\
    $\mathrm{prices}$, denoting the house prices.
\item[Variables:] \ \\
    $x$ for the samples.\\
    $y$ for the prices.\\
    $\domof(x)=\mathrm{samples}$.\\
    $\domof(y)=\mathrm{prices}$.
\item[Functions:] \ \\
    $f^*(x)$, the regression function to be learned.\\
    $\domofin(f^*)=\mathrm{samples}$, $\domofout(f^*)=\mathrm{prices}$.
\item[Predicates:] \ \\
    $\mathrm{eq}(y_1,y_2)$, a smooth equality predicate that measures how similar $y_1$ and $y_2$ are.\\
    $\domofin(\mathrm{eq}) = \mathrm{prices},\mathrm{prices}$.
\item[Axioms:] 
\begin{align}
    \forall \mathrm{Diag}(x,y) \ \mathrm{eq}(f^*(x),y)
\end{align}
Notice again the use of $\mathrm{Diag}$: 
when grounding $x$ and $y$ onto sequences of values, this is done by obeying a one-to-one correspondence between the sequences. In other words, we aggregate pairs of corresponding $\mathrm{samples}$ and $\mathrm{prices}$, instead of any combination thereof. 
\item[Grounding:] \ \\
    $\G(\mathrm{samples})=\R^6$.\\
    $\G(\mathrm{prices})=\R$.\\
    $\G(x) \in \R^m \times 6$, $\G(y) \in \R^m \times 1$. 
    Notice that this specification refers to the same number $m$ of examples for $x$ and $y$ due to the above one-to-one correspondence obtained with the use of  $\mathrm{Diag}$.\\
    $\G(\mathrm{eq}(\mathbf{u},\mathbf{v})) = \exp\big(-\alpha\sqrt{\sum_j(u_j-v_j)^2}\big)$,
    where the hyper-parameter $\alpha$ is a real number that scales how strict the smooth equality is.
    \footnote{
        Intuitively, the smooth equality is $\exp(-\alpha \  d(\mathbf{u},\mathbf{v}))$, where $d(\mathbf{u},\mathbf{v})$ is the Euclidean distance between $\mathbf{u}$ and $\mathbf{v}$.
        It produces a $1$ if the distance is zero; 
        as the distance increases, the result decreases exponentially towards $0$. 
        In case an exponential decrease is undesirable, one can adopt the following alternative equation:  $\mathrm{eq}(\mathbf{u},\mathbf{v})=\frac{1}{1+\alpha d(\mathbf{u},\mathbf{v})}$.} 
        In our experiments, we use $\alpha=0.05$.\\
    $\G(f^*(x) \mid \theta) = \mathtt{MLP}_\theta(x)$, where $\mathtt{MLP}_\theta$ is a multilayer perceptron which ends in one neuron corresponding to a price prediction, with a linear output layer (no activation function).
\item[Learning:] \ \\    
    The theory is constrained by the parameters of the model of $f^*$. LTN is used to estimate such parameters by maximizing the satisfaction of the knowledge-base, in the usual way. 
    Approximating $\forall$ using $\ApmeanError$ with $p=2$, as before, we randomly split the data set into 330 examples for training and 84 examples for testing.
    Figure \ref{f:crv_regression} shows the satisfaction level over 500 epochs.
    We also plot the Root Mean Squared Error (RMSE) between the predicted prices and the labels (i.e. actual prices, also known as target values).
    We visualize in Figure \ref{f:example_regression} the strong correlation between actual and predicted prices at the end of one of the runs.
\end{description}

\subsection{Unsupervised Learning (Clustering)}
\label{s:clustering}
In unsupervised learning, labels are either not available or are not used for learning. Clustering is a form of unsupervised learning whereby, without labels, the data is characterized by constraints alone.
LTN can formulate such constraints, such as: 
\begin{itemize}
\item clusters should be disjoint,
\item every example should be assigned to a cluster,
\item a cluster should not be empty,
\item if the points are near, they should belong to the same cluster,
\item if the points are far, they should belong to different clusters, etc.
\end{itemize}

\def\closexy{\mathit{close\_xy}}
\def\distantxy{\mathit{distant\_xy}}

\begin{figure}
\begin{center}
\includegraphics[width=.7\textwidth]{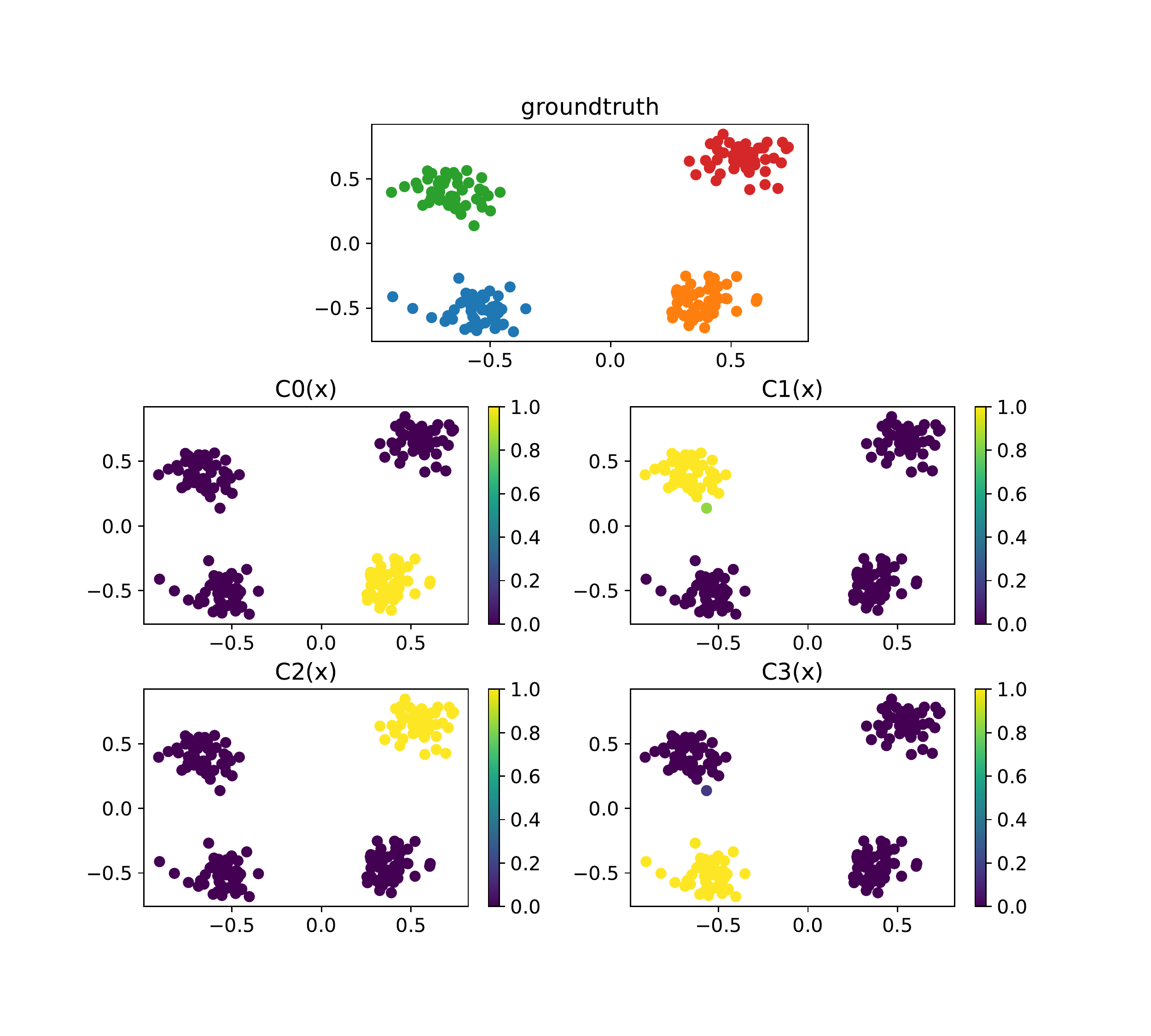}
\caption{LTN solving a clustering problem by constraint optimization: ground-truth (top) and querying of each cluster $C0$, $C1$, $C2$ and $C3$, in turn.}
\label{f:example_clustering}
\end{center}
\end{figure}

\begin{description}
\item[Domains:] \ \\
    $\mathrm{points}$, denoting the data to cluster.\\
    $\mathrm{points\_pairs}$, denoting pairs of examples.\\
    $\mathrm{clusters}$, denoting the cluster.
\item[Variables:] \ \\
    $x$, $y$ for all points. \\
    $\domof(x) = \domof(y) = \mathrm{points}$.\\
    $\domof(c) = \mathrm{clusters}$.
\item[Predicates:] \ \\
    $C(x,c)$, the truth degree of a given point belonging in a given cluster.\\
    $\domofin(C)=\mathrm{points},\mathrm{clusters}$.
\item[Axioms:]\ \\
\begin{align}
\forall x &\ \exists c \ C(x,c) \\
\forall c &\ \exists x \ C(x,c) \\
\label{eq:clustering_close} \forall (c, x, y : \lvert x - y \rvert < \mathrm{th_{close}} ) & \ \ (C(x,c) \leftrightarrow C(y,c)) \\
\label{eq:clustering_dist} \forall (c, x, y : \lvert x - y \rvert > \mathrm{th_{distant}}) & \ \ \lnot(C(x,c) \land C(y,c)) 
\end{align}

Notice the use of guarded quantifiers: all the pairs of points with Euclidean distance lower (resp. higher) than a value $\mathrm{th_{close}}$ (resp. $\mathrm{th_{distant}}$) should belong in the same cluster (resp. should not). 
$\mathrm{th_{close}}$ and $\mathrm{th_{distant}}$ are arbitrary threshold values that define some of the closest and most distant pairs of points. In our example, they are set to, respectively, $0.2$ and $1.0$.\\
As done in the example of Section \ref{example:multiclasssinglelabel}, the clustering predicate has mutually exclusive satisfiability scores for each cluster using a $\mathtt{softmax}$ layer.
Therefore, there is no explicit constraint about clusters being disjoint.

\item[Grounding:]\ \\
$\G(\mathrm{points}) = [-1,1]^2$.\\
$\G(\mathrm{clusters}) = \mathbb{N}^4$, we use one-hot vectors to represent a choice of 4 clusters.\\
$\G(x) \in [-1,1]^{m \times 2}$, that is, $x$ is a sequence of $m$ points. $\G(y) = \G(x)$.\\
$\mathrm{th_{close}} = 0.2$, $\mathrm{th_{distant}} = 1.0$.\\
$\G(c) = \left< [1,0,0,0],[0,1,0,0],[0,0,1,0],[0,0,0,1]\right>$.\\
$\G(C \mid \theta) : x,c \mapsto c^\top \cdot \mathtt{softmax}(\mathtt{MLP}_\theta(x))$, where $\mathtt{MLP}$ has 4 output neurons corresponding to the 4 clusters.

\item[Learning:]\ \\
We use the stable real product configuration to approximate the logical operators.
For $\forall$, we use $\ApmeanError$ with $p=4$. 
For $\exists$, we use $\Apmean$ with $p=1$ during the first 100 epochs, and $p=6$ thereafter, as a simplified version of the schedule used in Section \ref{s:ex_mnist}. 
The formula aggregator is approximated by $\ApmeanError$ with $p=2$.
The model is trained for a total of 1000 epochs using the Adam optimizer, which is sufficient for LTN to solve the clustering problem shown in Figure \ref{f:example_clustering}. Ground-truth data for this task was generated artificially by creating 4 centers, and generating $50$ random samples from a multivariate Gaussian distribution around each center.
The trained LTN achieves a satisfaction level of the clustering constraints of $0.857$.
\end{description}

\subsection{Learning Embeddings with LTN}
\label{s:ex_smokes}
A classic example of Statistical Relational Learning is the  smokers-friends-cancer example introduced in \cite{MLN}. Below, we show how this example can be formalized in LTN using semi-supervised embedding learning.

\begin{figure}
    \centering
    \includegraphics[width=0.8\textwidth]{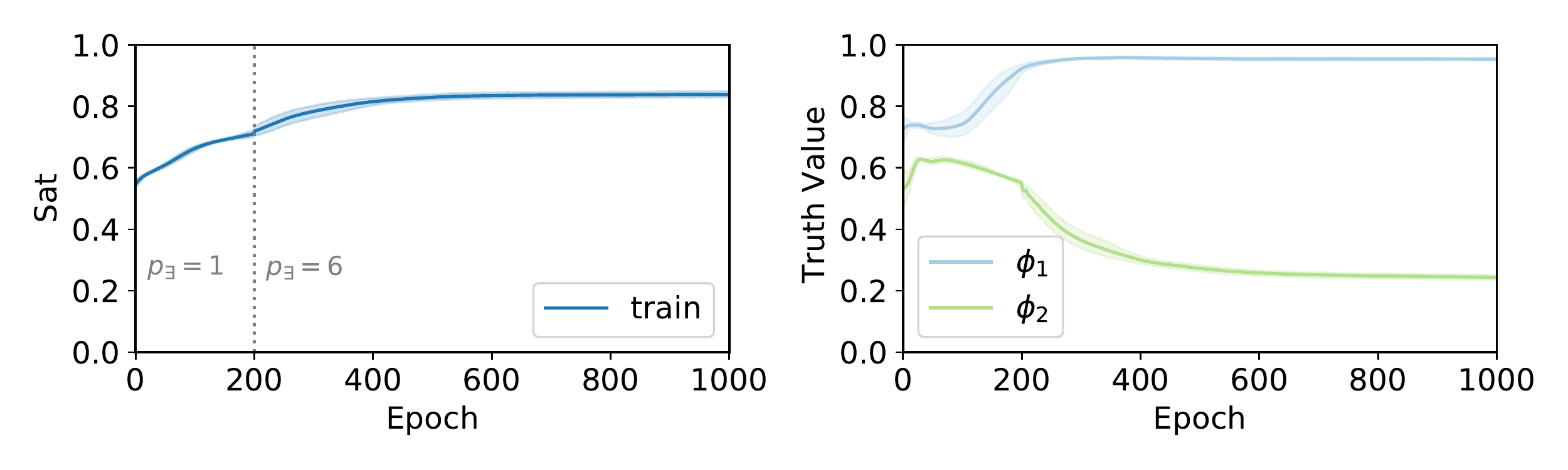}
    \caption{Smoker-Friends-Cancer example: Satisfiability levels during training (left) and truth-values of queries $\phi_1$ and $\phi_2$ over time (right).}
    \label{f:crv_smokes}
\end{figure}

\begin{figure}
    \centering
    \begin{subfigure}[t]{0.8\textwidth}
        \centering
        \includegraphics[width=.7\textwidth]{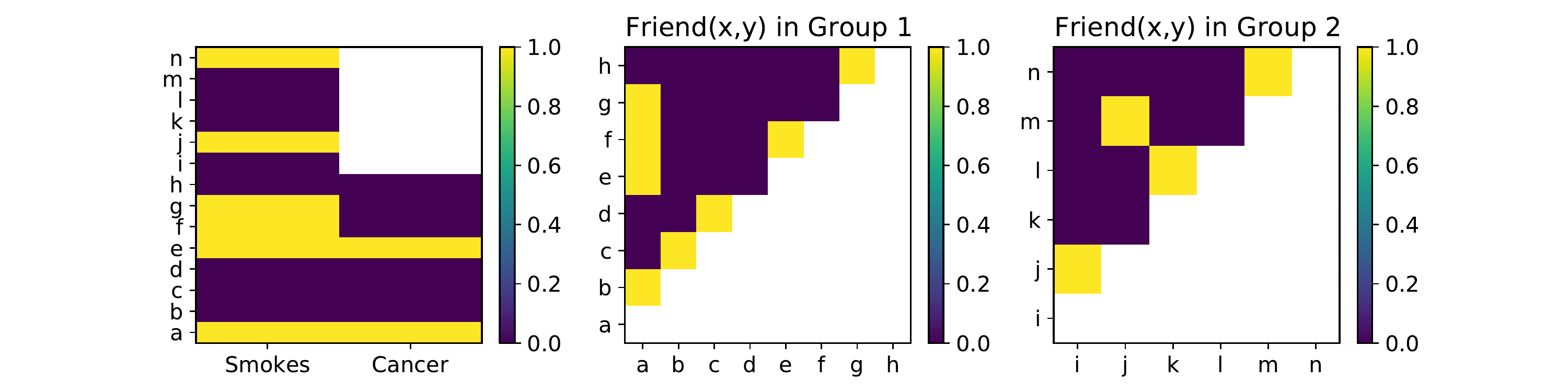}
        \caption{Incomplete facts in the knowledge-base: axioms for smokers and cancer for individuals $a$ to $n$ (left), friendship relations in group 1 (middle), and friendship relations in group 2 (right).}
        \label{f:smokes_incomplete_facts}
    \end{subfigure}
    ~\\
    \begin{subfigure}[t]{0.8\textwidth}
        \centering
        \includegraphics[width=.7\textwidth]{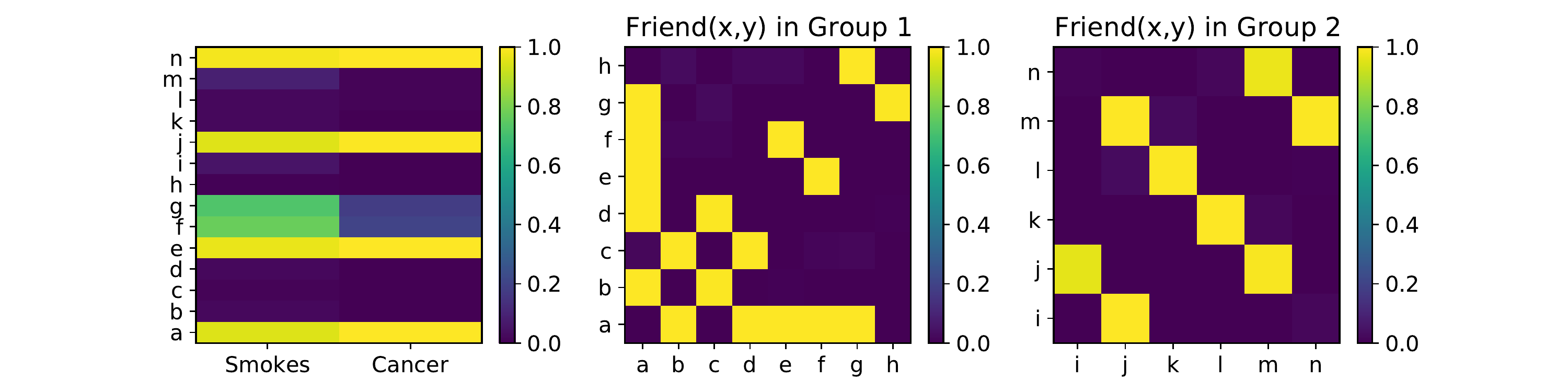}
        \caption{Querying all the truth-values using LTN after training: smokers and cancer (left), friendship relations (middle and right).}
        \label{f:smokes_inferred_facts}
    \end{subfigure}
    \caption{Smoker-Friends-Cancer example: Illustration of the facts before and after training.}
    \label{f:example_smokes_facts}
\end{figure}

\begin{figure}
    \centering
    \includegraphics[width=.7\textwidth]{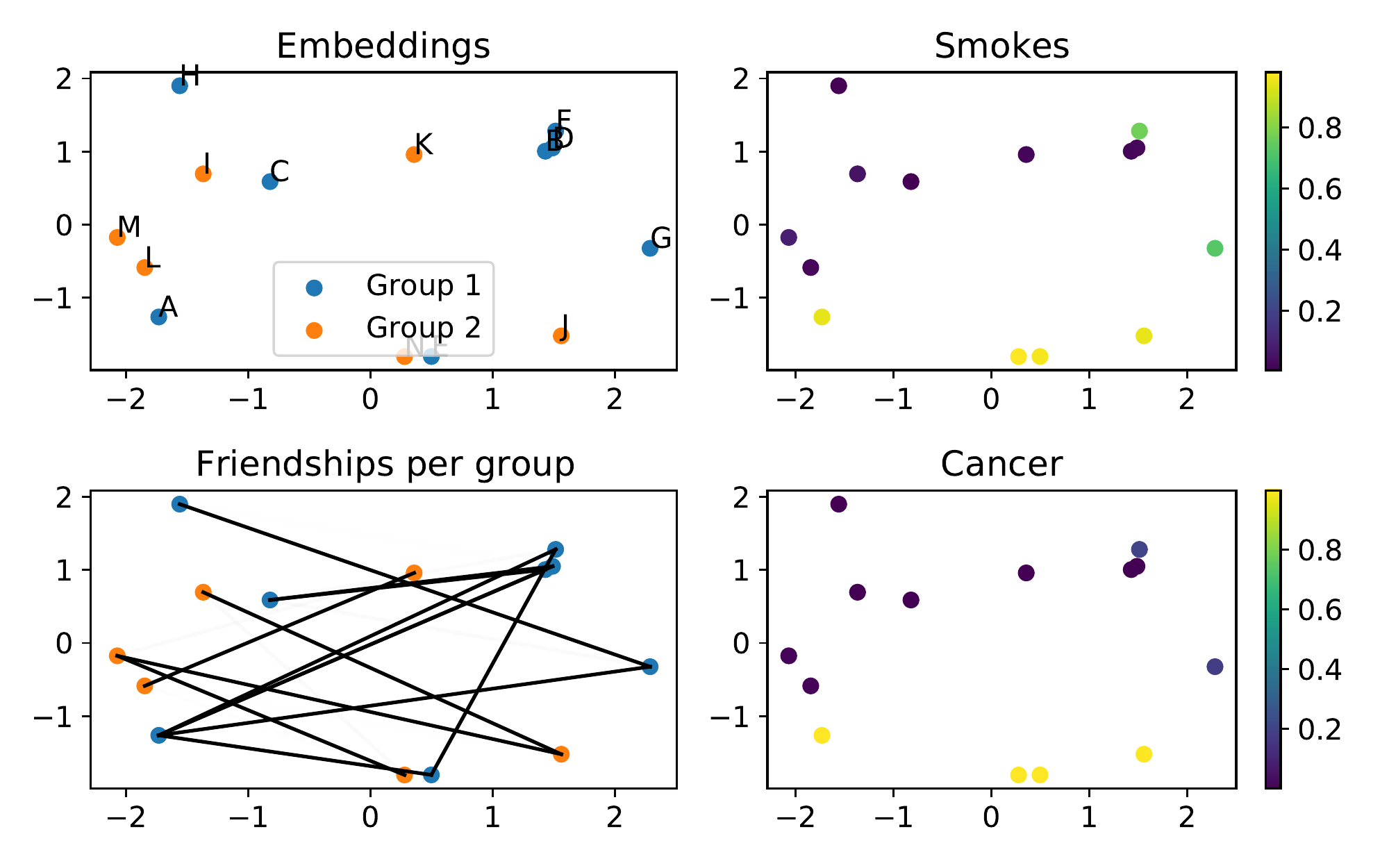}
    \caption{Smoker-Friends-Cancer example: learned embeddings showing the result of applying PCA on the individuals (top left); truth-values of smokes and cancer predicates for each embedding (top and bottom right); illustration of the friendship relations which are satisfied after learning (bottom left).}
    \label{f:example_smokes_embeddings}
\end{figure}

There are 14 people divided into two groups $\{a,b,\dots,h\}$ and $\{i,j,\dots,n\}$. 
Within each group, there is complete knowledge about smoking habits. 
In the first group, there is complete knowledge about who has and who does not have cancer. 
Knowledge about the friendship relation is complete within each group only if symmetry is assumed, that is, $\forall x,y \ (friends(x,y) \rightarrow friends(y,x))$.
Otherwise, knowledge about friendship is incomplete in that it may be known that e.g.\ $a$ is a friend of $b$, and it may be not known whether $b$ is a friend of $a$.
Finally, there is general knowledge about smoking, friendship, and cancer, namely that smoking causes cancer, friendship is normally symmetric and anti-reflexive, everyone has a friend, and smoking propagates (actively or passively) among friends. 
All this knowledge is represented in the axioms further below.
\begin{description}
\item[Domains:] \ \\
    $\mathrm{people}$, to denote the individuals.
\item[Constants:] \ \\
    $a,b,\dots,h,i,j,\dots,n$, the 14 individuals. Our goal is to learn an adequate embedding for each constant.\\
    $\domof(a)=\domof(b)=\dots=\domof(n)=\mathrm{people}$.
\item[Variables:] \ \\
    $x$,$y$ ranging over the individuals.\\
    $\domof(x)=\domof(y)=\mathrm{people}$.
\item[Predicates:] \ \\
    $S(x)$ for \emph{smokes}, $F(x,y)$ for \emph{friends}, $C(x)$ for \emph{cancer}.\\
    $\domof(S)=\domof(C)=\mathrm{people}$. $\domof(F)=\mathrm{people},\mathrm{people}$.
\item[Axioms:] \ \\
    Let $\mathcal{X}_1 = \{a,b,\dots,h\}$ and $\mathcal{X}_2 = \{i,j,\dots,n\}$ be the two groups of individuals.\\
    Let $\mathcal{S} = \{a,e,f,g,j,n\}$ be the smokers; knowledge is complete in both groups.\\
    Let $\mathcal{C} = \{a,e\}$ be the individuals with cancer; knowledge is complete in $\mathcal{X}_1$ only.\\
    Let $\mathcal{F} = \{(a,b),(a,e),(a,f),(a,g),(b,c),(c,d),(e,f),(g,h),(i,j),(j,m),(k,l),(m,n)\}$ be the set of friendship relations; knowledge is complete if assuming symmetry.\\
    These facts are illustrated in Figure \ref{f:smokes_incomplete_facts}.\\
    We have the following axioms:
    \begin{align}
        \qquad \qquad & F(u,v) && \text{for } (u,v) \in \mathcal{F} \\
        & \lnot F(u,v) && \text{for } (u,v) \notin \mathcal{F},\ u > v \\
        & S(u) && \text{for } u \in \mathcal{S} \\
        & \lnot S(u) && \text{for } u \in (\mathcal{X}_1 \cup \mathcal{X}_2 ) \setminus \mathcal{S} \\  
        & C(u) && \text{for } u \in \mathcal{C} \\
        & \lnot C(u) && \text{for } u \in \mathcal{X}_1 \setminus \mathcal{C}\\
        \label{eq:smokes_axiom_F_antireflexive}
        \forall x \ & \lnot F(x,x) && \\
        \label{eq:smokes_axiom_F_symmetric}
        \forall x,y \ & (F(x,y) \imp F(y,x)) && \\
        \forall x \exists y \ & F(x,y) && \\
        \forall x,y \ & ((F(x,y) \land S(x)) \imp S(y)) && \label{eq:smokes_axiom_propagate} \\
        \forall x \ & (S(x) \imp C(x)) && \label{eq:smokes_axiom_causecancer}\\
        \forall x \ & (\lnot C(x) \imp \lnot S(x)) &&  
    \end{align}
    Notice that the knowledge base is not satisfiable in the strict logical sense of the word.
    For instance, $f$ is said to smoke but not to have cancer, which is inconsistent with the rule $\forall x \ (S(x) \imp C(x))$.
    Hence, it is important to adopt a fuzzy approach as done with MLN or a many-valued fuzzy logic interpretation as done with LTN.
\item[Grounding:] \ \\
$\G(\mathrm{people}) = \R^5$. The model is expected to learn embeddings in $\R^5$.\\
$\G(a \mid \theta) = \mathbf{v}_\theta(a)$, \dots, $\G(n \mid \theta) = \mathbf{v}_\theta(n)$. 
Every individual is associated with a vector of 5 real numbers. 
The embedding is initialized randomly uniformly.\\
$\G(x \mid \theta) = \G(y \mid \theta) = \left< \mathbf{v}_\theta(a), \dots, \mathbf{v}_\theta(n) \right>$.\\
$\G(S \mid \theta) : x \mapsto \mathtt{sigmoid}(\mathtt{MLP\_S}_\theta(x))$, where $\mathtt{MLP\_S}_\theta$ has 1 output neuron.\\
$\G(F \mid \theta) : x,y \mapsto \mathtt{sigmoid}(\mathtt{MLP\_F}_\theta(x,y))$, where $\mathtt{MLP\_F}_\theta$ has 1 output neuron.\\
$\G(C \mid \theta) : x \mapsto \mathtt{sigmoid}(\mathtt{MLP\_C}_\theta(x))$, where $\mathtt{MLP\_C}_\theta$ has 1 output neuron.\\
The $\mathtt{MLP}$ models for $S$, $F$, $C$ are kept simple, so that most of the learning is focused on the embedding.

\item[Learning:]\ \\
We use the stable real product configuration to approximate the operators.
For $\forall$, we use $\ApmeanError$ with $p=2$ for all the rules, except for rules \eqref{eq:smokes_axiom_F_antireflexive} and \eqref{eq:smokes_axiom_F_symmetric}, where we use $p=6$. 
The intuition behind this choice of $p$ is that no outliers are to be accepted for the friendship relation since it is expected to be symmetric and anti-reflexive, but outliers are accepted for the other rules. 
For $\exists$, we use $\Apmean$ with $p=1$ during the first 200 epochs of training, and $p=6$ thereafter, with the same motivation as that of the schedule used in Section \ref{s:ex_mnist}. 
The formula aggregator is approximated by $\ApmeanError$ with $p=2$.

Figure \ref{f:crv_smokes} shows the satisfiability over 1000 epochs of training.
At the end of one of these runs, we query $S(x)$, $F(x,y)$, $C(x)$ for each individual; the results are shown in Figure \ref{f:smokes_inferred_facts}.
We also plot the principal components of the learned embeddings \cite{PCA} in Figure \ref{f:example_smokes_embeddings}.
The friendship relations are learned as expected: \eqref{eq:smokes_axiom_causecancer} "smoking implies cancer" is inferred for group 2 even though such information was not present in the knowledge base.
For group 1, the given facts for smoking and cancer for the individuals $f$ and $g$ are slightly altered, as these were inconsistent with the rules.
(the rule for smoking propagating via friendship \eqref{eq:smokes_axiom_propagate} is incompatible with many of the given facts). 
Increasing the satisfaction of this rule would require decreasing the overall satisfaction of the knowledge base, which explains why it is partly ignored by LTN during training.
Finally, it is interesting to note that the principal components for the learned embeddings seem to be linearly separable for the smoking and cancer classifiers (c.f. Figure \ref{f:example_smokes_embeddings}, top right and bottom right plots). 

\item[Querying:] \ \\  
To illustrate querying in LTN, 
we query over time two formulas that are not present in the knowledge-base:
\begin{align}
    \phi_1 &:& \forall p : & C(p) \imp S(p) \\
    \phi_2 &:& \forall p,q : & (C(p) \lor C(q)) \imp F(p,q)
\end{align}
We use $p=5$ when approximating $\forall$ since the impact of an outlier at querying time should be seen as more important than at learning time.
It can be seen that as the grounding approaches satisfiability of the knowledge-base, $\phi_1$ approaches \emph{true}, whereas $\phi_2$ approaches \emph{false} (c.f. Figure \ref{f:smokes_incomplete_facts}).

\end{description}

\subsection{Reasoning in LTN}
\label{s:ex_reasoning}
The essence of reasoning is to find out if a closed formula $\phi$ is the logical consequence of a knowledge-base $(\K,\G_\theta,\Theta)$.
Section \ref{s:def_reasoning} introduced two approaches to this problem in LTN:
\begin{itemize}
\item By simply \emph{querying after learning}\footnote{Here, learning refers to Section \ref{s:learning}, which is optimizing using the satisfaction of the knowledge base as an objective.}
    one seeks to verify if for the grounded theories that maximally satisfy $\K$, 
    the grounding of $\phi$ gives a truth-value greater than a threshold $q$.
    This often requires checking an infinite number of groundings.
    Instead, the user approximates the search for these grounded theories by running the optimization a fixed number of times only.
\item Reasoning by \emph{refutation} one seeks to find out a counter-example: a grounding that satisfies the knowledge-base $\K$ but not the formula $\phi$ given the threshold $q$.
A search is performed here using a different objective function.
\end{itemize}

We now demonstrate that reasoning by refutation is the preferred option using a simple example where we seek to find out whether $(A \lor B) \models_q A$. 

\begin{figure}
\centering
\includegraphics[width=0.8\textwidth]{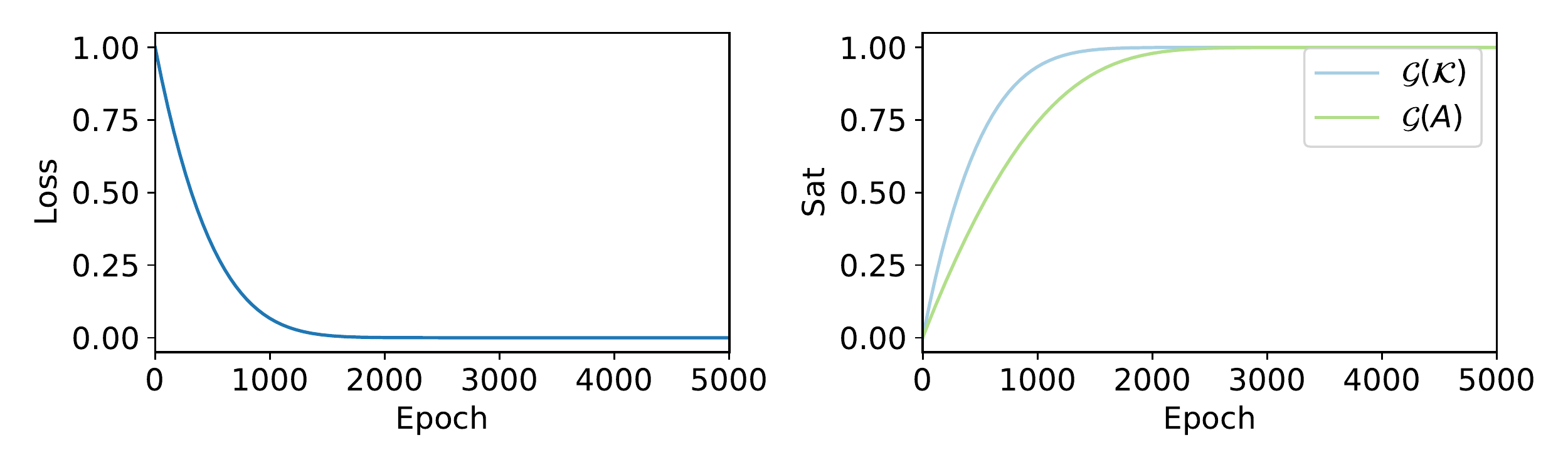}
\caption{Querying after learning: 10 runs of the optimizer with objective $\G^* = \argmax_{\G_\theta} (\G_\theta(\K))$. All runs converge to the optimum $\G_1$; the grid search misses the counter-example.
    }
\label{f:apx_reasoning_brave}
\end{figure}

\begin{figure}
\centering
\includegraphics[width=0.8\textwidth]{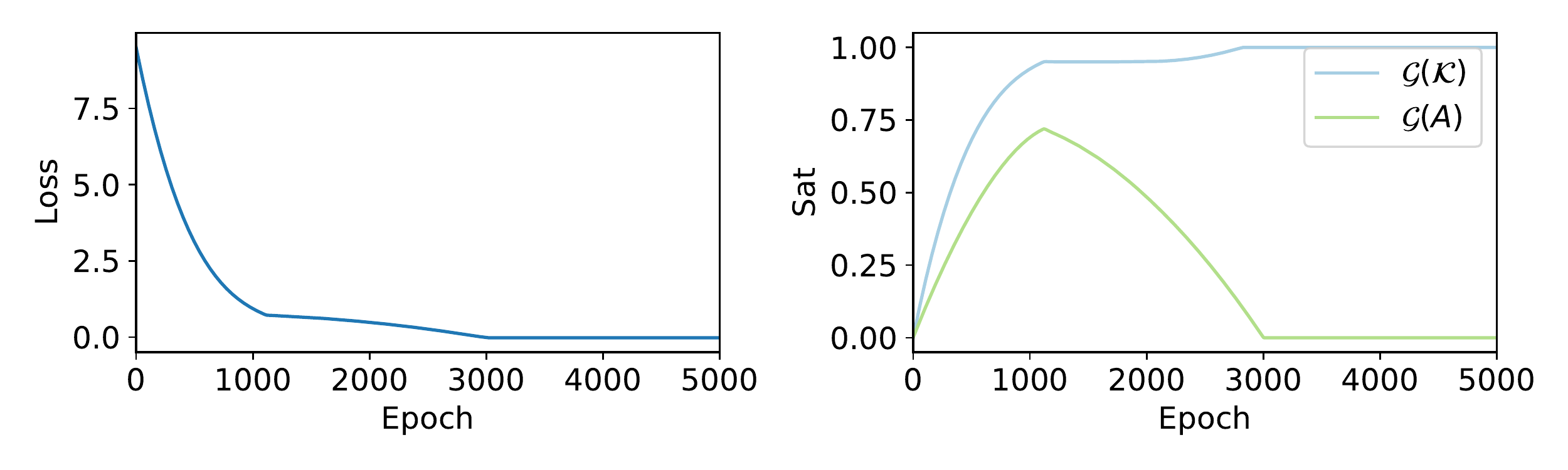}
\caption{Reasoning by refutation: one run of the optimizer with objective $\G^* = \argmin_{\G_\theta} (\G_\theta(\phi) + \mathtt{elu}(\alpha,\beta (q-\G_\theta(\K)))$,
    $q=0.95$, $\alpha=0.05$, $\beta=10$.
    In the first training epochs, the directed search prioritizes the satisfaction of the knowledge base.
    Then, the minimization of $\G_\theta(\phi)$ starts to weigh in more and the search focuses on finding a counter-example. 
    Eventually, the run converges to the optimum $\G_3$, which refutes the logical consequence.
    }
\label{f:apx_reasoning_refut}
\end{figure}

\begin{description}
    \item[Propositional Variables:]\ \\ 
    The symbols $A$ and $B$ denote two propositionial variables. 
    \item[Axioms:]\ \\
    \begin{equation}
        A \lor B
    \end{equation}
    \item[Grounding:]\ \\
    $\G(A)=a$, $\G(B)=b$, where $a$ and $b$ are two real-valued parameters.
    The set of parameters is therefore $\theta=\{a,b\}$.
    At initialization, $a=b=0$.\\
    We use the probabilistic-sum $\Sprod$ to approximate $\lor$, resulting in the following satisfiability measure:\footnote{
        We use the notation $\G(\K) := \Sat_{\phi \in \K}(\K,\G)$.
    }
    \begin{equation}
        \G_\theta(\K) = \G_\theta(A \lor B) = a + b - ab.
    \end{equation}
    There are infinite global optima maximizing the satisfiability of the theory, 
    as any $\G_\theta$ such that $\G_\theta(A)=1$ (resp. $\G_\theta(B)=1$) gives a satisfiability $\G_\theta(\K)=1$ for any value of $\G_\theta(B)$ (resp. $\G_\theta(A)$).
    As expected, the following groundings are examples of global optima:
    \begin{enumerate}[itemsep=1pt, start=1, label={$\G_{\arabic*}$:}]
        \item \label{a:reasoning-globalopt1} $\G_1(A)=1$,\ $\G_1(B)=1$, $\G_1(\K)=1$,
        \item $\G_2(A)=1$,\ $\G_2(B)=0$, $\G_2(\K)=1$,
        \item $\G_3(A)=0$,\ $\G_3(B)=1$, $\G_3(\K)=1$.
    \end{enumerate}
    \item[Reasoning:] \ \\
    $(A \lor B) \models_q A$?
    That is, given the threshold $q=0.95$, does every $\G_\theta$ such that $\G_\theta(\K) \geq q$ verify $\G_\theta(\phi) \geq q$.
    Immediately, one can notice that this is not the case. 
    For instance, the grounding $\G_3$ is a counter-example.

    If one simply reasons by querying multiple groundings after learning with the usual objective $\argmax_{(\G_\theta)}\G_\theta(\K)$, 
    the results will all converge to $\G_1$:  
    $\frac{\partial \G_\theta(\K)}{\partial a} = 1 - b$ and $\frac{\partial \G_\theta(\K)}{\partial b} = 1 - a$.
    Every run of the optimizer will increase $a$ and $b$ simultaneously until they reach the optimum $a=b=1$.
    Because the grid search always converges to the same point, no counter-example is found and the logical consequence is mistakenly assumed true. 
    This is illustrated in Figure \ref{f:apx_reasoning_brave}.
    
    Reasoning by refutation, however, the objective function has an incentive to find a counter-example with $\neg A$, as illustrated in Figure \ref{f:apx_reasoning_refut}.
    LTN converges to the optimum $\G_3$, which refutes the logical consequence.
    
\end{description}

\section{Related Work}
\label{s:relwork}
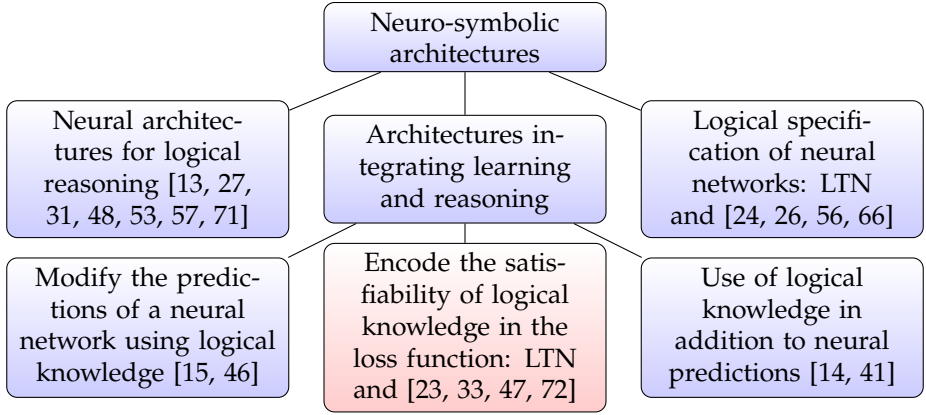
\begin{figure}
\begin{center}
  \begin{tikzpicture}[sibling distance=10em,
  every node/.style = {shape=rectangle, rounded corners,
    draw, align=center,text width=10em,
    top color=white, bottom color=blue!20},level distance=5em,sibling distance=12em]
  \node {Neuro-symbolic architectures}
  child { node {Neural architectures for logical reasoning
      \cite{DBLP:journals/jair/CohenYM20, Garcez2008,hohenecker2020ontology, marra2019neural,qu2019probabilistic,rocktaschel2017end,wang2019satnet}}
        }
  child{ node {Architectures integrating learning and reasoning}
      child[level distance=6em]  { node {Modify the predictions of a neural network using logical
          knowledge \cite{daniele2019knowledge,DBLP:conf/pkdd/MarraGDG19a} 
        }}
      child[level distance=6em]  {node[bottom color=red!20] {Encode the satisfiability of logical knowledge
          in the loss function: LTN and 
          \cite{fischer2019dl2,Harnessing,marra2019lyrics,xu_semantic_2018}
        }}
     child[level distance=6em]  { node {Use of logical knowledge in addition to neural predictions \cite{DAIAbductive,DeepProbLog}}}}
    child { node {Logical specification of neural networks: LTN and
 \cite{franca,Garcez2002,riegel2020logical,sourek2018lifted}
      }};
  \end{tikzpicture}
\end{center}
\caption{\label{fig:related-work-ontology} Three classes of neurosymbolic approaches with \emph{Architectures Integrating Learning and Reasoning} further sub-divided into three sub-classes, with LTN belonging to the sub-class highlighted in red.}
\end{figure}

The past years have seen considerable work aiming to integrate symbolic systems and neural networks. 
We shall focus on work whose objective is to build
computational models that integrate deep learning and logical reasoning into a so-called end-to-end (fully differentiable) architecture. We summarize a categorization in Figure~\ref{fig:related-work-ontology} where
the class containing LTN is further expanded into three
sub-classes. The sub-class highlighted in red is the one that contains LTN. The reason why one may wish to combine symbolic AI and neural networks into a neurosymbolic AI system may vary, c.f. \cite{3rdWave} for a recent comprehensive overview of approaches and challenges for neurosymbolic AI. 

\subsection{Neural architectures for logical reasoning}
\label{s:relwork_nn_for_logic}
These use
  neural networks to perform (probabilistic) inference on logical
  theories. Early work in this direction has shown correspondences
  between various logical-symbolic systems and neural network models
  \cite{Garcez2008,chcl,pinkas,DBLP:journals/apin/Shastri99,harmonicmind}.
  They have also highlighted the limits of current neural networks as
  models for knowledge representation. In a nutshell, current neural
  networks (including deep learning) have been shown capable of
  representing propositional logic, nonmonotonic logic programming,
  propositional modal logic, and fragments of first-order logic, but
  not full first-order or higher-order logic. Recently, there has been
  a resurgence of interest in the topic with many proposals emerging
  \cite{DBLP:journals/jair/CohenYM20,marra2019neural,qu2019probabilistic}. 
  In \cite{DBLP:journals/jair/CohenYM20}, each clause of a Stochastic Logic Program is converted into a factor graph with reasoning becoming 
  differentiable so that it can be implemented by deep networks. In
  \cite{minervini2020learning}, a differentiable unification algorithm
  is introduced with theorem proving sought to be carried out inside
  the neural network. Furthermore, in
  \cite{DBLP:journals/corr/abs-1809-02193,minervini2020learning}
  neural networks are used to learn reasoning strategies and logical
  rule induction.

Reasoning with LTN (Section \ref{s:def_reasoning}) is reminiscent of this category,
given that knowledge is not represented in a traditional logical language but in Real Logic.

\subsection{Logical specification of neural network architectures}
\label{s:relwork_logic_for_nn}
Here the goal is to use a logical language to specify the architecture of a neural network. Examples include
  \cite{DBLP:journals/jair/CohenYM20,franca,Garcez2002,riegel2020logical,sourek2018lifted}.
  In \cite{Garcez2002}, the languages of extended logic programming
  (logic programs with negation by failure) and answer set programming
  are used as background knowledge to set up the initial architecture
  and set of weights of a recurrent neural network, which is
  subsequently trained from data using backpropagation. In \cite{franca}, first-order logic programs in the
  form of Horn clauses are used to define a neural network that can
  solve Inductive Logic Programming tasks, starting from the most
  specific hypotheses covering the set of examples.
  Lifted relational neural
  networks \cite{sourek2018lifted} is a declarative framework where a
  Datalog program is used as a compact specification of a diverse range of existing advanced neural
  architectures, with a particular focus on Graph Neural Networks
  (GNNs) and their generalizations.
  In \cite{riegel2020logical} a
  weighted Real Logic is introduced and used to specify neurons in a
  highly modular neural network that resembles a tree structure,
  whereby neurons with different activation functions are used to
  implement the different logic operators.

  To some extent, it is also possible to specify neural architectures using logic in LTN.
  For example, a user can define a classifier 
  $P(x,y)$ as the formula $P(x,y) = (Q(x,y) \land R(y)) \lor S(x,y)$. 
  $\G(P)$ becomes a computational graph that combines the sub-architectures $\G(Q)$, $\G(R)$, and $\G(S)$ according to the syntax of the logical formula.

\subsection{Neurosymbolic architectures for the integration of inductive
learning and deductive reasoning}
These architectures seek to
  enable the integration of inductive and deductive reasoning in a
  unique fully differentiable framework
  \cite{daniele2019knowledge,fischer2019dl2,DeepProbLog,DBLP:conf/pkdd/MarraGDG19a,marra2019lyrics}.
  The systems that belong to this class combine a neural component
  with a logical component. The former consists of one or more neural
  networks, the latter provides a set of algorithms for performing
  logical tasks such as model checking, satisfiability, and logical
  consequence. These two components are tightly integrated so that
  learning and inference in the neural component are influenced by
  reasoning in the logical component and vice versa. Logic Tensor
  Networks belong to this category. Neurosymbolic
architectures for integrating learning and reasoning can be further separated into three
sub-classes: \begin{enumerate}
  \item Approaches that introduce additional layers to the
neural network to encode logical constraints which modify the
predictions of the network. This sub-class includes Deep
Logic Models \cite{DBLP:conf/pkdd/MarraGDG19a} and Knowledge Enhanced Neural
Networks \cite{daniele2019knowledge}. 
\item Approaches that integrate
logical knowledge as additional constraints in the objective function or loss function used to train the neural network (LTN and 
\cite{fischer2019dl2,Harnessing,marra2019lyrics}). 
\item Approaches that apply
(differentiable) logical inference to compute the consequences
of the predictions made by a set of base neural networks. Examples of
this sub-class are DeepProblog \cite{DeepProbLog} and Abductive Learning
\cite{DAIAbductive}.
\end{enumerate}

In what follows, we revise recent neurosymbolic
architectures in the same class as LTN: \emph{Integrating learning and reasoning}. 

\paragraph{Systems that modify the predictions of a base neural
  network:} Among the approaches that modify the predictions of the neural network
using logical constraints are Deep Logic Models
\cite{DBLP:conf/pkdd/MarraGDG19a} and Knowledge Enhanced Neural Networks
\cite{daniele2019knowledge}. Deep Logic Models (DLM) are a general
architecture for learning with constraints. Here, we will consider the
special case where constraints are expressed by logical formulas. In
this case, a DLM predicts the truth-values of a set of $n$
ground atoms of a domain $\Delta=\{a_1,\dots,a_k\}$. It consists of two models: a neural network $f(\bm x\mid\bm w)$ which takes
as input the features $\bm x$ of the elements of $\Delta$ and produces
as output an evaluation $\bm f$ for all the ground atoms, i.e.
$\bm f\in[0,1]^n$, and a probability distribution
$p(\bm y\mid \bm f,\bm \lambda)$ which is modeled by an undirected graphical model of the exponential family with each logical constraint characterized by a clique that contains the ground
atoms, rather similarly to GNNs. The model returns the assignment to the atoms that maximize the weighted truth-value of the constraints and minimize the
difference between the prediction of the neural network and a target value $\bm y$. Formally:

$$
\text{DLM}(\bm x\mid\bm\lambda,\bm w) =
\argmax_{\bm y}\left(\sum_c\lambda_c\Phi_c(\bm y_c)-\frac12||\bm y -
  \bm f(\bm x\mid\bm w)||^2\right)
$$

Each $\Phi_c(\bm y_c)$ corresponds to a ground propositional formula which is
evaluated w.r.t. the target truth assignment $\bm y$, and $\lambda_c$ is the
weight associated with formula $\Phi_c$. Intuitively, the upper
model (the undirected graphical model) should modify the
prediction of the lower model (the neural network) minimally 
to satisfy the constraints. $\bm f$ and $\bm y$
are truth-values of all the ground atoms obtained from the constraints
appearing in the upper model in the domain specified by the data
input. 

Similar to LTN, DLM evaluates constraints using fuzzy semantics. However, it considers only propositional connectives, whereas universal and existential quantifiers are supported in LTN.

Inference in DLM requires maximizing the prediction of the
model, which might be prohibitive in the presence of a large number of instances. In LTN, inference involves only a forward pass through the
neural component which is rather simple and can be carried out in parallel. However, in DLM
the weight associated with constraints can be
learned, while in LTN they are specified in the background
knowledge. 

The approach taken in Knowledge Enhanced Neural Networks (KENN)
\cite{daniele2019knowledge} is similar to that of DLM. Starting from
the predictions $\bm y = f_{nn}(\bm x\mid\bm w)$ made by a base neural
network $f_{nn}(\cdot\mid\bm w)$, KENN adds a \emph{knowledge enhancer}, which is a function that modifies $\bm y$
based on a set of weighted constraints formulated in terms of
clauses. The formal model can be specified as follows:

$$
\text{KENN}(\bm x\mid \bm\lambda,\bm w) =
\sigma(f'_{nn}(\bm x\mid \bm w) +
\sum_c\lambda_c\cdot
(\mathrm{softmax}(\mathrm{sign}(c) \odot f'_{nn}(\bm x\mid \bm w))
\odot \mathrm{sign}(c)))
$$
where $f'_{nn}(\bm x\mid \bm w)$ are the pre-activations of
$f_{nn}(\bm x\mid\bm w)$, $\mathrm{sign}(c)$ is a vector 
of the same dimension of $\bm y$ containing $1,-1$ and
$-\infty$, such that $\mathrm{sign}(c)_i=1$ 
(resp. $\mathrm{sign}(c)_i=-1$) if the $i$-th atom occurs positively (resp. negatively) in $c$, or $-\infty$ otherwise, and $\odot$ is the
element-wise product. KENN learns the weights $\bm \lambda$ of the
clauses in the background
knowledge and the base network parameters $\bm w$ by minimizing some
standard loss, (e.g. cross-entropy) on a set of training data. 
If the training data is
inconsistent with the constraint, the weight of the constraint will be
close to zero. This intuitively implies that the latent knowledge
present in the data is preferred to the knowledge specified in the
constraints. In LTN, instead, training data and logical constraints are
represented uniformly with a formula, and we require that they are
both satisfied. A second difference between KENN and LTN is the
language: while LTN supports constraints written in full first-order
logic, constraints in KENN are limited to universally quantified
clauses. 

\paragraph{Systems that add knowledge to a neural network by adding a term to the loss function:} In \cite{Harnessing}, a framework is proposed that learns simultaneously from labeled data and logical rules. The proposed architecture is made of a \emph{student}
network $f_{nn}$ and a \emph{teacher} network, denoted by $q$. The student network is trained to
do the actual predictions, while the teacher network encodes the
information of the logical rules. The transfer of information from the
teacher to the student network is done by defining a joint loss
$\mathscr{L}$ for both networks as a convex combination of the loss of the student and the teacher.
If $\tilde{\bm y}=f_{nn}(\bm x\mid\bm w)$ is the prediction of the
student network for input $\bm x$, the loss is defined as: 

$$
(1-\pi)\cdot\mathscr{L}(\bm y,\tilde{\bm y}) +
  \pi\cdot\mathscr{L}(q(\tilde{\bm y}\mid\bm x),\tilde{\bm y})
$$
where $q(\tilde{\bm y}\mid \bm x) =\exp\left(-\sum_c\lambda_c(1-\phi_c(\bm
  x,\tilde{\bm y}))\right)$ measures how much the predictions
$\tilde{\bm y}$ satisfy the constraints encoded in the set of
clauses $\{\lambda_c:\phi_c\}_{c\in C}$. Training is
iterative. At every iteration, the parameters of the student network
are optimized to minimize the loss that takes into account
the feedback of the teacher network on the predictions from the
previous step. The main difference between this approach and LTN 
is how the constraints are encoded in the loss. LTN integrates the
constraints in the network and optimizes directly their
satisfiability with no need for additional training data. Furthermore,
the constraints proposed in \cite{Harnessing} are universally
quantified formulas only. 

The approach adopted by \textsc{Lyrics} \cite{marra2019lyrics} is
analogous to the first version of LTN \cite{serafini2016logic}. Logical constraints are translated into a loss function that measures
the (negative) satisfiability level of the network. Differently from
LTN, formulas in \textsc{Lyrics} can be associated with weights that are
hyper-parameters. In \cite{marra2019lyrics}, a logarithmic loss function is also used when the product t-norm is adopted. Notice that weights
can also be added (indirectly) to LTN by introducing a $0$-ary
predicate $p_w$ to represent a constraint of the form $p_w\wedge\phi$. An
advantage of this approach would be that the weights could be learned.

In \cite{xu_semantic_2018}, a neural network computes the probability of some events being true.
The neural network should satisfy a set of propositional logic constraints on its output.
These constraints are compiled into arithmetic circuits for weighted model counting, which are then used to compute a loss function. 
The loss function then captures how close the neural network is to satisfying the propositional logic constraints.

\paragraph{Systems that apply logical reasoning on the predictions of
  a base neural network:} 
The most notable architecture in this category is
DeepProblog \cite{DeepProbLog}. DeepProblog extends the ProbLog framework for
probabilistic logic programming to allow the computation of
probabilistic evidence from neural networks. A ProbLog program is a
logic program where facts and rules can be associated with probability values. Such values can be learned. Inference in ProbLog to answer a query $q$ is performed by knowledge
compilation into a function $p(q \mid \bm\lambda)$ 
that computes the probability that $q$ is
true according to the logic program with relative frequencies $\bm \lambda$. In DeepProbLog, a neural network $f_{nn}$ that outputs a probability distribution
$\bm t=(t_1,\dots,t_n)$ over a set of atoms $\bm a=(a_1,\dots,a_n)$ is
integrated into ProbLog by extending the logic program with $\bm a$ and the respective probabilities $\bm t$. The probability of a query $q$ is then given by $p'(q\mid\bm \lambda,f_{nn}(x\mid \bm w))$, where $x$
is the input of $f_{nn}$ and $p'$ is the function corresponding to the logic program extended with
$\bm a$. Given a set of
queries $\bm q$, input vectors $\bm x$ and 
ground-truths $\bm y$ for all the queries,
training is performed by minimizing a loss function that measures
the distance between the probabilities predicted by the logic
program and the ground-truths, as follows: 
$$
\mathscr{L}(\bm y, p'(\bm q\mid\bm
\lambda,f_{nn}(\bm x\mid \bm w)))
$$
The most important difference between DeepProbLog and LTN concerns the
logic on which they are based. DeepProbLog adopts probabilistic
logic programming. The output of the base neural network is interpreted as the
probability of certain atoms being true. LTN instead is based on many-valued logic. 
The predictions of the base neural network are interpreted as fuzzy truth-values 
(though previous work \cite{van_krieken_semi-supervised_2019} also formalizes Real Logic as handling probabilities with relaxed constraints). 
This difference of logic leads to the second main difference between LTN and DeepProblog: their inference mechanism.  DeepProblog performs probabilistic inference
(based on model counting) while LTN inference consists of computing
the truth-value of a formula starting from the truth-values of its
atomic components. The two types of inference are incomparable. 
However, computing the fuzzy truth-value of a formula is more efficient
than model counting, resulting in a more scalable 
inference task that allows LTN to use full first-order logic with function symbols. In DeepProblog, to perform probabilistic inference, a closed-world
assumption is made and a function-free language is used. Typically, DeepProbLog clauses are compiled into
Sentential Decision Diagrams (SDDs) to accelerate inference considerably\cite{Darwiche}, although the compilation step of clauses into the SDD circuit is still costly. 

An approach that extends the predictions of a base neural network using
abductive reasoning is \cite{DAIAbductive}. Given a neural network
$f_{nn}(\bm x\mid\bm w)$ that produces a crisp output
$\bm y\in\{0,1\}^n$ for $n$ predicates $p_1,\dots,p_n$ and background
knowledge in the form of a logic program $p$, parameters $\bm w$ of
$f_{nn}$ are learned alongside a set of additional rules $\Delta_C$
that define a new concept $C$ w.r.t. $p_1,\dots,p_n$ such that, for
every object $o$ with features $\bm x_o$:
\begin{align}
\label{eq:abdutive-neuro-symb}
  \begin{array}{l@{\ \ \ \ }l} 
  p \cup f_{nn}(\bm x_o\mid\bm w)\cup\Delta_C \models C(o)
  & \mbox{if $o$ is an instance of $C$} \\ 
p \cup f_{nn}(\bm x_o\mid\bm w)\cup\Delta_C \models \neg C(o)
  & \mbox{if $o$ is not an instance of $C$}
  \end{array}
\end{align}
The task is solved by iterating the following three steps: 
\begin{enumerate}
\item Given the predictions of the neural network $\{f_{nn}(\bm x_o\mid\bm
  w)\}_{o\in O}$ on the set $O$ of training objects, search for the best $\Delta_C$ that
maximize the number of objects for which
\eqref{eq:abdutive-neuro-symb} holds;
\item For each object $o$, compute by abduction on $p\cup\Delta_C$,
  the explanation $\bm p'(o)$;
\item Retrain $f_{nn}$ with the training set $\{\bm x_o,\bm
  p'(o)\}_{o\in O}$. 
\end{enumerate}

Differently from LTN, in \cite{DAIAbductive} the optimization is done
separately in an iterative way. The
semantics of the logic is crisp, neither fuzzy nor probabilistic, and therefore not fully differentiable. Abductive reasoning is adopted, which is a potentially relevant addition for comparison with symbolic ML and Inductive Logic Programming approaches \cite{metagol}. 
 
Various other loosely-coupled approaches have been proposed recently such as \cite{ConceptLearner}, where image classification is carried out by a neural network in combination with reasoning from text data for concept learning at a higher level of abstraction than what is normally possible with pixel data alone. The proliferation of such approaches has prompted Henry Kautz to propose a taxonomy for neurosymbolic AI in  \cite{HenryKautz} (also discussed in \cite{3rdWave}), including recent work combining neural networks with graphical models and graph neural networks \cite{Battaglia,DBLP:conf/ijcai/LambGGPAV20,GNN}, statistical relational learning \cite{dILP,MLN}, and even verification of neural multi-agent systems \cite{lomuscio,borges}. 

\section{Conclusions and Future Work}
\label{s:concl}

In this paper, we have specified the theory and exemplified the reach of Logic Tensor Networks as a model and system for neurosymbolic AI. LTN is capable of combining approximate reasoning and deep learning, knowledge and data. 

%\subsection{On Learning and Reasoning}
For ML practitioners, learning in LTN (see Section \ref{s:learning}) 
can be understood as optimizing under first-order logic constraints relaxed into a loss function.
For logic practitioners, learning is similar to inductive inference:
given a theory, learning makes generalizations from specific observations obtained from data.
Compared to other neuro-symbolic architectures (see Section \ref{s:relwork}), 
the LTN framework has useful properties for gradient-based optimization (see Section \ref{s:stableproduct})
and a syntax that supports many traditional ML tasks and their inductive biases (see Section \ref{s:examples}),
all while remaining computationally efficient (see Table \ref{table:cnn}).

Section \ref{s:def_reasoning} discussed reasoning in LTN. Reasoning is normally under-specified within neural networks. Logical reasoning is the task of proving if some knowledge follows from the facts which are currently known.
It is traditionally achieved semantically using model theory or syntactically via a proof system.
The current LTN framework approaches reasoning semantically, although it should be possible to use LTN and querying alongside a proof system. 
When reasoning by refutation in LTN, to find out if a statement $\phi$ is a logical consequence of given data and knowledge-base $\K$, a proof by refutation attempts to find a semantic counterexample where $\lnot \phi$ and $\K$ are satisfied. If the search fails then $\phi$ is assumed to hold. This approach is efficient in LTN when we allow for a direct search to find counterexamples via gradient-descent optimization. It is assumed that $\phi$, the statement to prove or disprove, is known.
Future work could explore automatically inducing which statement $\phi$ to consider, possibly using syntactical reasoning in the process.

%\subsection{On the link with other semantics}
The paper formalizes Real Logic, the language supporting LTN.
The semantics of Real Logic are close to the semantics of Fuzzy FOL with the following major differences:
1) Real Logic domains are typed and restricted to real numbers and real-valued tensors,
2) Real Logic variables are sequences of fixed length, whereas FOL variables are a placeholder for any individual in a domain,
3) Real Logic relations relations are interpreted as mathematical functions, whereas Fuzzy Logic relations are interpreted as fuzzy set membership functions. 
Concerning the semantics of connectives and quantifiers, some LTN implementations correspond to semantics for t-norm fuzzy logic, but not all.
For example, the conjunction operator in stable product semantics is not a t-norm, as pointed out at the end of Section \ref{s:stableproduct}. 

%\subsection{On future work}
Integrative neural-symbolic approaches are known for either seeking to bring neurons into a symbolic system (neurons into symbols) \cite{DeepProbLog} or to bring symbols into a neural network (symbols into neurons) \cite{DBLP:journals/corr/abs-1910-06611}. LTN adopts the latter approach but maintaining a close link between the symbols and their grounding into the neural network. The discussion around these two options - neurons into symbols vs. symbols into neurons - is likely to take center stage in the debate around neurosymbolic AI in the next decade. LTN and related approaches are well placed to play an important role in this debate by offering a rich logical language tightly coupled with an efficient distributed implementation into TensorFlow computational graphs.

The close connection between first-order logic and its implementation in LTN makes LTN very suitable as a model for the neural-symbolic cycle
\cite{Garcez2008,DBLP:series/sci/2007-77}, which seeks to translate between neural and symbolic representations.
Such translations can take place at the level of the structure of a neural network, given a symbolic language \cite{Garcez2008}, or at the level of the loss functions, as done by LTN and related approaches \cite{DBLP:journals/jair/CohenYM20,DBLP:conf/icann/MarraGDG19,DBLP:conf/pkdd/MarraGDG19a}. LTN opens up a number of promising avenues for further research: 

Firstly, a continual learning approach might allow one to start with very little knowledge, build up and validate knowledge over time by querying the LTN network. Translations to and from neural and symbolic representations will enable reasoning also to take place at the symbolic level (e.g. alongside a proof system), as proposed recently in \cite{wagner_neural-symbolic_2021} with the goal of improving fairness of the network model.

Secondly, LTN should be compared in large-scale practical use cases with other recent efforts to add structure to neural networks such as the neuro-symbolic concept learner \cite{ConceptLearner} and high-level capsules which were used recently to learn the \textit{part-of} relation \cite{NIPS2019_9684}, similarly to how LTN was used for semantic image interpretation in \cite{LTNIJCAI}.

Finally, LTN should also be compared with Tensor Product Representations, e.g. \cite{TPRRNN}, which show that state-of-the-art recurrent neural networks may fail at simple question-answering tasks, despite achieving very high accuracy. Efforts in the area of transfer learning, mostly in computer vision, which seek to model systematicity could also be considered a benchmark \cite{Bengio2020A}. Experiments using fewer data and therefore lower energy consumption, out-of-distribution extrapolation, and knowledge-based transfer are all potentially suitable areas of application for LTN as a framework for neurosymbolic AI based on learning from data and compositional knowledge.

\section*{Acknowledgement}
We would like to thank Benedikt Wagner for his comments and a number of  productive discussions on continual learning, knowledge extraction and reasoning in LTNs.

\bibliographystyle{plain}
\bibliography{biblio}

\appendix
\clearpage
\section{Implementation Details}
\label{a:impl}

The LTN library is implemented in Tensorflow 2~\cite{tensorflow2015-whitepaper} and is available from GitHub\footnote{https://github.com/logictensornetworks/logictensornetworks}.
Every logical operator is grounded using Tensorflow primitives.
The LTN code implements directly a Tensorflow graph.
Due to Tensorflow built-in optimization, LTN is relatively efficient while providing the expressive power of FOL.

Table \ref{tab:network_architectures} shows an overview of the network architectures used to obtain the results of the examples in Section \ref{s:examples}.
The LTN repository includes the code for these examples.
Except if explicitly mentioned otherwise, the reported results are averaged over 10 runs using a 95\% confidence interval.
Every example uses a stable real product configuration to approximate Real Logic operators, and the Adam optimizer~\cite{kingma_adam_2017} with a learning rate of $0.001$ to train the parameters.

\newcommand{\Dense}{\mathrm{Dense}}
\newcommand{\Dropout}{\mathrm{Dropout}}
\newcommand{\Conv}{\mathrm{Conv}}
\newcommand{\MNISTConv}{\mathrm{MNISTConv}}

\begin{table}[h]
\centering
\begin{threeparttable}
\begin{tabular}{p{1cm} p{3cm} p{10cm}}
\toprule
Task & Network & Architecture\\
\midrule
\ref{s:ex_binary} & $\mathtt{MLP}$ & $\Dense(16)^\ast$, $\Dense(16)^\ast$, $\Dense(1)$\\ 
\ref{example:multiclasssinglelabel} & $\mathtt{MLP}$ & $\Dense(16)^\ast$, $\Dropout(0.2)$, $\Dense(16)^\ast$, $\Dropout(0.2)$,\\
  &  &  $\Dense(8)^\ast$, $\Dropout(0.2)$, $\Dense(1)$  \\ 
\ref{s:ex_multilabel} & $\mathtt{MLP}$ & $\Dense(16)^\ast$, $\Dense(16)^\ast$, $\Dense(8)^\ast$, $\Dense(1)$\\ 
\ref{s:ex_mnist} & $\mathtt{CNN}$ & $\MNISTConv$, $\Dense(84)^\ast$, $\Dense(10)$\\
 &$\mathrm{baseline-SD}$ & $\MNISTConv\times 2$, $\Dense(84)^\ast$, $\Dense(19)$, $\mathrm{Softmax}$\\
 &$\mathrm{baseline-MD}$ & $\MNISTConv\times 4$, $\Dense(128)^\ast$, $\Dense(199)$, $\mathrm{Softmax}$\\
\ref{s:ex_regression} & $\mathtt{MLP}$ &  $\Dense(8)^\ast$, $\Dense(8)^\ast$, $\Dense(1)$ \\
\ref{s:clustering} & $\mathtt{MLP}$ & $\Dense(16)^\ast$, $\Dense(16)^\ast$, $\Dense(16)^\ast$, $\Dense(1)$ \\
\ref{s:ex_smokes} & $\mathtt{MLP\_S}$ & $\Dense(8)^\ast$, $\Dense(8)^\ast$, $\Dense(1)$\\
& $\mathtt{MLP\_F}$ &  $\Dense(8)^\ast$, $\Dense(8)^\ast$, $\Dense(1)$\\
& $\mathtt{MLP\_C}$ &  $\Dense(8)^\ast$, $\Dense(8)^\ast$, $\Dense(1)$\\
\bottomrule    
\end{tabular}
\begin{tablenotes}
    \setlength\labelsep{0pt}
    \item $^\ast$ : layer ends with an $\mathtt{elu}$ activation 
    \item $\Dense(n)$ : regular fully-connected layer of $n$ units
    \item $\Dropout(r)$ : dropout layer with rate $r$
    \item $\Conv(f,k)$ : 2D convolution layer with $f$ filters and a kernel of size $k$
    \item $\mathrm{MP}(w,h)$ : max pooling operation with a $w \times h$ pooling window
    \item $\MNISTConv$ : $\Conv(6,5)^\ast$, $\mathrm{MP}(2,2)$, $\Conv(16,5)^\ast$, $\mathrm{MP}(2,2)$, $\Dense(100)^\ast$
\end{tablenotes}
\end{threeparttable}
\caption{Overview of the neural network architectures used in each example. 
Notice that in the examples, the networks are usually used with some additional layer(s) to ground symbols.
For instance, in experiment \ref{example:multiclasssinglelabel}, in $\G(P) : x,l \mapsto l^\top \mathtt{softmax}(\mathtt{MLP}(x))$,
the softmax layer normalizes the raw predictions of $\mathtt{MLP}$ to probabilities in $[0,1]$,
and the multiplication with the one-hot label $l$ selects the probability for one given class. 
}
\label{tab:network_architectures}
\end{table}

\clearpage
\section{Fuzzy Operators and Properties}
\label{a:operators}

This appendix presents the most common operators used in fuzzy logic literature and some noteworthy properties~\cite{hajek_metamathematics_1998,klement_triangular_2000,shi2009deep,van_krieken_analyzing_2020}.

\begin{table}[b]
\centering
\begin{tabular}{c c c c c }
     \hline
     Name 
          & $ a \land b $ 
          & $ a \lor b $ 
          & $a \rightarrow_R c $ 
          & $a \rightarrow_S c $ \\
     \hline
     Goedel 
          & $\min(a,b)$ 
          & $\max(a,b)$ 
          &  $ \begin{cases}
               1, & \text{if } a \leq c \\
               c, & \text{otherwise} 
               \end{cases} $ 
          & $\max(1-a, c)$
          \\
     Goguen/Product 
          & $a \cdot b$ 
          & $a+b-a \cdot b$ 
          & $\begin{cases}
               1, & \text{if } a \leq c \\
               \frac{c}{a}, & \text{otherwise}
               \end{cases} $ 
          & $1-a+a\cdot c$
          \\
     \L{}ukasiewicz 
          & $ \max(a+b-1,0) $ 
          & $\min(a+b, 1)$  
          & $\min(1-a+c, 1)$
          & $\min(1-a+c, 1) $
          \\
     \hline
\end{tabular}
\caption{Common Symmetric Configurations}
\end{table}

\subsection{Negation}

\begin{definition}
A negation is a function $N : [0,1] \rightarrow [0,1]$ that at least satisfies:
\begin{enumerate}[itemsep=1pt, start=1,label={N\arabic*.}]
     \item Boundary conditions: $N(0)=1$ and $N(1)=0$,
     \item Monotonically decreasing: $\forall(x,y)\in[0,1]^2,\ x \leq y \rightarrow N(x) \geq N(y)$.
\end{enumerate}
Moreover, a negation is said to be \emph{strict} if $N$ is continuous and strictly decreasing. 
A negation is said to be \emph{strong} if $\forall x \in [0,1],\ N(N(x))=x$.
\end{definition} 

We commonly use the standard strict and strong negation $\Ns(a) = 1-a$.

\subsection{Conjunction}

\begin{definition}
A conjuction is a function $C : [0,1]^2 \rightarrow [0,1]$ that at least satisfies:
\begin{enumerate}[itemsep=1pt, start=1,label={C\arabic*.}]
     \item boundary conditions: $C(0,0) = C(0,1) = C(1,0)=0$ and $C(1,1)=1$,
     \item monotonically increasing: $\forall(x,y,z)\in[0,1]^3,\ $ 
          if $x \leq y ,\ $ then $C(x,z)\leq C(y,z)$ and $C(z,x) \leq C(z,y)$.
\end{enumerate}
\end{definition}

In fuzzy logic, t-norms are widely used to model conjunction operators.
\begin{definition}
A t-norm (triangular norm) is a function $t : [0,1]^2 \rightarrow [0,1]$ that at least satisifies:
\begin{enumerate}[itemsep=1pt, start=1,label={T\arabic*.}]
     \item boundary conditions: $T(x,1)=x$,
     \item monotonically increasing,
     \item commutative,
     \item associative.
\end{enumerate}

\end{definition}

\begin{example}
Three commonly used t-norms are:
\begin{align}
     \Tmin(x,y) = & \min(x,y) \tag{minimum} \\
     \Tprod(x,y) = & x \cdot y \tag{product}\\
     \Tluk(x,y) = & \max(x+y-1,0) \tag{\L{}ukasiewicz}
\end{align}
\end{example}

\subsection{Disjunction}

\begin{definition}
A disjunction is a function $D : [0,1]^2 \rightarrow [0,1]$ that at least satisfies:
\begin{enumerate}[itemsep=1pt, start=1,label={D\arabic*.}]
\item boundary conditions: $D(0,0) = 0 $ and $D(0,1) = D(1,0) = D(1,1)=1$,
\item monotonically increasing: $\forall(x,y,z)\in[0,1]^3,\ $ 
     if $x \leq y ,\ $ then $D(x,z)\leq D(y,z)$ and $D(z,x) \leq D(z,y)$.
\end{enumerate}
\end{definition}

Disjunctions in fuzzy logic are often modeled with t-conorms.
\begin{definition}
A t-conorm (triangular conorm) is a function $S:[0,1]^2 \rightarrow [0,1]$ that at least satisfies:
\begin{enumerate}[itemsep=1pt, start=1, label={S\arabic*.}]
     \item boundary conditions: $S(x,0)=x$,
     \item monotonically increasing,
     \item commutative,
     \item associative.
\end{enumerate}
\end{definition}

\begin{example}
     Three commonly used t-conorms are:
     \begin{align}
          \Smin(x,y) = & \max(x,y) \tag{maximum} \\
          \Sprod(x,y) = & x + y - x \cdot y \tag{probabilistic sum}\\
          \Sluk(x,y) = & \min(x+y,1) \tag{\L{}ukasiewicz}
     \end{align}
     Note that the only distributive pair of t-norm and t-conorm is $\Tmin$ and $\Smin$ 
     -- that is, distributivity of the t-norm over the t-conorm, and inversely.
\end{example}

\begin{definition}
The $N$-dual t-conorm $S$ of a t-norm $T$ w.r.t. a strict fuzzy negation $N$ is defined as:
\begin{equation}
     \label{eq:demorgan1}
     \forall (x, y) \in [0,1]^2, S(x,y) = N(T(N(x),N(y))).
\end{equation}
If $N$ is a strong negation, we also get:
\begin{equation}
     \label{eq:demorgan2}
     \forall (x, y) \in [0,1]^2, T(x,y) = N(S(N(x),N(y))).
\end{equation}
\end{definition}

\subsection{Implication}

\begin{definition}
An implication is a function $I : [0,1]^2 \rightarrow [0,1]$ that at least satisfies:     
\begin{enumerate}[itemsep=1pt, start=1,label={I\arabic*.}]
     \item boundary Conditions: $I(0,0) = I(0,1) = I(1,1) = 1$ and $I(1,0) = 0$ 
\end{enumerate}

\end{definition}

\begin{definition}
There are two main classes of implications generated from the fuzzy logic operators for negation, conjunction and disjunction.
\begin{description}
     \item[S-Implications] \textit{Strong implications} are defined using
     $x \rightarrow y = \lnot x \lor y$ (\textit{material implication}).
     \item[R-Implications] \textit{Residuated implications} are defined using 
     $x \rightarrow y = \sup\{z \in [0,1] \mid x \land z \leq y \}$. 
     One way of understanding this approach is a generalization of \textit{modus ponens}: 
     the consequent is at least as true as the (fuzzy) conjunction of the antecedent and the implication. 
\end{description}
\end{definition}

\begin{example}
Popular fuzzy implications and their classes are presented in Table \ref{tab:implications}.

\begin{table}
     \centering     
     \begin{tabular}{|c|c|c|c|}
          \hline
          Name & $I(x,y)=$ & S-Implication & R-Implication \\
          \hline
          \thead{Kleene-Dienes \\ $\Ikd$} 
               & $\max(1-x,y)$
               & \thead{$S=\Smin$ \\ $N=\Ns$}
               & - \\
          \hline
          \thead{Goedel \\ $\Igodel$} 
               & $\begin{cases}
                    1, & x \leq y \\
                    y, & \text{otherwise}
               \end{cases}$
               & -
               & $T = \Tmin$ \\
          \hline
          \thead{Reichenbach \\ $\Ir$} 
               & $1 - x + x y$
               & \thead{$S=\Sprod$ \\ $N=\Ns$}
               & - \\
          \hline
          \thead{Goguen \\ $\Iprod$} 
               & $\begin{cases}
                    1, & x \leq y \\
                    y/x, & \text{otherwise}
               \end{cases}$
               & -
               & $T = \Tprod$ \\
          \hline
          \thead{\L{}ukasiewicz \\ $\Iluk$} 
               & $\min(1-x+y,1)$
               & \thead{$S=\Sluk$\\$N=\Ns$}
               & $T = \Tluk$ \\
          \hline
     \end{tabular}
     \caption{Popular fuzzy implications and their classes. 
     Strong implications (S-Implications) are defined using a fuzzy negation and fuzzy disjunction. 
     Residuated implications (R-Implications) are defined using a fuzzy conjunction. }
     \label{tab:implications}
     \end{table}
\end{example}

\subsection{Aggregation}

\begin{definition}
     An aggregation operator is a function $A : \bigcup\limits_{n \in \mathbb{N}} [0,1]^n \rightarrow [0,1]$ that at least satisfies:
     \begin{enumerate}[itemsep=1pt, start=1, label={A\arabic*.}]
          \item $A(x_1,\dots,x_n) \leq A(y_1,...,y_n)$ whenever $x_i \leq y_i$ for all $i \in \{1,\dots,n\}$,
          \item $A(x) = x$ forall $x \in [0,1]$,
          \item $A(0,\dots,0) = 0$ and $A(1,\dots,1) = 1$.
     \end{enumerate}
\end{definition}

\begin{example}
     Candidates for universal quantification $\forall$ can be obtained using t-norms with $A_T(x_i) = x_i$ and $A_T(x_1,\dots,x_n) = T(x_1,A_T(x_2,\dots,x_n))$:
     \begin{align}
          A_{\Tmin}(x_1,\dots,x_n) = & \min(x_1,\dots,x_n) \tag{minimum} \\
          A_{\Tprod}(x_1,\dots,x_n) = & \prod\limits_{i=1}^n x_i \tag{product}\\
          A_{\Tluk}(x_1,\dots,x_n) = & \max(\sum\limits_{i=1}^n x_i-n+1,0) \tag{\L{}ukasiewicz}
     \end{align}

     Similarly, candidates for existential quantification $\exists$ can be obtained using s-norms with $A_S(x_i) = x_i$ and $A_S(x_1,\dots,x_n) = S(x_1,A_S(x_2,\dots,x_n))$:
     \begin{align}
          A_{\Smin}(x_1,\dots,x_n) = & \max(x_1,\dots,x_n) \tag{maximum} \\
          A_{\Sprod}(x_1,\dots,x_n) = & 1-\prod\limits_{i=1}^n (1-x_i) \tag{probabilistic sum}\\
          A_{\Sluk}(x_1,\dots,x_n) = & \min(\sum\limits_{i=1}^n x_i,1) \tag{\L{}ukasiewicz}
     \end{align}

     Following are other common aggregators:
     \begin{align}
          \Amean(x_1,\dots,x_n) = & \frac{1}{n} \sum\limits_{i=1}^n x_i \tag{mean} \\
          \Apmean(x_1,\dots,x_n) = & \biggl( \frac{1}{n} \sum\limits_{i=1}^n x_i^p \biggr)^{\frac{1}{p}} \tag{p-mean}\\
          \ApmeanError(x_1,\dots,x_n) = & 1 - \biggl( \frac{1}{n} \sum\limits_{i=1}^n (1-x_i)^p \biggr)^{\frac{1}{p}} \tag{p-mean error}
     \end{align}

     Where $\Apmean$ is the generalized mean, and $\ApmeanError$ can be understood as the generalized mean 
     measured w.r.t. the errors. 
     That is, $\ApmeanError$ measures the power of the deviation of each value from the ground truth $1$.  
     A few particular values of $p$ yield special cases of aggregators. 
     Notably:
     \begin{itemize}[itemsep=1pt]
          \item $\lim\substack{p \to +\infty} \Apmean(x_1,\dots,x_n) = \max(x_1,\dots,x_n)$,
          \item $\lim\substack{p \to -\infty} \Apmean(x_1,\dots,x_n) = \min(x_1,\dots,x_n)$,
          \item $\lim\substack{p \to +\infty} \ApmeanError(x_1,\dots,x_n) = \min(x_1,\dots,x_n)$,
          \item $\lim\substack{p \to -\infty} \ApmeanError(x_1,\dots,x_n) = \max(x_1,\dots,x_n)$.
     \end{itemize}
     These "smooth" $\min$ (resp. $\max$) approximators are good candidates for $\forall$ (resp. $\exists$) 
     in a fuzzy context. The value of $p$ leaves more or less room for outliers depending
     on the use case and its needs.
     Note that $\ApmeanError$ and $\Apmean$ are related in the same way that 
     $\exists$ and $\forall$ are related using the definition 
     $\exists \equiv \lnot \forall \lnot$, 
     where $\lnot$ would be approximated by the standard negation.

     We propose to use $\ApmeanError$ with $p\geq 1$ 
     to approximate $\forall$ and $\Apmean$ with $p\geq 1$ to approximate $\exists$.
     When $p \geq 1$, these operators resemble the $l^p$ norm of a vector $u=(u_1,u_2,\dots,u_n)$, where $\left\| u \right\|_p = \left( |u_1|^p + |u_2|^p + \dotsb + |u_n|^p \right) ^{1/p}$.
     In our case, many properties of the $l^p$ norm can apply to $\Apmean$ (positive homogeneity, triangular inequality, ...).

\end{example}

\begin{table}
\centering
\begin{tabular}
        {| p{4cm} 
            | >{\centering}m{1.5cm}
            | >{\centering}m{1.5cm} 
            | >{\centering}m{1.5cm} 
            | >{\centering}m{1.5cm}
            | >{\centering\arraybackslash}m{2cm} | }
        \hline
        & \multicolumn{2}{c|}{$(\Tmin,\Smin,\Ns)$} 
        & \multicolumn{2}{c|}{$(\Tprod,\Sprod,\Ns)$}
        & $(\Tluk,\Sluk,\Ns)$ \\
        \hline 
        & $\Ikd$ & $\Igodel$ & $\Ir$ & $\Iprod$ & $\Iluk$ \\
        \hline
        Commutativity of $\land$, $\lor$ & \cmark & \cmark & \cmark & \cmark & \cmark \\
        \hline
        Associativity of $\land$, $\lor$ & \cmark & \cmark & \cmark & \cmark & \cmark \\
        \hline
        Distributivity of $\land$ over $\lor$  & \cmark & \cmark &  &  &  \\
        \hline
        Distributivity of $\lor$ over $\land$  & \cmark & \cmark &  &  &  \\
        \hline
        Distrib. of $\rightarrow$ over $\lor$,$\land$  & \cmark & \cmark &  &  &  \\
        \hline
        Double negation $\lnot \lnot p = p$ & \cmark & \cmark & \cmark & \cmark & \cmark \\
        \hline
        Law of excluded middle & & & & & \cmark \\
        \hline
        Law of non contradiction & & & & & \cmark \\
        \hline
        De Morgan's laws  & \cmark & \cmark & \cmark & \cmark & \cmark \\
        \hline 
        Material Implication & \cmark & & \cmark & & \cmark \\
        \hline
        Contraposition & \cmark & & \cmark & & \cmark \\
        \hline
\end{tabular}
\caption{Common properties for different configurations}
\label{table:summary_properties}
\end{table}
\clearpage
\section{Analyzing Gradients of Generalized Mean Aggregators}
\label{a:gradients}

\citep{van_krieken_analyzing_2020} show that some operators used in 
Fuzzy Logics are unsuitable for use in a differentiable learning setting.
Three types of gradient problems commonly arise in fuzzy logic operators.
\begin{description}
    \item[Single-Passing] The derivatives of some operators are non-null for only one argument.
    The gradients propagate to only one input at a time.
    \item[Vanishing Gradients] The gradients vanish on some part of the domain.
    The learning does not update inputs that are in the vanishing domain.
    \item[Exploding Gradients] Large error gradients accumulate and result in unstable updates.
\end{description}

Tables \ref{tab:summary_gradients_connectives} and \ref{tab:summary_gradients_aggreg} summarize 
their conclusions for the most common operators.
Also, we underline here exploding gradients issues that arise experimentally in $\Apmean$ and $\ApmeanError$, which are not in the original report.
Given the truth values of $n$ propositions $(x_1,\dots,x_n)$ in $[0,1]^n$:
\begin{enumerate}
    \item $ \Apmean(x_1,\dots,x_n) = \biggl( \frac{1}{n} \sum\limits_i x_i^p \biggr)^{\frac{1}{p}}$\\
        The partial derivatives are 
                $ \frac{\partial \Apmean(x_1,\dots,x_n)}{ \partial x_i} 
                = \frac{1}{n}^{\frac{1}{p}} \biggl( \sum_{j=1}^n x_j^p\biggr)^{\frac{1}{p}-1}x_i^{p-1}$.\\
        When $p>1$, the operator weights more for inputs with a higher true value --i.e. their partial derivative is also higher -- 
        and suits for existential quantification. 
        When $p<1$, the operator weights more for inputs with a lower true value and suits for universal quantification.
        \begin{description}
            \item[Exploding Gradients]
            When $p>1$, if $\sum_{j=1}^n x_j^p \to 0$, 
                then $\biggl(\sum_{j=1}^n x_j^p\biggr)^{\frac{1}{p}-1} \to \infty$ and the gradients explode. 
            When $p<1$, if $x_i \to 0$, then $ x_i^{p-1} \to \infty$. 
        \end{description}
    \item $ \ApmeanError(x_1,\dots,x_n) = 1 - \biggl( \frac{1}{n} \sum\limits_i (1-x_i)^p \biggr)^{\frac{1}{p}} $\\
        The partial derivatives are 
                $\frac{\partial \ApmeanError(x_1,\dots,x_n)}{\partial x_i} 
                = \frac{1}{n}^{\frac{1}{p}} \biggl( \sum_{j=1}^n (1-x_j)^p \biggr)^{\frac{1}{p}-1} (1-x_i)^{p-1}$.
        When $p>1$, the operator weights more for inputs with a lower true value --i.e. their partial derivative is also higher --
        and suits for universal quantification. 
        When $p<1$, the operator weights more for inputs with a higher true value and suits for existential quantification.

        \begin{description}
            \item[Exploding Gradients] \ \\
                When $p>1$, if $\sum_{j=1}^n (1-x_j)^p \to 0$, 
                    then $\biggl( \sum_{j=1}^n (1-x_j)^p \biggr)^{\frac{1}{p}-1} \to \infty$ and the gradients explode. 
                When $p<1$, if $1-x_i \to 0$, then $ (1-x_i)^{p-1} \to \infty$.
        \end{description}
\end{enumerate}

We propose the following \emph{stable product configuration} that does not have any of the aforementioned gradient problems:
\begin{align}
    \notzero(x) & = (1-\epsilon)x+\epsilon\\
    \notone(x) & = (1-\epsilon)x\\
    \Ns(x) &= 1-x \\
    \Tprod'(x,y) & = \notzero(x) \notzero(y) \\
    \Sprod'(x,y) &=  \notone(x)+ \notone(y)- \notone(x) \notone(y) \\
    \Ir'(x,y) &= 1 - \notzero(x) + \notzero(x) \notone(y) \\
    \Apmean'(x_1,\dots,x_n) 
        &= \biggl( \frac{1}{n} \sum\limits_{i=1}^n \notzero(x_i)^p \biggr)^{\frac{1}{p}}
        \qquad p \geq 1 \\
    \ApmeanError'(x_1,\dots,y_n) 
        &= 1 - \biggl( \frac{1}{n} \sum\limits_{i=1}^n (1-\notone(x_i))^p \biggr)^{\frac{1}{p}}
        \qquad p \geq 1
\end{align}
$N_S$ is the operator for negation, $\Tprod'$ for conjunction, $\Sprod'$ for disjunction, $\Iprod'$ for implication, $\Apmean'$ for existential aggregation, $\ApmeanError'$ for universal aggregation.

\begin{table}
\centering
\begin{tabular}{
    | p{2.3cm}
    | >{\centering}m{2.4cm}
    | >{\centering}m{2.4cm}
    | >{\centering\arraybackslash}m{2.4cm} | }
    \hline 
    & \thead{Single-Passing} & \thead{Vanishing} & \thead{Exploding} \\
    \hline 
    \multicolumn{4}{|l|}{\quad \textit{Goedel (mininum)}}\\
    \hline
    $\Tmin$,$\Smin$ & \xmark & & \\
    $\Ikd$ & \xmark & & \\
    $\Igodel$ & \xmark & \xmark & \\
    \hline
    \multicolumn{4}{|l|}{\quad \textit{Goguen (product)}}\\
    \hline
    $\Tprod$,$\Sprod$ & & (\xmark) & \\
    $\Ir$ & & (\xmark) &\\
    $\Ikd$ & & \xmark & (\xmark) \\
    \hline
    \multicolumn{4}{|l|}{\quad \textit{\L{}ukasiewicz}}\\
    \hline
    $\Tluk$, $\Sluk$ & & \xmark & \\
    $\Iluk$ & & \xmark & \\
    \hline
\end{tabular}
\caption{Gradient problems for some binary connectives. 
(\xmark) means that the problem only appears on an edge case.}
\label{tab:summary_gradients_connectives}
\end{table}

\begin{table}
\centering
\begin{tabular}{
    | p{2.4cm}
    | >{\centering}m{2.4cm}
    | >{\centering}m{2.4cm}
    | >{\centering\arraybackslash}m{2.4cm} | }
    \hline
    & \thead{Single-Passing} & \thead{Vanishing} & \thead{Exploding} \\
    \hline
    $A_{\Tmin}$/$A_{\Smin}$ & \xmark & & \\
    \hline
    $A_{\Tprod}$/$A_{\Sprod}$ & & \xmark & \\
    \hline
    $A_{\Tluk}$/$A_{\Sluk}$ & & \xmark & \\
    \hline
    $\Apmean$ & & & (\xmark) \\
    \hline
    $\ApmeanError$ & & & (\xmark) \\
    \hline
\end{tabular}
\caption{Gradient problems for some aggregators. 
(\xmark) means that the problem only appears on an edge case.}
\label{tab:summary_gradients_aggreg}
\end{table}

\end{document}